\documentclass[11pt]{article}
\usepackage{geometry}
\usepackage{authblk}
\usepackage[utf8]{inputenc}
\usepackage{amsmath,amsfonts}
\usepackage{xcolor}
\usepackage{algorithmic}
\usepackage{algorithm}
\usepackage{array}
\usepackage{subcaption}
\usepackage{textcomp}
\usepackage{stfloats}
\usepackage{url}
\usepackage{graphicx}
\usepackage{hyperref}
\usepackage{subscript}
\usepackage{pifont}
\usepackage{xspace} 
\usepackage{calc}

\usepackage{makecell}
\usepackage{multirow}

\usepackage{siunitx}
\usepackage[normalem]{ulem}
\robustify\bfseries
\robustify\uline
\def\Decimal{.00}
\def\Uline#1{\Ulinehelp#1 }
\def\Ulinehelp#1.#2 {
  #1.#2\setbox0=\hbox{#1\Decimal}\hspace{-\wd0}{\if\relax#2\relax
    \uline{\phantom{#1.0}}\else\uline{\phantom{#1.#2}}\fi}
    }

\usepackage{pdflscape}
\usepackage{afterpage}
\usepackage{capt-of}
\usepackage{cleveref}

\usepackage[numbers]{natbib}

\usepackage{arydshln}

\usepackage{easyReview}

\crefname{paragraph}{section}{sections}
\Crefname{paragraph}{Section}{Sections}

\usepackage{etoolbox}
\makeatletter
\pretocmd{\NAT@citexnum}{\@ifnum{\NAT@ctype>\z@}{\let\NAT@hyper@\relax}{}}{}{}
\makeatother

\newcommand{\ie}{\textit{i.e.}\xspace}
\newcommand{\eg}{\textit{e.g.}\xspace}

\newcommand\blfootnote[1]{%
  \begingroup
  \renewcommand\thefootnote{}\footnote{#1}%
  \addtocounter{footnote}{-1}%
  \endgroup
}

\title{On the use of Graphs \\for Satellite Image Time Series}

\author[1]{Corentin Dufourg}
\author[1,2]{Charlotte Pelletier}
\author[3]{Stéphane May}
\author[1,4]{Sébastien Lefèvre}
\affil[1]{Universit\'e Bretagne Sud, IRISA, UMR CNRS 6074, Vannes, France}
\affil[2]{Institut Universitaire de France (IUF)}
\affil[3]{Centre National d'\'Etudes Spatiales (CNES), Toulouse, France}
\affil[4]{UiT The Arctic University of Norway, Tromsø, Norway}

\date{}

\begin{document}

\maketitle

\begin{abstract}
The Earth's surface is subject to complex and dynamic processes, ranging from large-scale phenomena such as tectonic plate movements to localized changes associated with ecosystems, agriculture, or human activity. Satellite images enable global monitoring of these processes with extensive spatial and temporal coverage, offering advantages over in-situ methods. In particular, resulting satellite image time series (SITS) datasets contain valuable information. To handle their large volume and complexity, some recent works focus on the use of graph-based techniques that abandon the regular Euclidean structure of satellite data to work at an object level. Besides, graphs enable modelling spatial and temporal interactions between identified objects, which are crucial for pattern detection, classification and regression tasks.
This paper is an effort to examine the integration of graph-based methods in spatio-temporal remote-sensing analysis. In particular, it aims to present a versatile graph-based pipeline to tackle SITS analysis. It focuses on the construction of spatio-temporal graphs from SITS and their application to downstream tasks. The paper includes a comprehensive review and two case studies, which highlight the potential of graph-based approaches for land cover mapping and water resource forecasting. It also discusses numerous perspectives to resolve current limitations and encourage future developments.
\\

\blfootnote{\noindent DOI: \href{http://doi.org/10.1109/MGRS.2025.3622200}{10.1109/MGRS.2025.3622200} \\ 2473-2397 © 2025 IEEE. All rights reserved, including rights for text and data training of artificial intelligence and similar technologies.

}

\end{abstract}

\section{Introduction}
\label{sec:intro}
The Earth is constantly evolving. Its surface follows complex dynamics orchestrated by a wide variety of phenomena acting on all scales, from global movements of tectonic plates, atmosphere and oceans to local changes linked to the life cycles of forests, agriculture or human impacts. Observing the Earth from space enables global studies that are not possible with in-situ measurements, both in terms of spatial coverage and temporal sampling frequency, as satellites capture images of any location in the world at high spatial and temporal resolutions, enabling the study of natural and artificial phenomena on the Earth's surface. According to the Satellite Database maintained by the Union of Concerned Scientists~\cite{grimwood2023}, there were more than one thousand operational satellites dedicated to Earth observation in May 2023.

Thanks to the high satellite revisit capacity, recent data can be arranged as satellite image time series (SITS), forming a sequence of satellite images of the same location but taken at different times. These data provide a wealth of information for real-world applications, such as ecology, healthcare, risk assessment, land management, agriculture, meteorology and climatology~\cite{miller2024deep}. However, SITS represent a large volume of complex data that requires automatic tools for their efficient processing.

\subsection{From object-based to graph-based image analysis}
Indeed, SITS can be voluminous spatially, temporally and spectrally, depending on their resolution in each of these dimensions, leading to a high variability of recorded measurements and a memory-costly storage and processing. By processing datacubes at a pixel scale, most of the existing techniques~\cite{blaschke2001s} struggle to handle these problems. Historically, similar issues have arisen with high-resolution satellite images. Beyond computational complexity, partially solved by hardware~\cite{christophe2011remote} and data management and distribution~\cite{ma2015remote,sudmanns2020big, gorelick2017google, killough2018overview} progress, the Object-Based Image Analysis (OBIA) paradigm was used~\cite{blaschke2008, lang2019geobia} to overcome these limitations. This framework extends the consideration of features to spatial areas rather than independent pixels, by abandoning the regular Euclidean structure of satellite data (\ie, the pixel grid) and considering objects as working units.

OBIA tries to mimic some aspects of human vision such as spatial concepts that group local elements, \ie, pixels, to identify real objects as geographical regions, like a river, a building or a field. To describe these objects, most of the OBIA methods use inherent object attributes such as spectral, textural and geometrical features. However, in some applications, an object is better described by its context and the other surrounding objects (\eg, islands, urban parks), or by its inner components and the relations between them (\eg, residential areas). Therefore incorporating a structural approach with object relationships to OBIA seems primordial~\cite{tiede2014new}. To overcome this issue, a natural and flexible way of modeling relationships between objects is to structure data with a graph.

The graph-based approach retains the advantages of OBIA for representing scenes as sets of geographic objects, but also allows the mathematical tools of graph theory to be used to analyze them. In this context, the definition of an object is extended beyond the purely semantic sense to include any pixel sets aggregated together by an arbitrary criterion. In the most granular case, even individual pixels can be treated as objects. In the remainder of the paper, we will consider this generic definition of an object. Graph representation is memory efficient, suitable not only for computations, but also for human interpretation, thanks to its powerful synthesis and visualization capabilities~\cite{wang2017graphs}. Graphs are highly flexible mathematical tools that can be used to model all kinds of local contextual interactions, \eg, geographical relationships, but also to take into account long-term connections, semantic relationships and experts' prior knowledge of the domain~\cite{sun2022remote}.


Concerning SITS, extensions of OBIA to the temporal dimension have already been proposed to monitor specific objects over time~\cite{petitjean2012, halabisky2018harnessing}.
The representation of spatio-temporal objects in geographic information systems was studied early on, particularly for efficient integration of time in spatial databases~\cite{siabato2018survey}. To efficiently represent phenomena, various methods have been successively developed, from those based on the discretization of space-time into different objects at each change~\cite{armstrong1988, langran1988, hornsby2000, hornsby2002}, to those based on events describing the state of a spatial object at a given instant by the succession of past changes undergone~\cite{peuquet1995, claramunt1995, claramunt1996, worboys2005event}. To combine these two aspects of object evolution, it is possible to focus on both the different states at given times and the changes that led to these states. This can be done by considering the different states as different objects (as in the discretization-based approach) interacting with each other, each relationship signifying a change (as in the event-based approach)~\cite{delmondo2013}. The latter concept clearly extends OBIA with a graph-based approach in a spatio-temporal framework. Thus, spatial and spatio-temporal connections are considered between the different states of the objects, with spatio-temporal relationships between different dates, but also spatial relationships between different entities at the same date to preserve information on the current environment. Relationships between objects are very important, as they retain their context and evolution.

\begin{figure*}[!t]
    \centering
    
    \begin{minipage}[c]{\linewidth}
    \captionsetup{justification=centering}
        \begin{minipage}{.56\linewidth}
            \begin{subfigure}[c]{.29\linewidth}
                \captionsetup{position=top,skip=-10pt}
                \subcaption{Satellite Image\\ Time Series}
        	    \includegraphics[width=\linewidth]{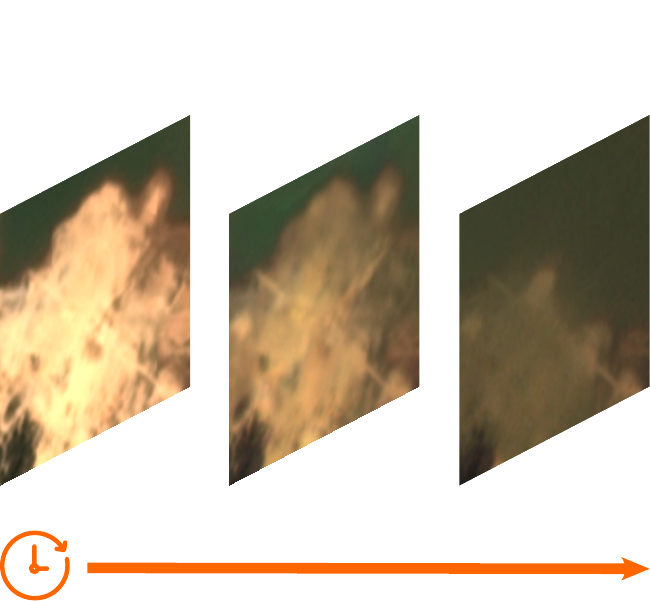}
        	\end{subfigure}
        	\hfill
        	\begin{subfigure}[c]{.03\linewidth}
        	    \captionsetup{position=top,skip=-10pt}
        	    \subcaption*{~\\~}
        	    \includegraphics[width=\linewidth]{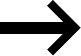}
        	\end{subfigure}
        	\hfill
        	\begin{subfigure}[c]{.29\linewidth}
        	    \captionsetup{position=top,skip=-10pt}
        	    \subcaption{Spatio-temporal objects}
        	    \includegraphics[width=\linewidth]{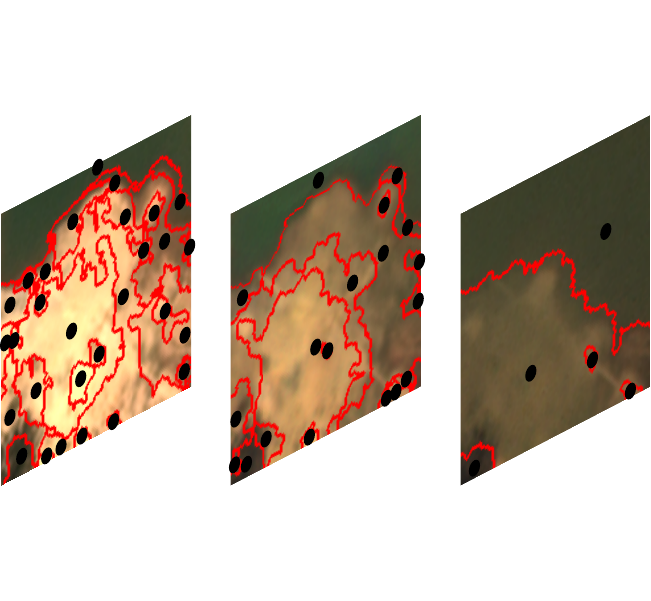}
        	\end{subfigure}
        	\hfill
        	\begin{subfigure}[c]{.03\linewidth}
        	    \captionsetup{position=top,skip=-10pt}
        	    \subcaption*{~\\~}
        	    \includegraphics[width=\linewidth]{img/pipeline/2_arrow.pdf}
        	\end{subfigure}
        	\hfill
        	\begin{subfigure}[c]{.29\linewidth}
        	    \captionsetup{position=top,skip=-10pt}
        	    \subcaption{Spatio-temporal graph}
        	    \includegraphics[width=\linewidth]{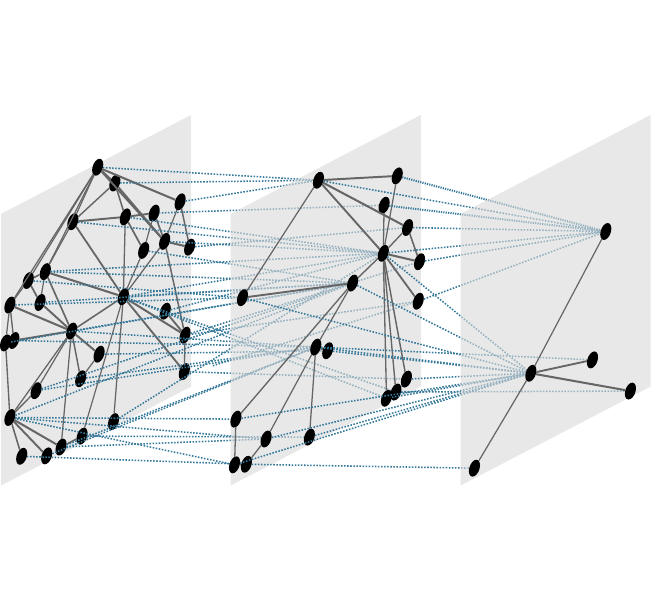}
        	\end{subfigure}
    	\end{minipage}
    	\hfill
        \begin{minipage}{.43\linewidth}
            \begin{minipage}[B]{\linewidth}
	            \centering
	            \begin{subfigure}[b]{.39\linewidth}
	                \centering
	                \subcaption{Visualization}
	                \includegraphics[width=\linewidth]{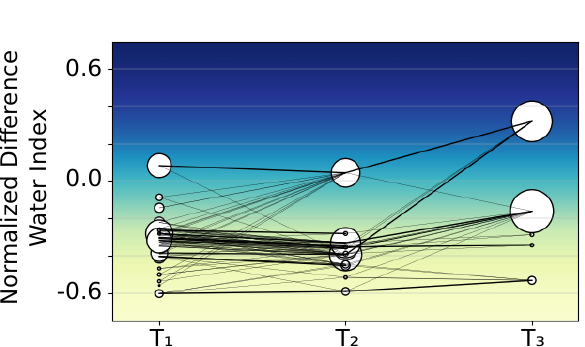}
	            \end{subfigure}
	            \hfill
	            \begin{subfigure}[b]{.44\linewidth}
	                \centering
	                \subcaption{Pattern mining}
	                \includegraphics[width=\linewidth]{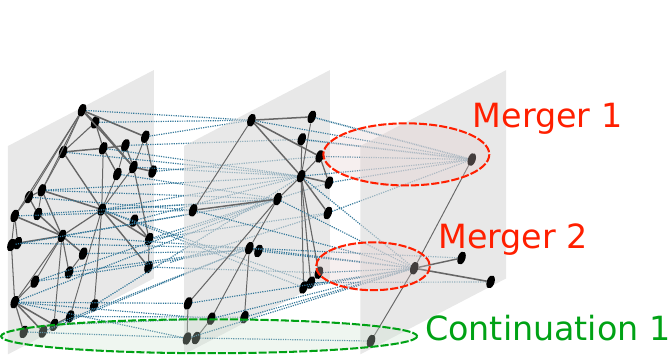}
	            \end{subfigure}
	            \hspace{.05\linewidth}
            \end{minipage}\\
            \begin{subfigure}[c]{.7\linewidth}
                \vspace{.2\baselineskip}
                \includegraphics[width=\linewidth]{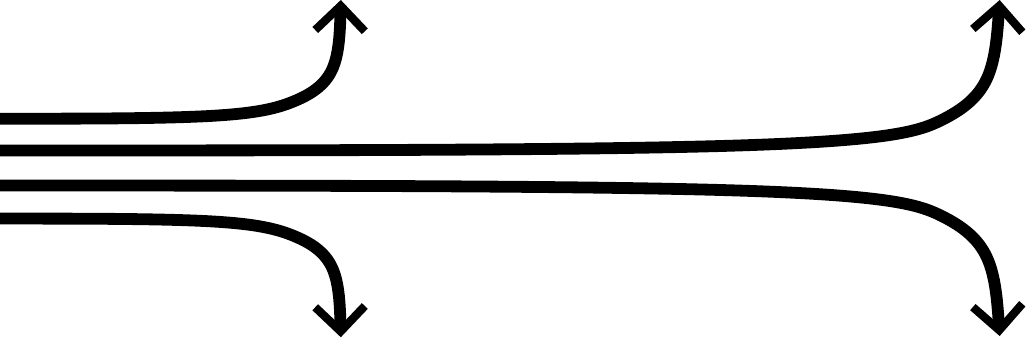}
            \end{subfigure}\\
            \begin{minipage}[T]{\linewidth}
	            \centering
	            \hfill
	            \begin{subfigure}[T]{.33\linewidth}
	                \centering
	                \includegraphics[width=\linewidth]{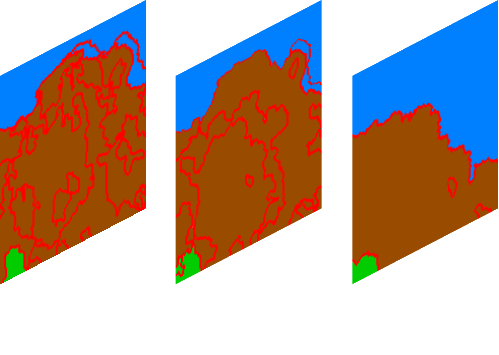}
	                \subcaption{Classification}
	            \end{subfigure}
	            \hfill
	            \begin{subfigure}[T]{.54\linewidth}
	                \centering
	                \includegraphics[width=\linewidth]{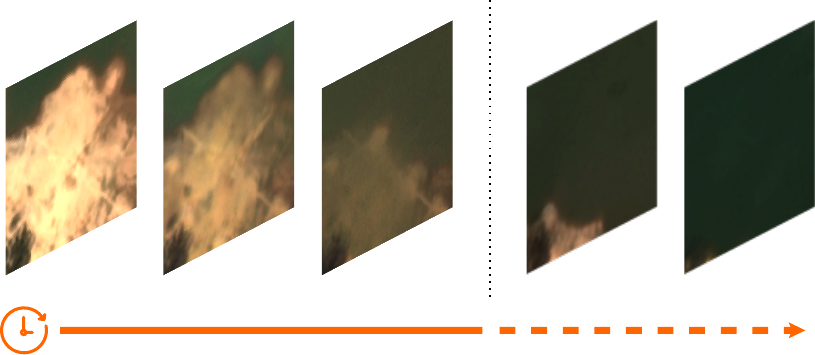}
	                \subcaption{Forecasting}
	            \end{subfigure}
            \end{minipage}
        \end{minipage}
    \end{minipage}
    
    \captionsetup{justification=justified, skip=10pt}
    \caption{
    Pipeline of a spatio-temporal graph used as an intermediate representation of satellite image time series (SITS) for different tasks. (a)~SITS enable the study of phenomena occurring at the Earth's surface (\Cref{sec:SITSacqui}). (b) The entities involved in the captured phenomena are identified, modeled as a set of spatio-temporal objects and outlined in red (\Cref{sec:st_obj}). (c)~The model is completed by spatial and spatio-temporal relationships between entities, forming a spatio-temporal graph (\Cref{sec:inter_obj_relations}). The plot shows entities as black nodes, spatial edges as grey lines and spatio-temporal edges as blue dotted lines. This representation can then be exploited for the targeted application. (d)~Graphs can serve as a visualization tool for experts (\Cref{sec:task_expert_analysis}). The plot illustrates the normalized difference water index (NDWI) temporal profile of the entities, in the same way as in~\cite{guttler2017}, and suggests a soil covering event by water. (e)~Graphs can help to detect predefined or frequent patterns (\Cref{sec:task_graph_pattern_mining}). In the illustration, merger and continuation patterns, as defined in~\cite{xu2021}, are highlighted. (f)~Graphs can be used for classification tasks in time and space (\Cref{sec:task_classification}). (g)~Graphs can efficiently handle the way entities dynamically interact with each other and thus provide anticipation of the future state of a component of the Earth system as part of a forecasting task (\Cref{sec:task_forecasting}).
    }
    \label{fig:pipeline_intro}
\end{figure*}

\subsection{Related surveys and Research questions}

Beyond SITS modeling, graphs have a rich history of applications, including computational chemistry, recommendation systems, image processing, traffic forecasting and physics simulation~\cite{ortega2018graph}. Among these applications, graphs can serve as a powerful tool for interpreting large volumes of data, especially with specific graph data processing techniques that have evolved considerably over the years~\cite{hamilton2017representation, cai2018comprehensive}.
In the field of image processing, graphs have been established as a natural representation for modeling spatial relationships and structures, as highlighted in the seminal book \textit{Image Processing and Analysis with Graphs}~\cite{lezoray2012image}. Extending this perspective to Earth observation, graph-based techniques have also been employed to represent and analyze remote sensing knowledge, as reviewed by \citet{sun2022remote}.
In fact, nowadays, increasing attention is being paid to graph neural networks (GNNs) and their transfer to the field of remote sensing~\cite{ma2024deep, zhao2025beyond}. A recent review offers an overview of GNNs applied to remote sensing and proposes two taxonomies based on either input data or tasks~\cite{khlifi2023}. It identifies the benefits of deep graph learning in remote sensing, while pointing to future directions for improving the interaction between these two fields. The development of spatio-temporal graphs for Earth observation data, including SITS, is recommended for future research by~\citet{zhao2025beyond}, who provide the readers with a qualitative review on GNNs for Earth observation, covering a broad spectrum of application fields and data sources. In contrast, another recent review on SITS \cite{miller2024deep} highlights the potential of graphs for SITS analysis, but provides only a preliminary overview.

These studies have highlighted the benefits of graph structure as a powerful tool for representing and analyzing images, and more specifically remote sensing data. A focus on spatio-temporal data is lacking, despite the interest in such graph-based approaches. In fact, creating a graph from SITS is also complex, due to the wide freedom of assumptions and possible interactions between the objects. Thus, our study provides a broader view of graphs specifically used for satellite-based spatio-temporal data, investigating both graph extraction from SITS and analysis methods, including current trends in deep graph learning. \Cref{fig:pipeline_intro} illustrates the whole graph-based pipeline studied in this paper to process SITS.
With a comprehensive review and two case studies, this paper will hopefully provide the remote sensing and graph learning communities with guidance to overcome the challenges of creating and operating SITS using graphs. To stimulate cross-domain interactions, we guide this paper through three research questions:
\begin{itemize}
    \item How to efficiently model satellite image time series using graphs?
    \item How relevant information can be extracted from such graphs?
    \item How to adapt the proposed pipeline to a given application and solve an Earth observation problem?
\end{itemize}

To answer these questions, the paper is organized as follows. \Cref{sec:background} covers the basics of graph theory and the mathematical considerations required to exploit spatio-temporal graphs. \Cref{sec:graphExtract} describes a methodology for modeling SITS data into a spatio-temporal graph structure, while \Cref{sec:graphAppli} examines existing approaches for processing this structure depending on the downstream task. \Cref{sec:exp} presents two case studies of spatio-temporal graphs for SITS. Finally, \Cref{sec:ccl} concludes our study by discussing potential improvements and future challenges for the graph data model in remote sensing.

\section{Background on Graph Theory}
\label{sec:background}

Graphs are a fundamental tool in many fields for modeling interactions between elements. Those elements are specific to each application domain, \eg, the atoms in a molecule, the junctions in a road network or the people in a social network. In remote sensing, the elements can be the objects identified by OBIA, but we will maintain a general focus on graph theory in this section. It covers the basics, setting out the definitions and terminology used in the paper. The principles of graph theory are addressed in numerous textbooks~\cite{bondy1976graph, west2001introduction}.

\begin{figure}[!b]
    \centering
    \begin{minipage}{.6\linewidth}
    \subfloat[]{
	   \raisebox{-0.5\height}{\includegraphics[clip,width=.3\linewidth]{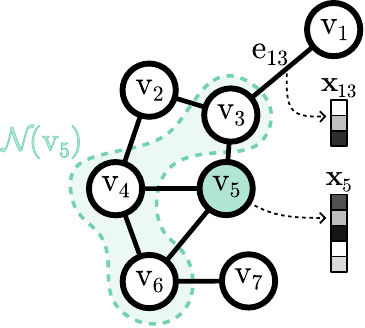}}
    }
    \hfill
    \subfloat[]{
	   \raisebox{-0.5\height}{\includegraphics[clip,width=.3\linewidth]{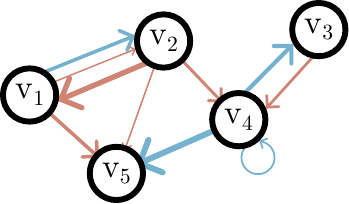}}
    }
    \end{minipage}
	\caption{Examples to illustrate graph terminology. (a)~An undirected graph with node and edge attributes. The green shaded area, denoted by $\mathcal{N}(\mathrm{v}_5)$, depicts the neighborhood of node $\mathrm{v}_5$. (b)~A multi-relational graph with weighted directed edges. The edge width represents the edge weight and orange and blue colored-edges represents two types of relationship.}
    \label{fig:graph_def}
\end{figure}

\subsection{Graph definition}
\label{sec:background:graph}

A \textit{graph} is a set of interconnected elements. Formally, it is represented as $\mathcal{G}=(\mathcal{V},\mathcal{E})$ where $\mathcal{V}=\left\{\mathrm{v}_i\right\}_{i=1}^{n}$ is the set of $n$ elements, called \textit{nodes}, linked together by \textit{edges} $\mathcal{E} \subseteq \mathcal{V} \times \mathcal{V}$. An edge $\mathrm{e}_{ij} \in \mathcal{E}$ corresponds to a link between the connected nodes $\mathrm{v}_i \in \mathcal{V}$ and $\mathrm{v}_j \in \mathcal{V}$. The pair of nodes composing an edge can be ordered $\mathrm{e}_{ij}=\left(\mathrm{v}_i,\mathrm{v}_j\right)$, and the graph is said to be \textit{directed}, or unordered $\mathrm{e}_{ij}=\mathrm{e}_{ji}=\left\{\mathrm{v}_i,\mathrm{v}_j\right\}$, and the graph is then \textit{undirected}.

An important notion related to the context of a node is its \textit{neighborhood}. A node's neighborhood is the set of nodes to which it is connected. The neighbors of $\mathrm{v}_i$ are denoted by $\mathcal{N}(\mathrm{v}_i) = \left\{\mathrm{u}\in\mathcal{V} \mid \left\{\mathrm{u},\mathrm{v}_i\right\}\in\mathcal{E} \right\}$. The cardinality of a node's neighborhood is also called the node's \textit{degree}. In the case of a directed graph, the incoming and outgoing neighborhoods and degrees must be distinguished, as relationships are not bidirectional.

It is worth noting that additional information can be associated with the graph. For example, an \textit{attributed} graph $\mathcal{G}=(\mathcal{V}, \mathcal{E}, \mathcal{X}_\mathcal{V}, \mathcal{X}_\mathcal{E}, \mathcal{X}_\mathcal{G})$ is one where each node, each edge and the graph itself can be associated with specific properties that describe its distinctive features, respectively represented by $\mathcal{X}_\mathcal{V}=\left\{\mathbf{x}_i \mid \mathrm{v}_i \in \mathcal{V} \right\} \in \mathbb{R}^{|\mathcal{V}| \times f_\mathcal{V}}$, $\mathcal{X}_\mathcal{E}=\left\{\mathbf{x}_{ij} \mid \mathrm{e}_{ij} \in \mathcal{E}\right\} \in \mathbb{R}^{|\mathcal{E}| \times f_\mathcal{E}}$ and $\mathcal{X}_\mathcal{G} \in \mathbb{R}^{f_\mathcal{G}}$, with $f_\mathcal{V}$, $f_\mathcal{E}$ and $f_\mathcal{G}$ the dimensions of feature vectors. A graph can also have weights on its edges to model the strength of interaction and is then referred as a \textit{weighted} graph, with $w_{ij} \in \mathbb{R}$ being the weight of $\mathrm{e}_{ij}$. The edges can also represent different kinds of relationships with a \textit{multi-relational graph} having $\mathcal{E} = \mathcal{E}^1 \sqcup  \mathcal{E}^2 \sqcup \cdots \sqcup  \mathcal{E}^r \subseteq \mathcal{V} \times \mathcal{R} \times \mathcal{V}$, where $\mathcal{R}$ is a set of $r$ relationship types and $\sqcup$ is the disjoint union operator. To go further than its topological connections and features, a graph may also be endowed with some spatial geometry, being thus a \textit{geometric} graph~\cite{bronstein2017geometric}. 
The geometrical information  $\mathcal{X}_{P_\mathcal{V}} = \left\{ \mathbf{p}_i \mid \mathrm{v}_i \in \mathcal{V} \right\} \in \mathbb{R}^{|\mathcal{V}| \times d}$ is usually carried by nodes, in addition to other features, on a $d$-dimensional manifold. The most common geometrical feature example for remote sensing data is the spatial coordinates. 
These extended definitions can obviously be combined together according to the degree of realism and complexity required. \Cref{fig:graph_def} shows examples of graphs. In particular, it represents an attributed undirected graph and a multi-relational graph with weighted directed edges.

Previous definitions act as generalizations of the most elementary graph definition, adding further information and improving the model. Differently than those, some graphs can be grouped together into families, such as bipartite graphs and complete graphs, thanks to specific structures and properties~\cite{thomas2023graph}. Another family is that of trees, which correspond to undirected graphs with no cycle. While these tree structures are widely used in remote sensing notably to model hierarchies~\cite{bosilj2018,maia2021}, this paper deals with graphs in general, and therefore with graphs that also include cycles. This enables more comprehensive modeling of phenomena, for example the representation of spatially highly connected networks such as roads, or the tracking of cyclical processes such as successive divisions and mergers of regions over time. Consequently, this paper does not focus on the different graph families, but studies the use of graphs for SITS in its generality via their modeling with a spatio-temporal graph, which could be in practice a tree or any other type of graphs. Readers interested in the use of trees for SITS can refer to several existing application works~\cite{lemen2008, alonso2013, tuna2020component}.

\subsection{Spatio-temporal graph}
\label{sec:background:stg}

A particular graph at the heart of this work is the \textit{spatio-temporal graph}. It describes relationships between elements which lie on a spatio-temporal manifold. These elements can be points or closed domains, in both spatial and temporal dimensions. The relationships, modeled by edges, can be categorized either as spatial, or spatio-temporal. \Cref{sec:graph_creation} describes how objects can be extracted from SITS and provides a formal definition of the different relationships.

A spatio-temporal graph is therefore a multi-relational graph with a set of elements linked between them by spatial and spatio-temporal links, \ie, $\mathcal{G}^{ST} = \left( \mathcal{V}, \mathcal{E} \right)$ with $\mathcal{E} = \mathcal{E}^S \sqcup  \mathcal{E}^{ST}$. Intuitively, $\mathcal{V} = \left\{ \mathrm{v}_i \right\}_{i=1}^n$ is the set of graph nodes representing $n$ elements $\mathcal{O} = \left\{ \mathrm{o}_i \right\}_{i=1}^n$. If a relationship exists between two elements, whether spatial or spatio-temporal, an edge is added to the graph. The edges within $\mathcal{E}^S$ and $\mathcal{E}^{ST}$ correspond to spatial and spatio-temporal relationships, respectively. Since the elements represented by nodes lie on a spatio-temporal manifold, a spatio-temporal graph is inherently geometric.

Although various works in geographic information science and remote sensing used different taxonomies---spatial object temporal adjacency graph~\cite{lemen2009}, geospatial-temporal semantic graph~\cite{brost2014}, evolution graph~\cite{guttler2017}---we follow the definition of~\citet{delmondo2013} and \citet{zeghina2024deep} by calling this structure a spatio-temporal graph.

In the graph literature, there are other mathematical formulations for describing a dynamic phenomenon that changes over time. The most used are the Discrete-Time Dynamic Graph (DTDG) and the Continuous-Time Dynamic Graph (CTDG)~\cite{kazemi2020representation,barros2021survey}. A DTDG is similar to a spatio-temporal graph with a sequence of spatial graphs, but without spatio-temporal relationships to link the elements. A DTDG is therefore less expressive than a spatio-temporal graph. A CTDG represents evolution of a graph as a sequence of node/edge addition and deletion events. However, in existing works using graphs for SITS, objects represented by nodes are usually defined for each date independently. Thus the node set can change greatly between two acquisitions. In a CTDG framework, this can be expressed as a deletion of all previous nodes, followed by the addition of a set of new nodes. This makes it difficult to track the temporal dynamics of a phenomenon using this formulation. Therefore, DTDG and CTDG are unsuitable to model dynamic phenomena captured by SITS, thus explaining our choice to define the spatio-temporal graph as a static graph (\ie, with a fixed topology) with two types of neighborhood, one spatial and the other spatio-temporal.

\subsection{Graph-related tasks}
\label{sec:background:task}

On spatio-temporal graphs, as on any other graph, different tasks exist to extract information and apply it to real cases~\cite{chami2022}. The most common task is \textit{classification}, which involves assigning predefined categories to nodes, edges or an entire graph. For example, node classification consists of associating each node of the graph with a label (\ie, a predefined class). If categories are not predefined, the task is called \textit{clustering}. A special case of node clustering is \textit{pattern mining}, which identifies recurrent structures within a graph.

Sometimes, a discrete classification is not relevant for a specific application and continuous values are required. In that case, a \textit{regression} is applied to nodes, edges or graphs to predict a corresponding continuous value. If the objective is to understand the relationship between the graph and an extrinsic variable, it is coined as \textit{extrinsic regression}, in contrast to \textit{intrinsic regression} which makes prediction on the same space as the input data. The intrinsic regression of a future state on the basis of the current state and previous states is known as \textit{forecasting}.

Another edge-level task is the \textit{link prediction} that consists in making a prediction regarding missing, unobserved or potential connections between nodes.

All these tasks can be carried out by graph analysis experts in their field, but can also be performed by specialized algorithms and automatic learning methods. In the latter case, the supervised tasks (\eg, classification or regression) require labels at the graph, node or edge level. Ground truth labels can also be partial or unavailable due to tedious annotation work, lack of experts or, sometimes, the difficulty of accessing this information. In this case, other learning paradigms should be used, such as semi-supervised, weakly supervised or unsupervised learning.

Learning from graphs involves taking into account both the graph attributes and their topology. Each graph having its specific structure, learning approaches can be separated according to two main perspectives: (i) transductive, and (ii) inductive. Transductive approaches, such as DeepWalk~\cite{perozzi2014deepwalk} and node2vec~\cite{grover2016node2vec}, focus on optimizing embeddings for specific graph structures, meaning they can only operate on known graph structures. These methods are useful for inferring information about or between observed nodes in the graph (\eg predicting labels for all nodes given a partial labeling). In contrast, inductive approaches can produce embeddings on previously unseen graphs and nodes, as they learn a function that maps input to continuous representations, rather than the embedding themselves. This category encompasses all GNN approaches that follow the message-passing framework (which will be presented in \Cref{sec:supervised_methods}), along with methods that model relational functions~\cite{rossi2018deep}.

With their flexibility, graphs are a data structure suitable for a large number of applications. We will detail specific remote-sensing downstream tasks for SITS-based graphs in \Cref{sec:graphAppli}. The design of a spatio-temporal graph also depends on these applications, and its construction pipeline is detailed in the next section.\par

\section{From SITS to graph}
\label{sec:graphExtract}

Graphs can be used for remote sensing in a number of ways, and the flexibility offered by such a structure requires strong assumptions on the data. These assumptions depend both on the acquisition setting and application, and thus must be carefully studied. Taking the example of a building height estimation task, if the data used is acquired at nadir, the building itself contains little information about its height, so it may be crucial for the graph to model the relationship between a building and its cast shadow. With a different off-nadir acquisition, The vertical component of the building appears. Relative size differences between different buildings are informative, and such relationships could help to establish a common frame of reference for height estimation. If, however, the application is different, wishing to determine whether a building has more of a residential, commercial or industrial purpose, spatial relationships with parks, streets and neighboring buildings provide a better understanding of the surrounding context, as the function of a building is often linked to its location. In the context of SITS, the possible relationships are even more extensive by adding the temporal dimension to the spatial and feature dimensions of the data. The design of a graph therefore depends on the desired application and the available data. We identify usually three main steps for the extraction of a spatio-temporal graph from a SITS: (i) collecting the multi-temporal data according to the desired application and understanding the characteristics and constraints of the selected satellite modalities; (ii) identifying the objects that will be used as nodes in the spatio-temporal graph; (iii) considering and integrating the temporal and spatial dimensions of SITS into the graph by linking the objects together with edges.
Each step requires particular caution with regards to the assumptions made, and is associated with limitations that we highlight below.

\subsection{Collecting the satellite image time series}
\label{sec:SITSacqui}

Before exploiting satellite images, we need to know the context in which they were acquired to adjust the graph construction choices. The flexibility of graph means it can handle any type of data, but downstream performance results from its design. The data acquisition context acts as a constraint for better data modeling.

About this context, each satellite mission is thought up to respond to a specific problem with its own needs, whether for studying water resources, monitoring agriculture, or military purposes for example. Depending on the application, satellite measurements for Earth observation have different modalities and can be optical or radar~\cite{zhu2018review}. This means that each SITS is unique with different characteristics, depending on instrument specifications and circulating orbit. Hence, the context to design graphs for SITS applications encompasses temporal, spatial and spectral resolutions, the technical specifications of the acquisition device and the physical properties of the areas observed.

In this section, the focus is given to the spatial and temporal characteristics, overlooking data storage and transmission constraints, and preprocessing such as radiometric correction~\cite{richter2023atmospheric} and geometric realignment~\cite{feng2021advances}. Specifically, we can identify three common types of SITS based on their spatial resolution and revisit time, \ie, the time it takes for the satellite to acquire new data from the same location on Earth.

The first one enables a high temporal sampling (from less than an hour to one or two days), but with a low spatial resolution (hectometric at best). This kind of SITS is provided either by geostationary satellites or wide-swath satellites in low orbit to capture large, highly dynamic structures.

On the contrary, some SITS can have a high spatial resolution (from a few decimeters to several tens of meters) and a low temporal resolution, to focus on local details of long-term phenomena. Site revisits can be regular, such as once a month for low-earth orbit satellites, or irregular, if acquisitions are made on demand by an agile satellite than can point off-nadir.

Finally, acquisition settings can achieve both high spatial and temporal resolution. Adding identical satellites on the same low-Earth orbit, but delayed, forms constellations and improves revisit time without compromising spatial resolution. However, deploying mega-constellations raises the question of low-Earth orbit congestion~\cite{boley2021satellite}.

These differences in acquisition methods mean that SITS have a wide range of spatial resolutions, from a few decimeters to several kilometers, and different temporal samplings, regular or not. There is a third concern regarding data resolution. For optical SITS, it corresponds to spectral resolution ranging from panchromatic acquisition, to multispectral with a few wavelengths, often in the visible and infrared range, and hyperspectral with acquisitions over hundred of narrowed contiguous spectral bands. It should be noted that there is a trade-off between spatial and spectral resolutions, meaning that a system with high spectral resolution can only provide medium or low spatial resolution. For radar data, we can consider the number of frequency bands and the polarization of the acquisition as the radar counterpart of spectral resolution. The ability to distinguish subtle differences in measured energies is called radiometric resolution.

The analysis of SITS is likely to extend beyond the multispectral and radar aspects currently studied. For example, some works manipulate multitemporal hyperspectral images in the form of time series at the pixel level. However, only a few exploit these satellite acquisitions as SITS, exploiting jointly spatial, temporal and spectral dimensions~\cite{halimi2015unmixing,guo2021multitemporal,liu2021bayesian,borsoi2023dynamical,li2024transformer}, certainly due to the lack of methods to process such complex and large data. Moreover, if we consider future satellite planning, the Optical Constellation in Three Dimensions (CO3D) will add depth to the spatial dimension thanks to three-dimensional acquisitions, going beyond the stereo acquisitions of Pléiades and Pléiades Neo, and thus providing multi-temporal 3D point clouds~\cite{lebegue2020co3d}, which can be processed using graph-based methods.

With such a wide range of acquisition techniques, it is possible to imagine large SITS in the extreme case of a substantial number of multispectral acquisitions at very high spatial resolution. This possibility reinforces the need for automatic methods capable of efficiently processing large SITS, as Earth science has entered the era of Big Data~\cite{vance2024big}. We argue in this review that spatio-temporal graphs are capable of handling such data by summarizing a dense datacube to a structured network of objects.

\subsection{Graph creation}
\label{sec:graph_creation}

Once the SITS has been acquired and its characteristics are known, a spatio-temporal graph can be created to effectively represent this data. Considering the definition of a spatio-temporal graph given in \Cref{sec:background:stg}, our formalism allows objects, spatial and spatio-temporal relationships to be defined in different ways, which will be examined in the following sections.

\subsubsection{Entity representation}
\label{sec:st_obj}

The OBIA paradigm, which forms the foundation of our work, builds on the idea that satellite-observable phenomena are inherently spatial and that their constituent entities can be characterized and analyzed. Studied phenomena lie on a spatio-temporal manifold $\mathcal{M} = \mathcal{S} \times \mathcal{T}$, where $\mathcal{S}$ represents the spatial dimensions and $\mathcal{T}$ represents time. A spatio-temporal object $\mathrm{o}_i$ defined by an index $i$ and represented by the node $\mathrm{v}_i$ on a spatio-temporal graph is described by the temporal domain $\mathcal{T}_i \subseteq \mathcal{T}$ during which the entity is observed, by the set of points $\mathcal{P}_i$ occupied in $\mathcal{M}$, and by $\mathrm{x}_i$ any properties inherent to the object. The spatial domain $\mathcal{S}_i$ of the object $\mathrm{o}_i$ is not necessarily spatially continuous, as shown in the definition of the spatial domain in the next section. Mathematically, the object $\mathrm{o}_i$ is thus defined as $\mathrm{o}_i = \left( \mathcal{T}_i, \mathcal{P}_i, \mathrm{x}_i \right)$.

Depending on the characteristics of the acquisition and the downstream task, the entities modeled in the graph can be very diverse and the assumptions used to identify them will differ, both in terms of spatial and temporal domains and object properties. In the following sections, the different options to extract objects from SITS are formalized and studied in light of the existing literature. First, different ways to define $\mathcal{P}_i$ from a single image are provided. Then, this definition of $\mathcal{P}_i$ is extended to a multi-temporal setting using two common points of view. Finally, object feature extraction methods to set $\mathrm{x}_i$ are presented.

\paragraph{Spatial domain}

An entity is represented above all by the evolution of its spatial domain, which corresponds to the points~$\mathcal{P}_i$ in the spatio-temporal manifold~$\mathcal{M}$. It is therefore possible to identify the spatial domain independently for each acquisition date. For the sake of clarity, the assumptions used to define the spatial domain of an entity will be first studied for single-date acquisitions, as a SITS is composed of individual images. The following overview of the various spatial domain definitions is thus given respecting this simplification, considering~$\mathcal{S}_i^t$ as the spatial domain of an object~$\mathrm{o}_i$ at time~$t$.

\begin{figure}
    \centering

    \begin{minipage}{.44\linewidth}
        
        \begin{subfigure}{.43\linewidth}
            \captionsetup{position=top,justification=centering}
            \includegraphics[width=\linewidth]{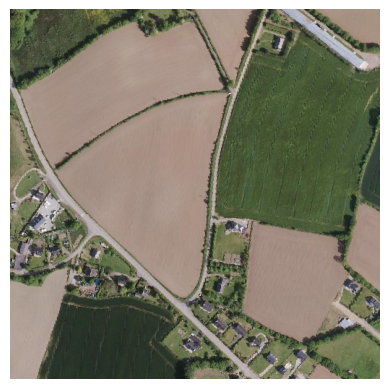}
            \subcaption{Satellite image}
        \end{subfigure}
        \hfill
        \begin{subfigure}{.53\linewidth}
            \captionsetup{position=top,justification=centering}
            \includegraphics[width=\linewidth]{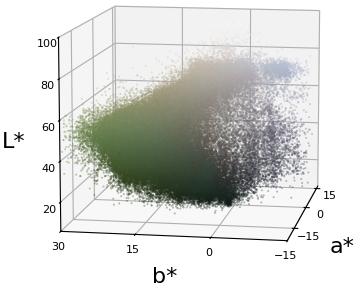}
            \subcaption{CIELAB color~space}
        \end{subfigure}
        
    \end{minipage}\\
    \vspace{.5em}
    \begin{minipage}{.66\linewidth}
        \begin{subfigure}{.3\linewidth}
            \captionsetup{position=bottom,justification=centering}
            \includegraphics[width=\linewidth]{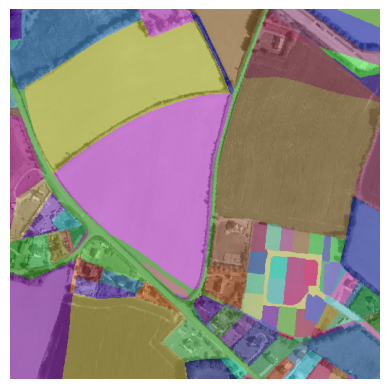}
            \subcaption{Land register parcels}
            \label{fig:spatialObject_parcel}
        \end{subfigure}
        \hfill
        \begin{subfigure}{.3\linewidth}
            \captionsetup{position=bottom,justification=centering}
            \includegraphics[width=\linewidth]{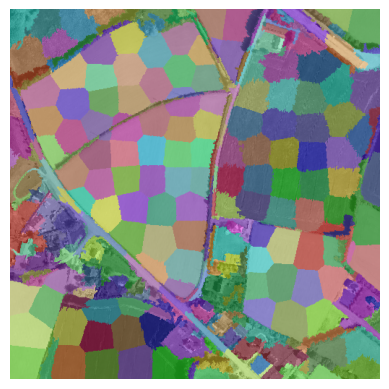}
            \subcaption{Superpixels\\~}
            \label{fig:spatialObject_slic}
        \end{subfigure}
        \hfill
        \begin{subfigure}{.3\linewidth}
            \captionsetup{position=bottom,justification=centering}
            \includegraphics[width=\linewidth]{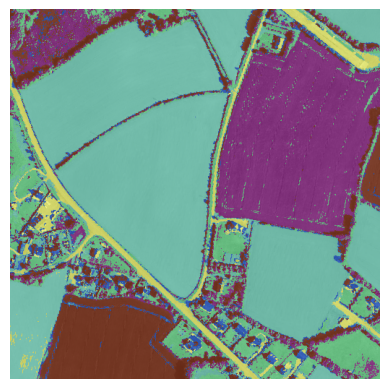}
            \subcaption{Clustering\\~}
            \label{fig:spatialObject_kmeans}
        \end{subfigure}
    \end{minipage}
    
    \caption{
        Various spatial mappings of a satellite image into objects, showed in random colors. (a-b)~Original satellite image pixels plotted respectively along the spatial grid and in the CIELAB color space. (c-e)~Spatial domain of objects defined respectively from a land register, a segmentation algorithm, and a spectral similarity criterion.
    }
    \label{fig:spatialObject}
\end{figure}

When it comes to spatial mapping between satellite images and objects, the objects identified tend to correspond as closely as possible to the elements of the observed phenomenon. This requires either an identification by the user, auxiliary data such as land registers~\cite{garnot2020} (as in \Cref{fig:spatialObject_parcel}) or land cover maps~\cite{zou2023}, or going through task-dependent expert algorithms as in~\cite{yi2014, xue2019} for ocean applications.

In some cases, manual mapping, auxiliary data or expert algorithms are unavailable or unsuitable due to too much variability between observed objects. Therefore it may be useful to approximate real entities with abstract objects based on some assumptions. This automates the identification process by defining objects with pre-established rules.

The most common spatial assumption is to consider objects as regions, \ie, sets of connected pixels, that are spectrally homogeneous. Segmentation algorithms, such as Simple Linear Iterative Clustering (SLIC)~\cite{achanta2012}, Felzenszwalb~\cite{felzenszwalb2004} and Multi-Resolution Segmentation from eCognition~\cite{benz2004}, can be used for this purpose. \Cref{fig:spatialObject_slic} shows an example of such a case. Depending on their nature, images can also be discretized into convex polygons according to a mesh~\cite{cohen1996optical,horritt2001predicting, ge2024non}, as in the finite element method. The choice of segmentation and meshing algorithms depends on the task, and the size and shape of the objects observed~\cite{hossain2019}. State-of-the-art segmentation methods rely on AI-powered approaches using pre-trained foundation models, such as the Segment Anything Model~\cite{kirillov2023segment}, adapted to a wide variety of objects. An extreme case of a spatial region, useful for coarse resolution images with sub-pixel entities, is the one-pixel-one-object mapping, as used in~\cite{tulczyjew2022} for super-resolution. Note that patch-based methods, such as Convolutional neural networks (CNNs) using the context around each pixel or Vision Transformers arbitrarily slicing the image into rectangular regions, are outside the scope of this study. In fact, even if CNNs and Transformers can be considered as specific cases of GNNs, their data structure does not imply any semantic meaning related to the physical entities observed. This therefore goes beyond our view of graphs for SITS as an extension of OBIA.

Another spatial assumption is that pixels with similar values act in the same way, forming an object even if its parts are disconnected~\cite{heas2004a,heas2004b,heas2005}, as illustrated in \Cref{fig:spatialObject_kmeans}. Consequently, mapping this set of similar disconnected pixels to an object implies a greater importance of feature dimension over spatial dimension and may result in some impulse noise in the downstream task.

In either case, it should be noted that homogeneity and similarity criteria to aggregate pixels into objects can be calculated from raw data, but also from learned semantic features~\cite{jampani2018}. Another solution is end-to-end object learning to obtain objects that are more specific to a task, and therefore perform better in the downstream processing of the graph~\cite{yang2020superpixel, ma2021fast, xu2023esnet}. However, the learning cost is higher and the adaptability poorer, requiring re-training for each different task. The storage advantage is also lost, as the full SITS is always needed in memory for each learning iteration, whereas a spatio-temporal graph only needs to be built once as a data preprocessing step.

Finally, these processes aggregate pixels of an image to identify different objects, each corresponding to an entity.
In short, objects are identified by their spatial domains, resulting from assumptions on the data to model real entities. Hence, to pursue these efforts to be realistic, the evolution of these entities over time, in terms of both their spatial domain and their attributes, must be taken into account.

\begin{figure}[h]
    \centering
    \subfloat[Endurantism]{
	   \includegraphics[clip,width=.42\linewidth]{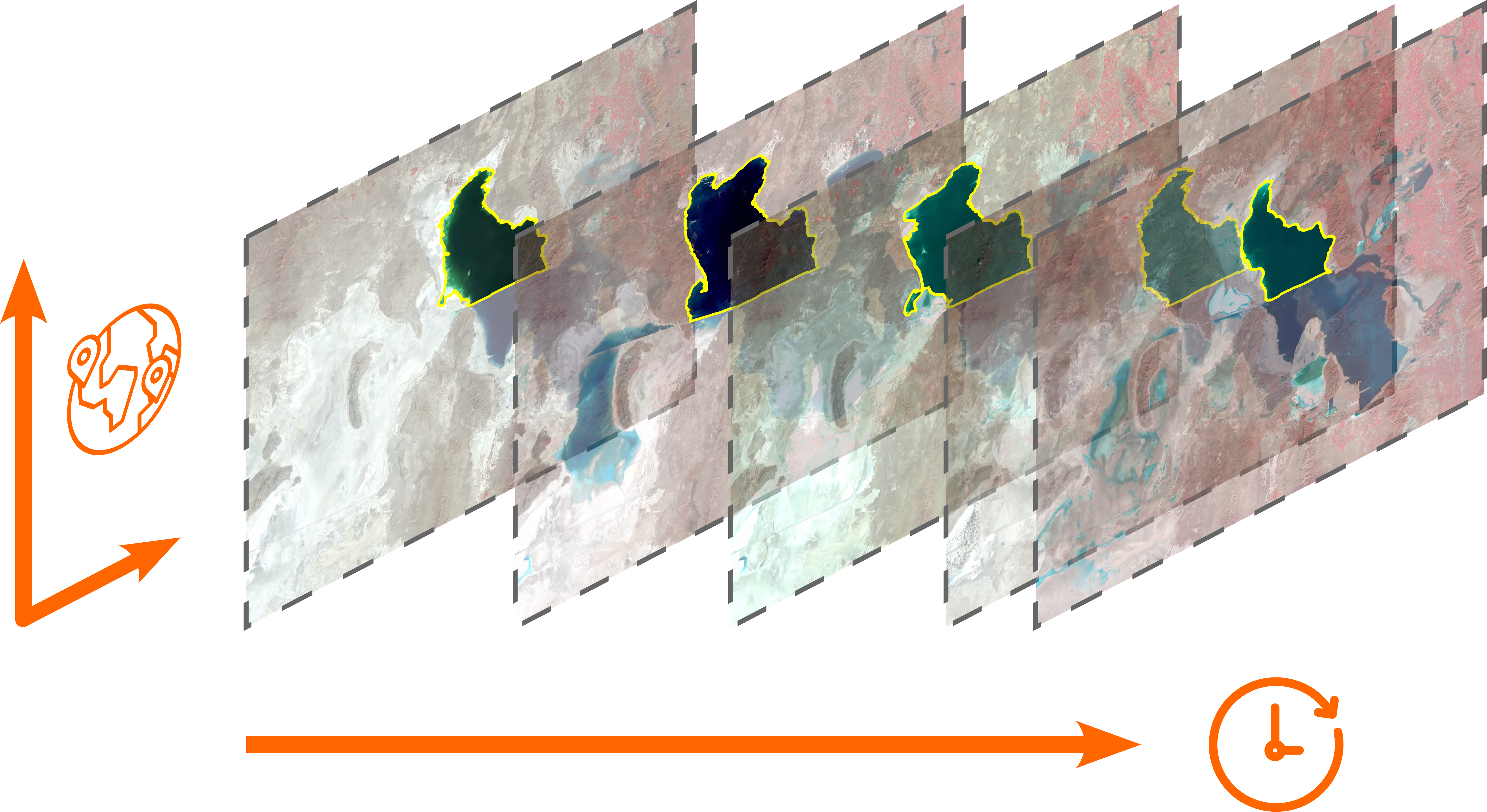}
	   \label{fig:spacetime_endure}
    }
    \hfill
    \subfloat[Perdurantism]{
	   \includegraphics[clip,width=.42\linewidth]{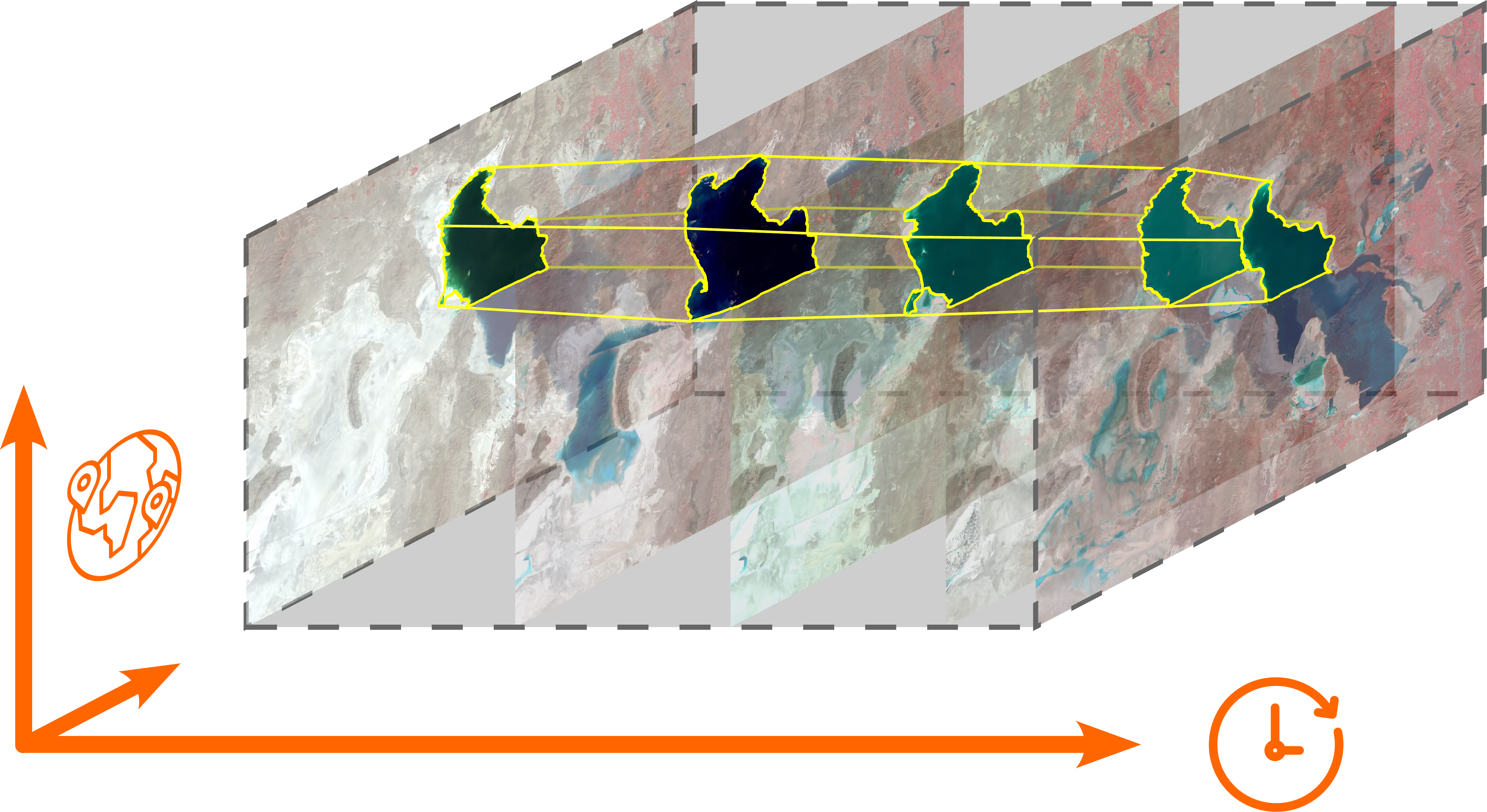}
	   \label{fig:spacetime_perdure}
    }
	\caption{Illustration of the endurantist and perdurantist paradigms for an object-based analysis of SITS. Acquisitions depict the evolution of Great Salt Lake water area; credit USGS. (a)~According to endurantists, an object is wholly present at each moment of its life, with intrinsic properties as spectral signatures and spatial extent. The yellow outline shows the same object, but at different times and therefore with changing properties. (b)~According to perdurantists, an object is a set of successive instantaneous temporal parts, occupying a spatio-temporal region. The yellow outline designates the whole spatio-temporal object where certain temporal parts are highlighted as slices of the object.}
    \label{fig:spacetime}
\end{figure}

\paragraph{Temporal domain}
When analyzing a SITS, an entity may be observed over several dates, and the various acquisitions are not necessarily independent. Thus, the spatial domain of an object may persist and evolve over time. Then, considering a temporal dimension to objects allows us to take into account the temporal granularity of SITS, leading to spatio-temporal objects. Similar to the spatial dimension, temporal assumptions must be made to define spatio-temporal objects on the basis of available data and application requirements. The way objects persist over time is also linked to the metaphysical paradigm considered to integrate space and time. There are two views, neither of which enjoys consensus: endurantism and perdurantism. They are known as three-dimensionalism and four-dimensionalism, by considering our world lying in three dimensional space or four dimensional space-time~\cite{sider2001}. The former treats physical objects as entities that are observable entirely at each moment of their lifetime, \ie, at a given date the whole entity is present (see \Cref{fig:spacetime_endure}).
Those objects support time-evolving properties, such as spatial extent and spectral reflectance changes. Consequently, it separates the two realms of space and time. The latter postulates that objects have extension in both time and space, treating space and time as a unified and fundamental domain. At each moment, only one part of the object is present, not the object as a whole (see \Cref{fig:spacetime_perdure}). We refer the reader to the work of \citet{galton2004fields} for a more in-depth perspective on endurantism and perdurantism in relation to an object-based representation of the world.

To put these concepts in relation with the object-based analysis of SITS, the endurantists consider an object $\mathrm{o}_i$ as a time-indexed set of regions $f_i(t)$ in the spatial manifold, meaning the object changes over time without being subdivided; $f_i : \mathcal{T}_i \rightarrow \mathcal{S}_i$ with $\mathcal{S}_i$ a subset of $\mathcal{S}$ delimiting the spatial domain, which varies over time for the object $\mathrm{o}_i$.
The pixel set occupied by an object $\mathrm{o}_i$ according to endurantism thus corresponds to its complete trajectory in the spatio-temporal manifold, \ie, $\mathcal{P}_i = \bigcup_{t \in \mathcal{T}_i} \left\{ f_i(t) \right\}  \subseteq \mathcal{M}$.
Therefore, satellite acquisitions correspond to snapshots of the object's properties at a given time.
This view is adopted in~\cite{yi2014} for object tracking, identifying individual objects at each date and associating them over time to study changes in their properties.

From another perspective illustrated in~\Cref{fig:spacetime_perdure}, perdurantism depicts the world at a given moment as a time slice $\mathcal{S}_i^t$ of spatio-temporal objects, $\mathcal{P}_i = \bigcup_{t \in \mathcal{T}_i} \left\{ \mathcal{S}_i^t \right\}  \subseteq \mathcal{M}$.
Consequently, objects can have different lifetimes depending on the SITS, ranging from one to several acquisition dates.
Therefore, as geographical entities have a time-varying shape and a lifetime independent of the SITS duration, their corresponding spatio-temporal objects are 3D-shaped with spatial and temporal parts. Then, joint spatio-temporal segmentation is used to take into account both the spatial and the temporal geometry of objects~\cite{lemen2009,li2021}. However, while considering both the spatial and temporal aspects of the data, these methods can be cumbersome and not scalable. Moreover, they still focus on 2D segmentation per acquisition, even though temporal continuity is taken into account. We note that most existing segmentation algorithms, for which an efficient implementation exists, are tailored for single images, not sequences of images.
This explains that the literature often disregards the temporal information, and thus the 3D-shape of spatio-temporal objects. To model SITS in the object-based paradigm, two main strategies based on simplification assumptions are used.

The first one considers objects living throughout the whole SITS acquisition period. The temporal dimension of these spatially-fixed objects can be modeled as a multivariate time series of their attributes. Objects can be identified either from an image at a reference date, asserting that the land units defining a given scene undergo minimal changes during the time interval covered by the SITS~\cite{censi2021, interdonato2020, zangari2021}, or from a multi-temporal composite image, over-segmenting the spatial domain to capture the static and dynamic boundaries of the entities~\cite{desclee2006forest, watkins2019comparison}.

The second strategy assumes that geographical entities are independent between acquisition dates, due to a low revisit rate for example, and objects are identified only spatially in each image over time~\cite{guttler2017}, facilitating the use of spatial geometric information. This case, while different from endurantism in its approach, is similar in its application to object identification. However, links modeled by spatio-temporal edges in a perdurantist setting will have a different meaning than spatio-temporal edges in an endurantist perspective, demonstrating the value of thinking through the paradigm to be adopted before modeling a SITS by a graph. In fact, establishing spatio-temporal relationships between the different temporal parts of an object is crucial in this setting (\Cref{sec:temporalRelations}).

As studied by \citet{galton2004fields}, the vision used for the spatio-temporal manifold depends both on the desired application and on the characteristics of the objects over time. On the one hand, endurantism is more likely to be adopted for tracking object movements because of its ability to handle spatial changes, for example displacement, as a time series of a object attributes~\cite{yi2014}. On the other hand, perdurantism is often better suited to analyze the spectral evolution of regions or follow the splits and merges of parcels~\cite{guttler2017} thanks to its ability to isolate objects into independent spatio-temporal regions. However, whichever approach is chosen, they are not mutually exclusive, and are both consistent with the notion of spatio-temporal objects, and by extension spatio-temporal graphs.

Let us note that one or the other point of view, i.e. endurantism or perdurantism, is not directly claimed by the cited articles and results from our personal interpretation of the works to illustrate the notions in a tangible way.

\paragraph{Object feature extraction}
\label{sec:obj_feat}

Once the objects are well defined in space and time, it is important to consider their inherent characteristics to get the most out of them. In many cases, it is those features that carry useful information for downstream tasks.

There are two ways of extracting meaningful features from objects of arbitrary shape, whether a 2D object extracted from a single satellite image or a 3D spatio-temporal object from a SITS. The features can be predefined and hand-crafted, or they can be data-dependent and learned.

\begin{figure}
    \centering
    \subfloat[Hand-crafted features]{
	   \includegraphics[clip,width=.45\linewidth]{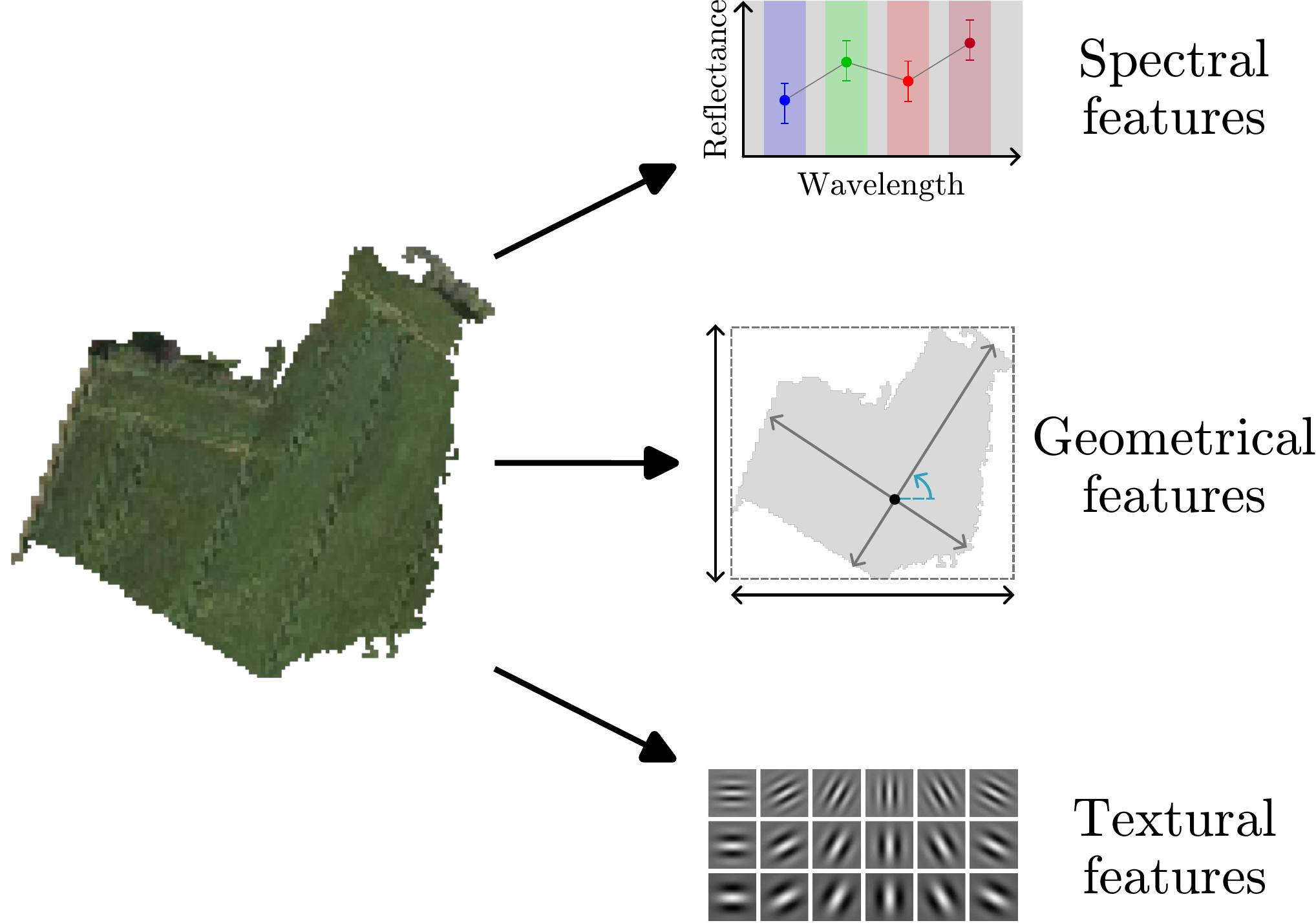}
	   \label{fig:featureExtraction:handcrafted}
    }
    \hfill
    \subfloat[CNN-based features]{
	   \includegraphics[clip,width=.45\linewidth]{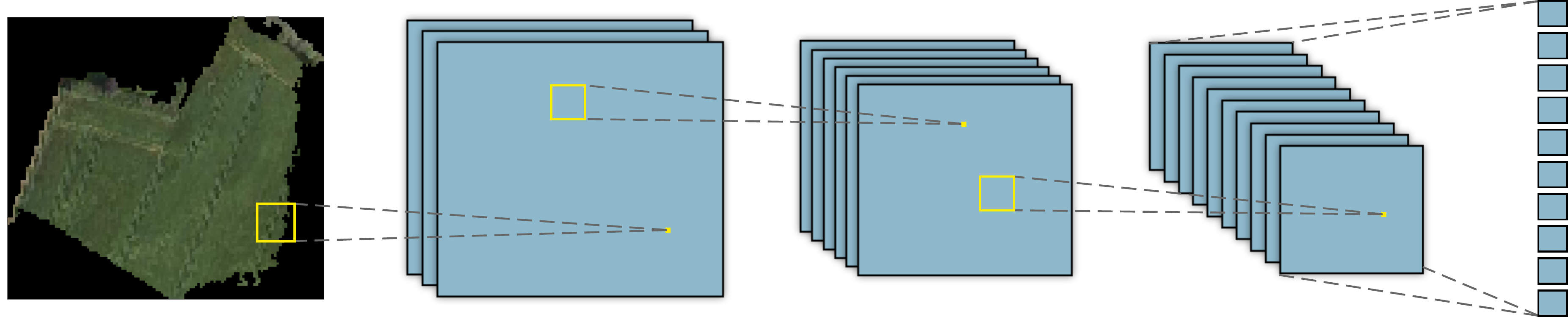}
	   \label{fig:featureExtraction:cnn}
    }
    \vspace{1.5em}
    \subfloat[GNN-based features]{
	   \includegraphics[clip,width=.45\linewidth]{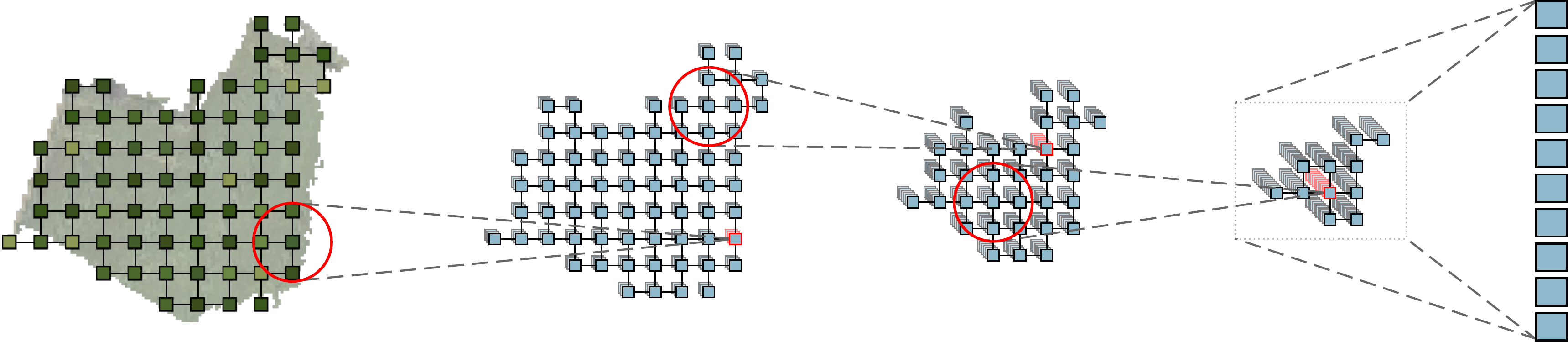}
	   \label{fig:featureExtraction:gnn}
    }
    \hfill
    \subfloat[Pixel-Set Encoder]{
	   \includegraphics[clip,width=.45\linewidth]{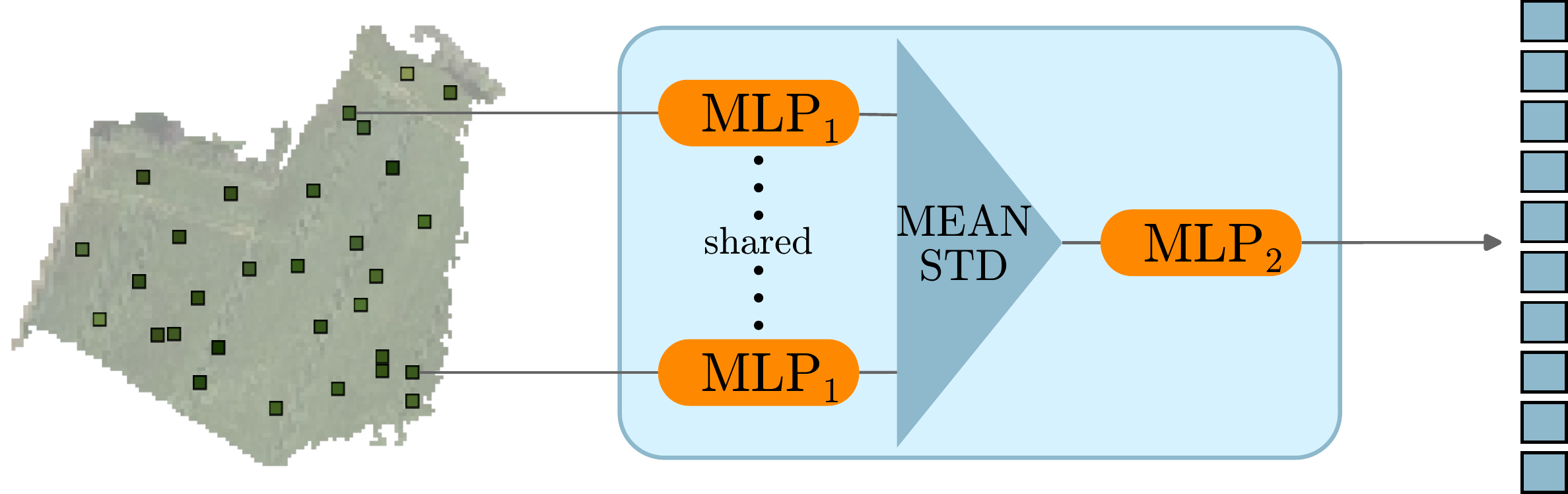}
	   \label{fig:featureExtraction:pse}
    }    
    
	\caption{Four ways to extract spatial object features with the region boundaries and satellite images as input. (a)~Compute spectral, geometrical and textural features in a manually-defined process. (b)~Extract deep features using a convolutional neural network on the padded or cropped region. (c)~Apply graph neural network to the region's pixel graph. (d)~Sample a fixed number of random pixels and compute statistical descriptors using a pixel-set encoder; adapted from~\cite{garnot2020}.}
    \label{fig:featureExtraction}
\end{figure}

The first category computes features designed and selected by experts of the field depending on the input data and the application. As in \Cref{fig:featureExtraction:handcrafted}, these features are usually divided in three domains: spectral, textural, and geometrical~\cite{blaschke2014geographic}.

Spectral features are linked to the way observed entities interact with light, informing about chemical compositions and structures. For optical acquisition, it is linked with the reflection spectrum of incident light at different wavelengths. Spectral features usually include the region's average reflectance at given wavelengths, but also computed spectral indices such as Normalized Difference Vegetation Index (NDVI) and Normalized Difference Water Index (NDWI)~\cite{xue2017significant, ma2019waterspectral}. For radar signal, features can be linked to the polarization of the signal or to difference in signal intensity due to wave scattering.

Texture is more related to the spatial distribution of pixel values inside the object. Range, second-order statistics, entropy, histogram of values, Gabor wavelet decomposition, Haralick features, and pattern spectrum are some existing estimators used to analyze image texture~\cite{humeau2019, petrou2021image}.

Geometrical features tend to describe the position, size and shape of objects. The position of a region can be described by the coordinates of its center of mass, and its size by the area of the region. Shape is more complex to encode with a limited number of features, but compactness, solidity, elongation, Hu moments and histogram of oriented gradients are some useful estimators~\cite{zhang2004}.

The advantages of hand-crafted features lie in the integration of prior knowledge that is fully understandable and explainable. However, this can also be a hard process if the task is too complex. This means that some of the selected features may be redundant, not discriminant and, in the worst case, add noise, with the risk of falling into the curse of dimensionality~\cite{georganos2018less}.

This introduces the second way of extracting object features, through machine learning. In particular, deep learning automatically extracts features that are data-dependent and therefore more adaptable to a target task. Unlike hand-crafted features, there is less control over the meaning of these deep features, and they generally have no real-world interpretation. Nevertheless, the deep learning features can be exploited into an object-based pipeline to make the most of the amount of data at their disposal~\cite{ma2024deep}.

CNNs are known to be powerful in processing regular images and extracting descriptors. Therefore, a CNN can be applied to the entire input image, and then the features extracted from each object are the average value of the CNN feature vectors over their region~\cite{diao2022}. The fact that region segmentation is only used after feature computation implies that OBIA principles are only partially applied.

To use CNNs as a feature extractor by region, an early work~\cite{audebert2016} applies a CNN backbone on multi-scale patches centered on each region. This allows the extraction of inner-object features but also contextual ones. It seems therefore more difficult to retain fine-grained information about objects, as their boundaries have been ignored and their spectral, textural and geometrical characteristics have been potentially poisoned by their surroundings.

To overcome this issue, learning with the object context is abandoned---it can be further taken into account thanks to a neighborhood graph---in favor of retaining the initial region geometry, by padding the area around the region with zeros to get a rectangular patch for the CNN backbone~\cite{garnot2019, ma2019, ahmad2023, kavran2023}. Preserving the geometry of the initial region can be important, so that the CNN can extract the region's inherent features, in particular its geometrical properties. \Cref{fig:featureExtraction:cnn} illustrates this solution.

Another way of respecting object geometry when using convolution is to use deformable CNNs, which modify the shape of the convolution kernel with a learned offset based on region boundaries~\cite{zhao2022}.

It is also possible to use graph neural networks to extract spectral and textural features while fully respecting object boundaries, as illustrated in \Cref{fig:featureExtraction:gnn}. The authors of \cite{jia2024} and~\cite{cao2024} exploit them to preserve low-frequency information from their superpixel regions, which already denote a certain homogeneity, and they demonstrate the effectiveness of GNN-based features compared to more conventional statistical indicators.

Finally, on some images with a medium spatial resolution, such as those acquired by \mbox{Sentinel-2}, textural information is not well captured and therefore not very meaningful. Taking this into account, the Pixel-Set Encoder~\cite{garnot2020} module, plotted in \Cref{fig:featureExtraction:pse}, focuses on spectral information, learning statistical descriptors of the object's spectral distribution. It also uses simple hand-crafted geometrical features to account for the shape and the size of the region.

It is more common to meet these descriptors, whether hand-crafted or deep, with 2D spatial objects but they can also be extended to spatio-temporal objects, with features computed jointly on the spatial and temporal domains~\cite{tuna2020}, with temporal profiles of spatial features~\cite{censi2021}, or with time-specific features, such as phenological parameters~\cite{valero2016production, jonsson2002seasonality}.

In the context of Earth observation, geographical coordinates of the regions in the form of spatial positional encoding may be useful~\cite{klemmer2023positional}. Similarly, temporal~\cite{garnot2020} or thermal~\cite{nyborg2022generalized} positional encoding and other SITS metadata can also be used to attach additional information to objects.

\subsubsection{Inter-entity relationships}
\label{sec:inter_obj_relations}

Now that SITS have been divided into several objects, the identified entities can be used independently, which is common in most OBIA works. However, as exposed previously, the spatial and temporal context of each entity, in addition to its intra-object information, is crucial to understand the nature of the object and its evolution~\cite{dufourg2023b}.

Certain design choices for spatio-temporal objects may have led to the incorporation of temporal information into their features, such as temporal profiles~\cite{censi2021,interdonato2020} and temporal position encoding~\cite{keisler2022,lam2023learning}. Some objects also already possess intra-object spatial information in their features. While such information may be sufficient, it is limited to the intra-object context and can hardly be extended to long-term monitoring, which involves discretizing the temporal domain in addition to the spatial one. This limitation can be overcome by combining intra- and inter-object information to understand the overall temporal and spatial context at different scales.
Specifically, this context is maintained within the spatio-temporal graph by means of relationships of various kinds between entities. In the following, we define and describe \textit{spatial}, then \textit{spatio-temporal} relationships.

\paragraph{Spatial relationships}
\label{sec:spatialRelations}

SITS provide spatial information that can be exploited with spatial relationships between entities in a spatio-temporal graph. This type of relationship assumes that entities observed at the same time can be linked. We therefore formally consider a relationship as spatial if the time domains of the two linked objects are identical, $\forall \mathrm{e}_{ij} \in \mathcal{E}^S, \mathrm{e}_{ij} = \left(\mathrm{v}_i, \mathrm{v}_j\right) \mid \mathcal{S}_i \neq \mathcal{S}_j \; \text{and} \; \mathcal{T}_i = \mathcal{T}_j$. Hence, spatial relationships of an object represent its context during its lifetime.

\begin{figure}
	\centering

	\subfloat[Region adjacency]{
		\includegraphics[clip,width=.22\linewidth]{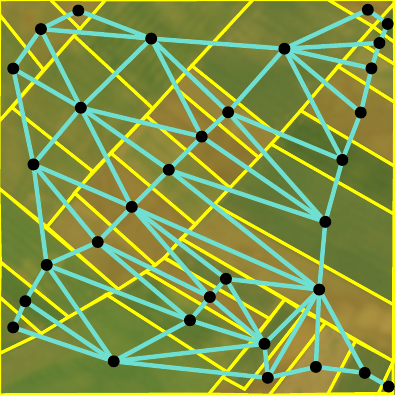}
		\label{fig:spaEdge:adj}
	}
	\hfill
	\subfloat[$\epsilon$-ball]{
		\includegraphics[clip,width=.22\linewidth]{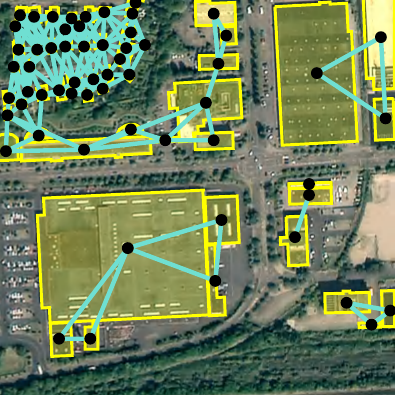}
		\label{fig:spaEdge:eball}
	}
	\hfill
	\subfloat[Feature similarity]{
		\includegraphics[clip,width=.22\linewidth]{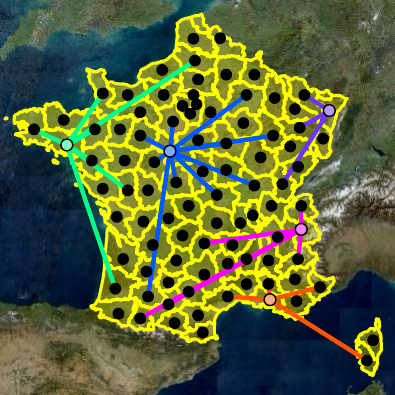}
		\label{fig:spaEdge:featsim}
	}
	\hfill
	\subfloat[Learned edges]{
		\includegraphics[clip,width=.22\linewidth]{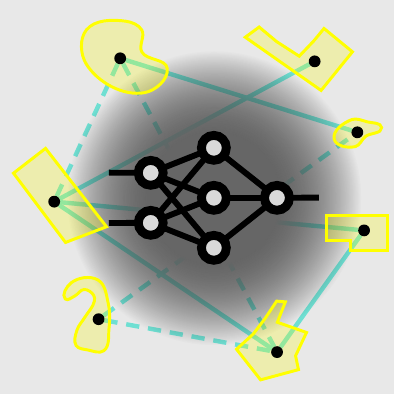}
		\label{fig:spaEdge:learned}
	}
	
	\caption{Different design of spatial relationships in spatio-temporal graphs, illustrated by a few use cases. (a)~Topological relationships, such as region adjacency, illustrate a direct spatial connection between two objects. For example, it is used here to link two adjacent fields and helps to share certain physical characteristics of the local environment, such as temperature or air moisture, external to the very nature of the entity being modeled. (b)~Proximity-based relationships, here denote with an $\epsilon$-ball, can help recognize spatial groupings of objects sharing common characteristics, as is the case with residential, industrial and commercial building functions. (c)~Similarity of climatic features, here for the departments of metropolitan France. One node per climate zone is highlighted and connected to a few other nodes near or far. (d)~Learned edges in an end-to-end manner.}
	
	\label{fig:spatial_edges_def}
\end{figure}

The most direct use of spatial relationships is to connect neighboring regions to counter the limitations induced by hard segmentation. Inspired by Tobler's First Law of Geography---``Everything is related to everything else, but near things are more related than distant things''~\cite{tobler1970computer}---, region adjacency graphs identify the adjacency connection between entities as spatial relationships, as in \Cref{fig:spaEdge:adj} with crop fields. Adjacency is a type of topological relationship among others but spatial relationships in spatio-temporal graphs can be extended to all kinds of topological relationships, from the region connection calculus 8 (RCC8)~\cite{randell1992rcc8} to the 9-intersection model~\cite{clementini1993small}, as long as they are semantically adapted to the application case. Based on the positions of objects, it is also possible to define proximity graphs such as $\epsilon$-ball graph, $k$-nearest-neighbor graph, and Delaunay triangulation~\cite{lezoray2012image}. An example of $\epsilon$-ball graph is depicted in \Cref{fig:spaEdge:eball}, which links buildings close to each other.

Topological relationships are based on the local geometrical context of an object, but it can be crucial to allow long-range connections between distant entities for certain applications. For example, on a global scale, different regions of the world share the same type of climate and may therefore share similar spatio-temporal features, as it is the case in \Cref{fig:spaEdge:featsim}. Thus, two distant objects can be connected by the similarity of their features~\cite{ning2024harnessing}. It is also possible to learn these task-dependent relationships in an end-to-end manner as in~\cite{cachay2021world, liu2021self, zhao2024adaptive}~(\Cref{fig:spaEdge:learned}).

It should be pointed out that the same graph can handle different types of spatial relationships, making it a heterogeneous graph. For example, both adjacency and similarity edges can be used to make the most of short- and long-distance relationships~\cite{interdonato2020}.
Finally, the importance of the spatial relationship can also be established using a weighted graph, \eg, border length or geographical distance can affect the rate of information exchange between two regions.

\paragraph{Spatio-temporal relationships}
\label{sec:temporalRelations}

SITS are mainly exploited for their temporal dimension, as they enable us to study dynamics over time that independent images cannot provide by their very nature. This temporal nature is exploited through temporal and spatio-temporal relationships in graphs or through time series analysis at the pixel~\cite{pelletier2019temporal} or object level~\cite{petitjean2012}. These edges can be exploited conjointly with spatial relationships (see \Cref{sec:graphAppli}).

The temporal context of an object can be crucial to track its evolution, and thus it can be modeled by temporal relationships between spatio-temporal objects. Following the philosophy of definition of spatial relationship, we consider a temporal relationship if the two linked objects have the same spatial domains,  \ie, $\forall \mathrm{e}_{ij} \in \mathcal{E}^T, \mathrm{e}_{ij} = \left(\mathrm{v}_i, \mathrm{v}_j\right) \mid \mathcal{S}_i = \mathcal{S}_j \; \text{and} \; \mathcal{T}_i \neq \mathcal{T}_j$. This definition implies that the space domain does not evolve over time, which in fact represents only a few scenarios. By relaxing the constraint on the spatial domains, we extend it to spatio-temporal edges for all relationships that connect spatio-temporal objects with different temporal domains:  $\forall \mathrm{e}_{ij} \in \mathcal{E}^{ST}, \mathrm{e}_{ij} = \left(\mathrm{v}_i, \mathrm{v}_j\right) \mid \mathcal{T}_i \neq \mathcal{T}_j$. This comprehensive definition enables modeling the spatio-temporal context of each entity, and therefore its past and future, through relationships over varying distances, even if the acquisition dates are irregular. Temporal order can even be modeled by relationships directed from past to future with an oriented graph, providing a good structure for studying the causality of events.

\begin{figure}
	\centering
    \subfloat[(Semi-)manual tracking]{
		\includegraphics[clip,width=.3\linewidth]{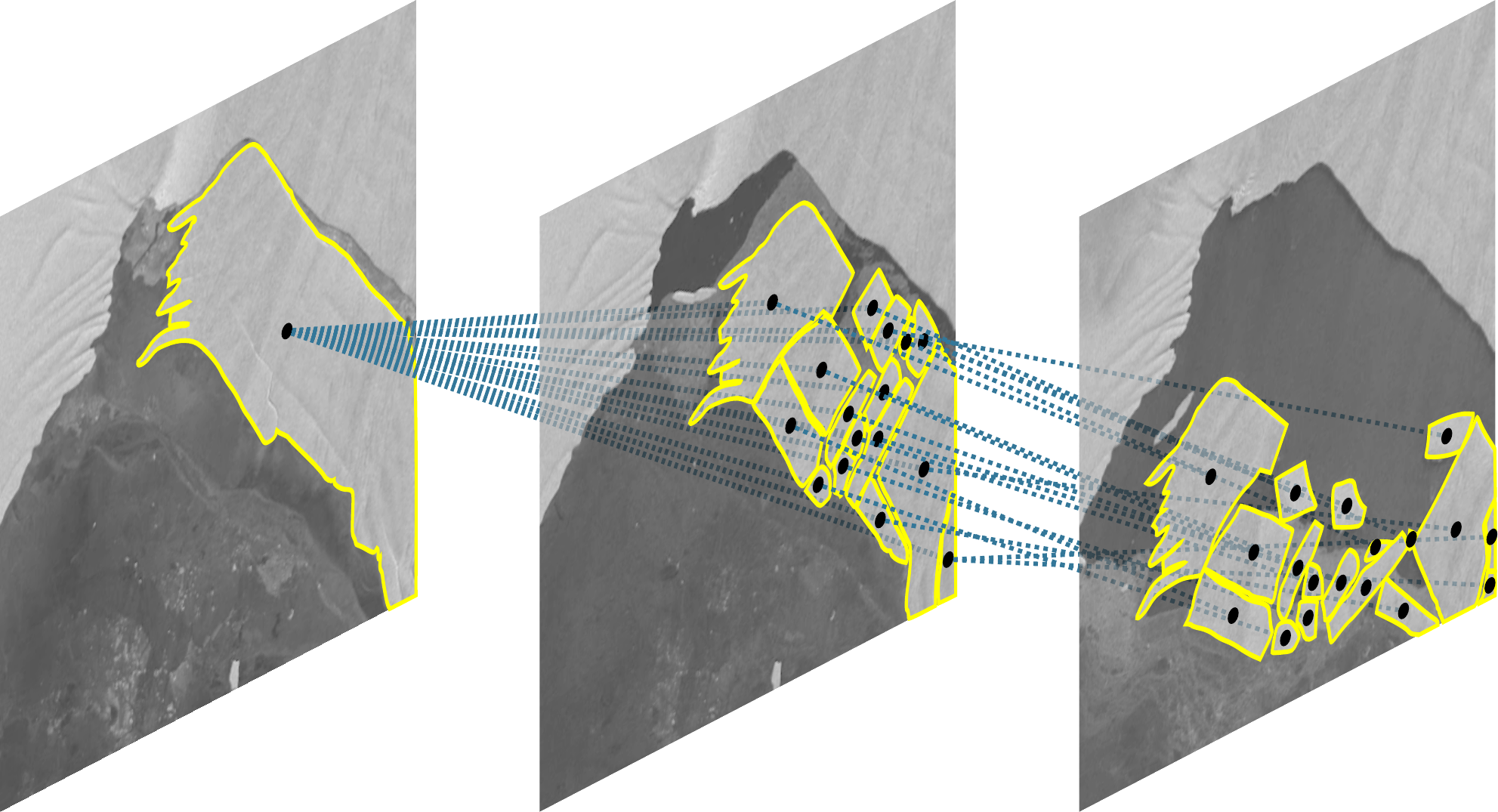}
		\label{fig:tempo_edges_def_manual}
	}
	\hfill
	\subfloat[Spatial overlap]{
		\includegraphics[clip,width=.3\linewidth]{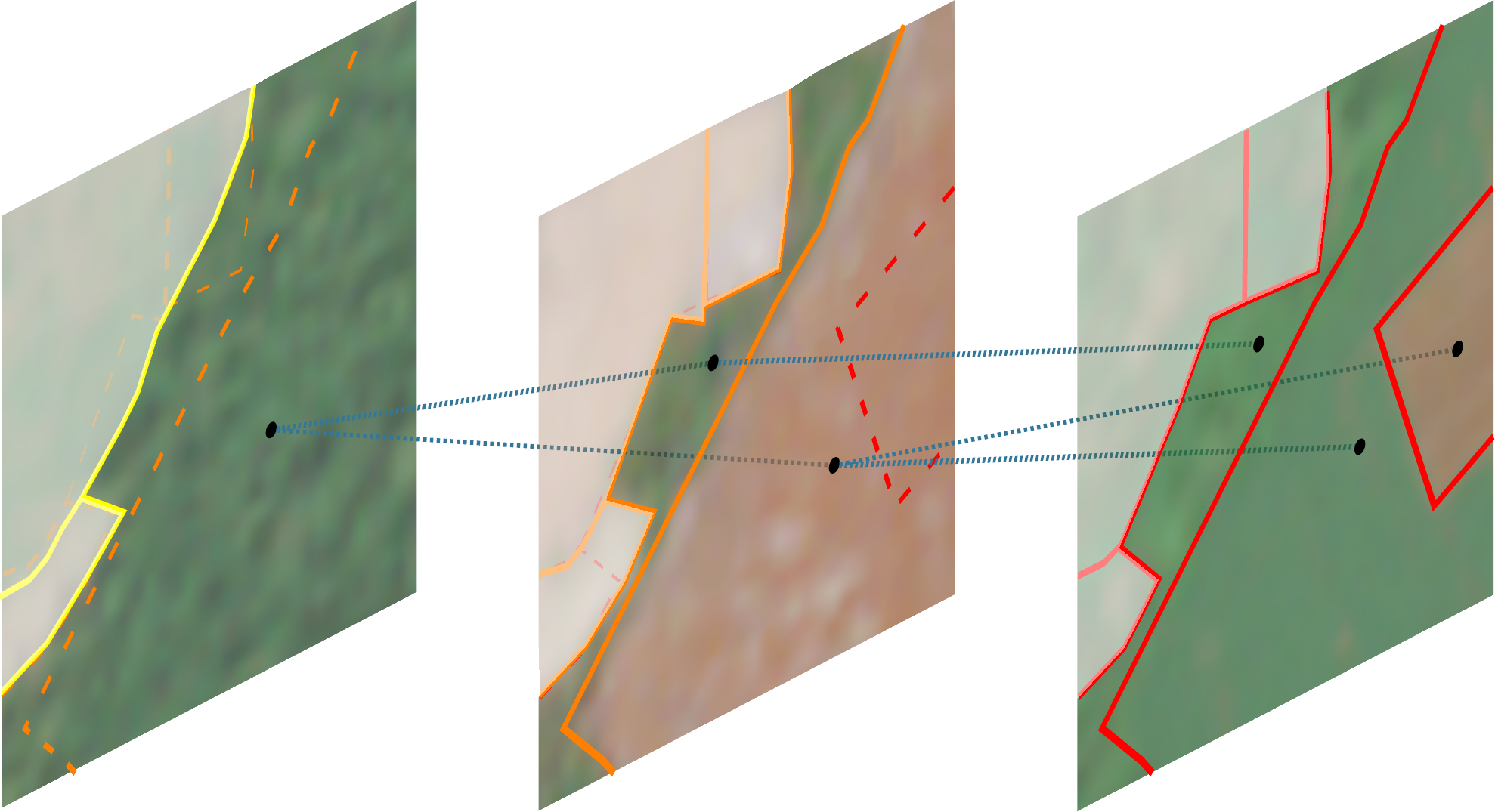}
		\label{fig:tempo_edges_def_overlap}
	}
	\\
	\vspace{1em}
	\subfloat[Feature similarity]{
		\includegraphics[clip,width=.3\linewidth]{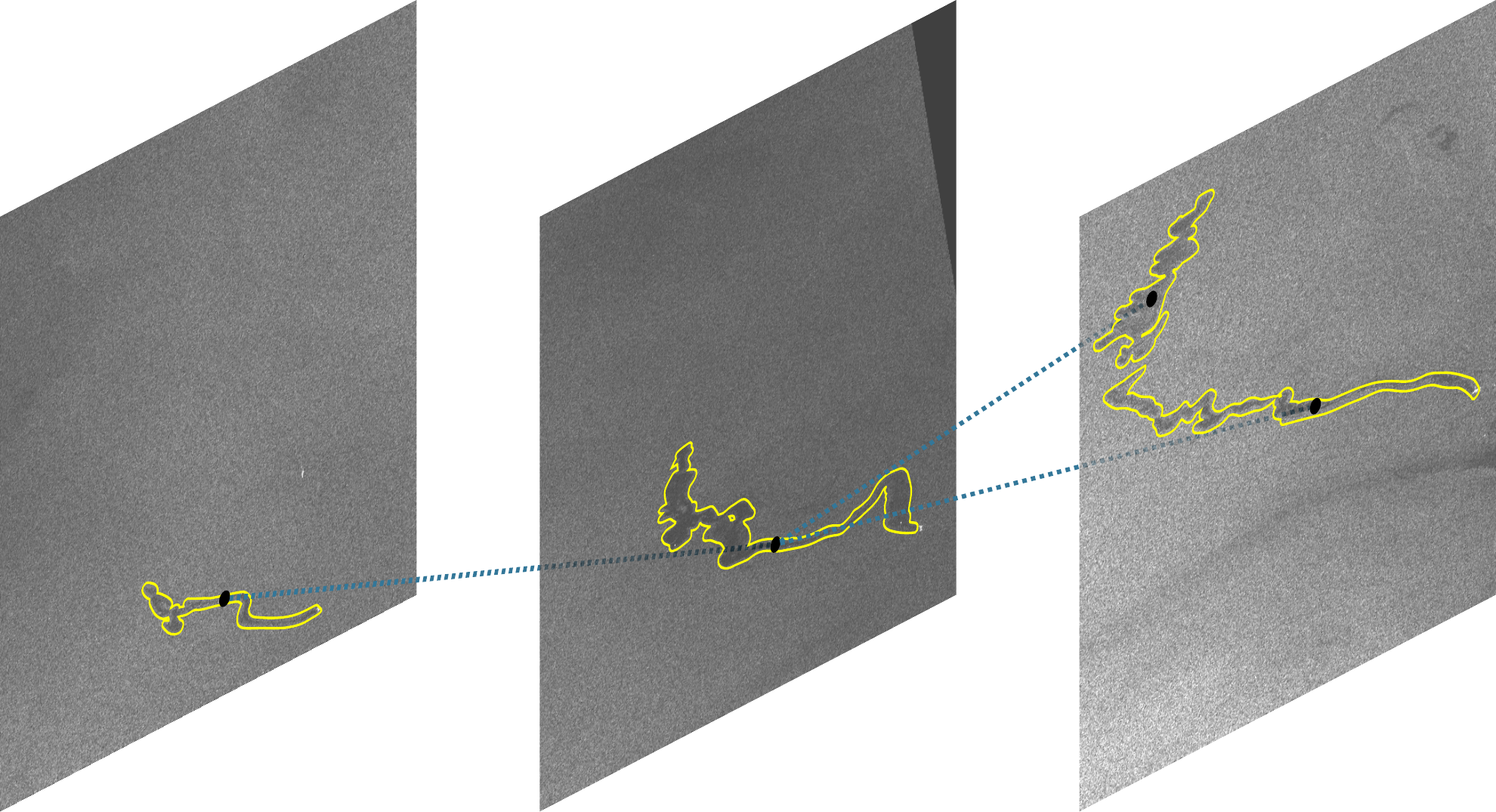}
		\label{fig:tempo_edges_def_similarity}
	}
	\hfill
	\subfloat[Periodic connections]{
		\includegraphics[clip,width=.5\linewidth]{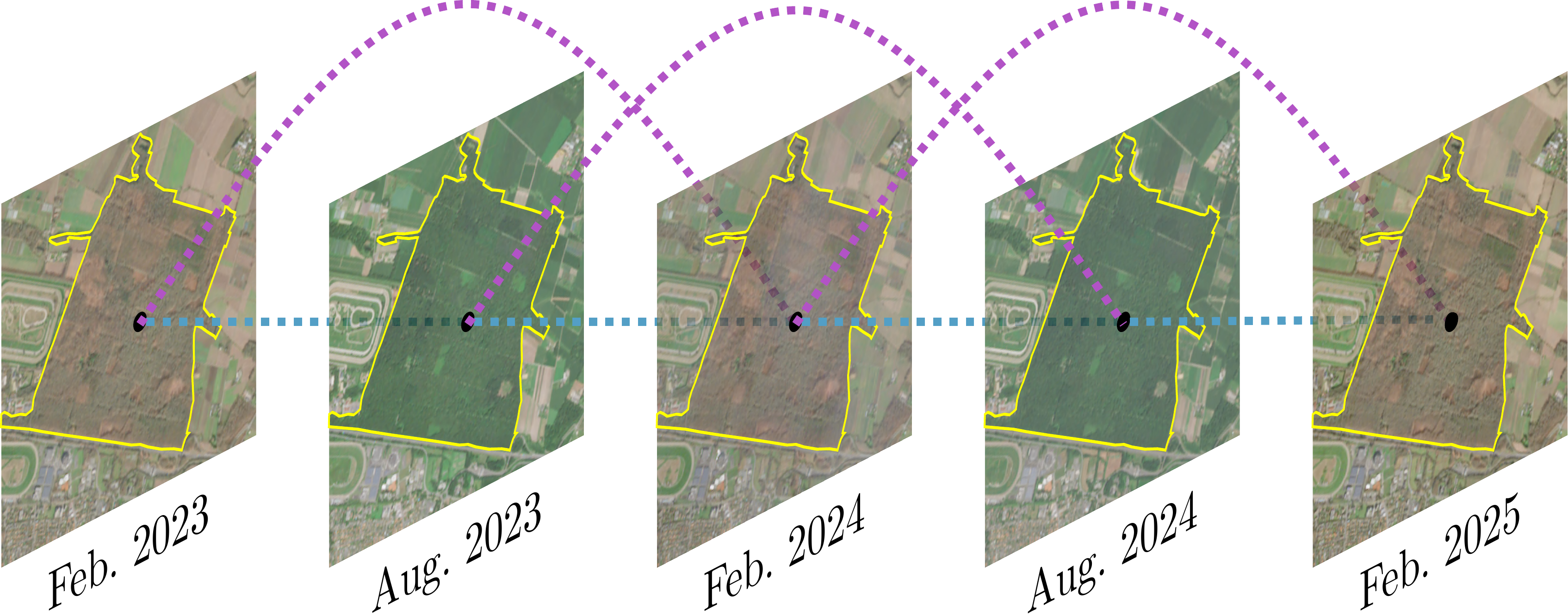}
		\label{fig:tempo_edges_def_periodic}
	}
	
	\caption{Different designs of spatio-temporal relationships in spatio-temporal graphs, illustrated by a few use cases. (a)~Object tracking by an expert or a domain-specific algorithm, here for ice shelf monitoring. The fragmentation and large displacement of objects motivate the \mbox{(semi-)manual} design of edges by thematic experts. (b)~Spatial overlap of objects from different timestamps, here for forest and crop monitoring. Slash-and-burn agriculture provides nutrients to the soil, which explains this design choice focused on spatial extent. (c)~Similarity of object features, here for oil spill monitoring. Oil slicks have a distinct color that contrasts with the ocean background, and their large displacement can be handled by a feature similarity criterion. (d)~Object periodicity, here to capture the phenology of a forest. In the Northern Hemisphere, temperate forests become greener in spring and summer, followed by browning in autumn and winter. Long-range spatio-temporal edges can handle these seasonal cycles.}
	\label{fig:tempo_edges_def}
\end{figure}

It can be easy to establish spatio-temporal connections when the identity of the objects is already known thanks to prior tracking, whether manual or automatic~\cite{martesen2017, delmondo2010, xue2019, yi2014}. Such a spatio-temporal edge design is plotted in~\Cref{fig:tempo_edges_def_manual} for ice shelf tracking. However, this type of link can be more difficult to make when objects are defined with few priors, as is often the case with SITS due to the heterogeneity of the scenes observed.

Most works consider spatio-temporal links between temporally successive objects, which could be considered adjacent in the geometric framework of 4D space-time, thus sharing a common boundary in both space and time. In other terms, objects are linked if their timestamps are consecutive and their spatial footprints overlap significantly~\cite{gueguen2006, cheung2015}. This is based on the assumption that the evolution of an entity is mainly linked to the past evolution of its spatial region, as in the case of slash-and-burn agriculture illustrated in~\Cref{fig:tempo_edges_def_overlap}, and the influence of neighboring regions. This approach can be used to represent changes in attributes between two entities, as well as elementary geographical changes such as mergers and divisions. However, this approach fails to track fast object displacement relative to the sampling frequency, as the spatial footprint of these objects does not necessarily overlap between two consecutive frames. It is important to take account acquisition granularity when modeling moving objects~\cite{hornsby2002}.

Spatio-temporal relationships based on feature similarity through time can handle both the evolution of regions and large entity displacement, as it is the case for oil spills plotted in~\Cref{fig:tempo_edges_def_similarity}.
By doing a clustering on the object feature space, objects belonging to the same cluster and localized on two consecutive dates are linked by a spatio-temporal edge~\cite{heas2004b,heas2005}. This definition of spatio-temporal relationship allows long-range interactions between spatially unaligned objects by assuming that objects with similar characteristics act in similar ways over time, and therefore that their information is worth sharing. This setting makes it possible to link geographically distant but spectrally similar areas, in order to study global dynamics or diffuse processes.

In addition to the edges between objects of consecutive dates, periodic connections with medium- and long-term dates make it possible to take into account noise and seasonal nuisance~\cite{wang2022}. Medium-term connections ensure that at least one neighbor can denoise a possibly cloudy acquisition, while long-term links connect two acquisitions from different years but from the same month. An example of these long-term links for forest seasonality due to phenology is shown in~\Cref{fig:tempo_edges_def_periodic}. Thanks to these spatio-temporal connections, subsequent analysis of the graph can distinguish long-term trends from abrupt changes. So as not to depend on \textit{a priori} information on cycle length, it is possible to learn spatio-temporal links between objects of different dates, both near and far~\cite{ahmad2023}. This should give a better understanding of repetitive temporal phenomena over long periods thanks to these long-distance dependencies without \textit{a priori} information.

These different approaches define which entities are temporally related to each other, but not to what extent. The importance of the connection can be established using a weighted graph by considering the adjacency rate of the regions~\cite{rejichi2011}, the similarity of their features~\cite{rejichi2014a}, or potentially by considering the temporal distance between the objects as it can be expressed with the relative temporal positional encoding~\cite{foumani2024improving}.

More generally, it is interesting to note that the temporal dimension is increasingly being exploited through the use of graphs, thanks to the recent advancements in GNNs. For example, the Transformer architecture can be seen as exploiting a complete graph over a time series, \ie, a graph in which each pair of timestamps is connected. Recent advances in graphs for time series~\cite{jin2024survey} could be derived and adapted to the case of SITS.

We can see similarities between spatial and spatio-temporal relationships. This is due to the way in which spatio-temporal objects are defined, depending on whether the temporal dimension is completely distinguished from the spatial dimensions, or whether they belong to the same space-time.

\subsection{Fidelity and complexity}
Despite the advantages of a graph-based approach, modeling SITS as an intermediate step necessarily induces additional complexity for data representation compared to direct data use. However, the processing of this representation can be less costly, leading to a compromise between the accuracy of the representation and its processing. It is therefore important to know the cost of such a representation in terms of model fidelity, storage and computational complexity. The theoretical complexities reported in this section are consistent with experiments, both in terms of runtime and memory usage. For space reasons, these experiments are not reported in the paper, but are available on request from the corresponding author.

\subsubsection{Model fidelity}
Fidelity refers to the model's ability to accurately represent reality. In SITS case, since the spatio-temporal graph acts as an intermediate representation of the data, the information must be preserved as much as possible. The first step in processing SITS is segmentation into objects, followed by the extraction of the embedding to represent these objects. Model fidelity therefore depends on both the SITS segmentation and the object embeddings.

An ideal embedding from a fidelity point of view should contain the same information---radiometric, textural, and geometric---as in its corresponding object. However, it is also crucial for an embedder to extract higher-order features that are useful for the downstream task. A good embedding must be a compromise between an embedding capable of reconstructing the region identically and a lossy embedding with generalization capabilities. For example, a mean aggregator is not a high-fidelity encoder as it loses a lot of information, but focuses on some part of the radiometric data that could be crucial for specific applications. Recently, learning methods for encoding objects have given good results (\Cref{sec:obj_feat}), but the level of fidelity of embeddings and their ability to generalize is more difficult to assess than for deterministic methods, as they depend on the training data.

The choice of an embedder, and consequently the level of complexity required to obtain accurate representations, is strongly linked to the object identification method. Obviously, a homogeneous region will be simpler to encode than a textured one, but the complexity is transferred to the inter-object level with a higher number of small entities and more complex relationships. The notion of fidelity also applies to the spatial footprint of objects. If the contours of an object do not correctly follow an observable entity, this can influence and limit the performance of an application, as shown in the use case of land cover mapping (\Cref{sec:gden_segm_error}). Therefore the fidelity of a spatio-temporal graph is measured by the fidelity of the segmentation used and the object embedder.

Beyond the encoding of acquired values, spatio-temporal graphs also contain spatial and spatio-temporal relationships which add higher-level information, improving the model's fidelity to reality. Despite a higher memory cost than pure OBIA approaches in which relationships between objects are not explicitly modeled, this additional information serves as a guide for the downstream task, increasing efficiency, as shown in~\cite{dufourg2023b}.

\subsubsection{Storage complexity}
The storage complexity for the raw SITS data is $O(C {\times} T {\times} H {\times} W)$ with $C$ the number of input channels, $T$ the number of acquisition dates, $H$ and $W$ the height and width of images in pixel.

Concerning pure OBIA models, objects attributes and the object-pixel map are stored, leading to a memory cost of $O(f_\mathcal{V} {\times} \left|\mathcal{V}\right|) + O(T {\times} H {\times} W)$, with $f_\mathcal{V}$ the object embedding dimension and $\mathcal{V}$ the set of objects. The number of objects is limited by the number of SITS pixels, \ie, $\left|\mathcal{V}\right| \leq T {\times} H {\times} W$. The higher memory cost comes from the object-pixel map which is needed in order to return from the object- to pixel-scale. It associates an object index with each SITS pixel.

Spatio-temporal graphs add spatial and spatio-temporal edges to extend the OBIA paradigm. Attributed edges $\mathcal{E}$ can be stored in memory respecting an edge list representation, $O(\left|\mathcal{E}\right|) + O(f_\mathcal{E} {\times} \left|\mathcal{E}\right|)$, $f_\mathcal{E}$ being the edge features dimension. In total, by considering the object features, the object-pixel map, the spatial and spatio-temporal edges and their features, the memory cost of a spatio-temporal graph is $O(f_\mathcal{V} {\times} \left|\mathcal{V}\right|) + O(T {\times} H {\times} W) + O(\left|\mathcal{E}\right|) + O(f_\mathcal{E} {\times} \left|\mathcal{E}\right|)$.
Depending on the task, the object-pixel map is not always necessary to be stored. In this case, the spatio-temporal graph has a lower storage complexity than the raw data. The data compression ratio can thus be approximated by $\displaystyle \frac{O(C {\times} T {\times} H {\times} W)}{O(f_\mathcal{V} {\times} \left|\mathcal{V}\right|) + O(T {\times} H {\times} W) + O(\left|\mathcal{E}\right|) + O(f_\mathcal{E} {\times} \left|\mathcal{E}\right|)}$ and depends on the number of nodes, edges, and the feature dimensions of the graph.

\subsubsection{Time complexity}
The use of an intermediate representation of the data necessarily adds a supplementary computational cost for its construction. The time complexity results from the cost of each individual step of the construction process, from object identification and characterization, to the addition of spatial and spatio-temporal edges. Each step can be achieved in numerous ways as seen previously. For object identification, a time complexity of $O(C {\times} T {\times} H {\times} W)$, $O(C {\times} T {\times} H {\times} W {\times} log\left(T {\times} H {\times} W\right))$ or $O(I {\times} C {\times} T {\times} H {\times} W {\times} |\mathcal{V}|)$ can be achieved with respectively SLIC, Felzenszwalb and  $k$-means algorithms, $I$ being the number of iterations of $k$-means. Object feature extractors can be relatively simple, such as the one for mean values which has a time complexity of $O(T {\times} H {\times} W)$, but also complex, such as with neural networks. Adding edges increases the time complexity differently depending on the criterion. If objects are connected by adjacency, the computational cost is $O(T {\times} H {\times} W)$. If edges are based on a similarity criterion, the time complexity is $O(f_\mathcal{V} {\times} \left|\mathcal{V}\right|^2)$ for a naive implementation calculating all pairwise distances, whereas it is $O(\left|\mathcal{V}\right| log \left|\mathcal{V}\right|)$ with an optimized approach based on a k-d tree or a ball tree. It should be noted that the time complexity is highly dependent on the design choices made for the spatio-temporal graph, and can therefore act as a decision factor.

This construction cost can be balanced by the usually much lower computational complexity to process graph values than raw SITS data for applications. It can be explained by the divide-and-conquer approach used in the OBIA paradigm. To illustrate this gap, the time complexity of the convolution operation on a SITS and a spatio-temporal graph can be compared. Convolutions are commonly used in computer vision to take into account the surrounding context of a pixel. The operation has been extended to the graph domain to achieve similar behavior on nodes and their neighbors. The convolution on a SITS with a kernel of size $K^T {\times} K^H {\times} K^W$ has a time complexity of $O(C^3 {\times} T {\times} H {\times} W {\times} K^T {\times} K^H {\times} K^W)$. The time complexity of a convolution on a graph is $O(f_\mathcal{V} {\times} |\mathcal{E}| + {f_\mathcal{V}}^2 {\times} |\mathcal{V}|)$ (based on GCN computations~\cite{kipf2017gcn}). Time savings are thus linked to the number of nodes and edges in the graph. It is therefore important from a computational point of view to choose wisely which relations to model, as their quantity influences computational time.

\section{From graph to task}
\label{sec:graphAppli}

Due to physical laws, all phenomena on Earth evolve in a similar way if conditions and causes are the same. Consequently, to understand these phenomena, we need to identify and exploit the spatio-temporal patterns to which they are linked, and spatio-temporal graphs are an effective tool for this purpose. They are adapted both to human understanding and to machine calculations and modeling. This section will therefore look at different ways of using spatio-temporal graphs to understand and exploit their patterns, from least to most automated, in the context of Earth observations.

\subsection{Expert graph analysis}
\label{sec:task_expert_analysis}

\begin{figure}[!b]
    \centering
    \includegraphics[width=.4\linewidth]{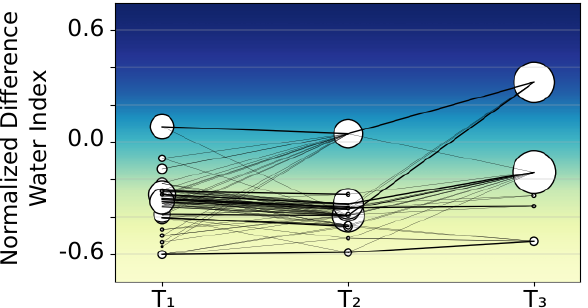}
    \caption{Graph as a visualization tool for experts. The plot illustrates the normalized difference water index (NDWI) temporal profile of the entities, in the same way as in~\cite{guttler2017}, and suggests a soil covering event by water. The size of a node is proportional to the area of the corresponding region, and the weight of an edge denotes the alignment between the two linked regions.}
    \label{fig:task_visu}
\end{figure}

First, graphs can serve as a visualization tool for human interpretation of spatial and temporal dynamics. For example, the dynamic processes and scenarios of ocean eddies are efficiently stored in the spatio-temporal and spatial connections of graphs in~\cite{yi2014}. As a result, the graphical representation of a specific eddy can highlight the dynamic changes in its lifetime and the interactions with all the other eddies, to help better understand their development. This representation is also used to illustrate the movement trajectory of the foreshore during its life cycle~\cite{xu2021}. Another visualization practice, presented in~\cite{guttler2014exploring,guttler2017} and illustrated in~\Cref{fig:task_visu}, uses spatio-temporal graphs to monitor the evolution of agricultural parcels and natural areas. This tool is also adapted to Synthetic Aperture Radar (SAR) data for flood analysis~\cite{debusscher2019object}. The visual interpretation of the graph structure, \ie, the merging and splitting of objects and the temporal profile of node attributes makes it possible to identify seasonal phenomena such as leaf fall, agricultural harvests, and dry summer conditions.

This type of analysis uses graphs to track the evolution of a given object. Although this evolution can be studied visually, the computation of some statistical indicators on the graphs can help in this task. The simplest, counting the number of nodes and edges per graph, can give a first idea of the dynamics under study, but it is possible to go further. For example, the evolution of a specific area over a given period can be analyzed by calculating a variation score on the attributes of the objects~\cite{guttler2017,debusscher2019object}. It is also possible to compute several spatial coverage indicators to understand the spreading or shrinking of an area~\cite{guttler2017}.

Finally, some basic graph operators assist in capturing some predefined patterns from which further analyses can be conducted. For instance, structural operators such as node in- and out-degrees are used in~\cite{xu2021,li2021, o2025persistence} to identify the appearance, disappearance, splitting, merging, and development of entities. Morphological changes in entities can also be detected by comparing the attributes of successive objects, as in~\cite{xu2021} with the area attribute.

These simple tools greatly help experts to understand the dynamics of a study area in relation to global phenomena and specific changes. Whether using pattern statistics to detect exceptional periods, occurrence maps to identify particular locations, or other indicators that can be linked to external data and visual representations, experts can summarize knowledge and deduce rules for apprehending the phenomena under study in the short or long term, and locally or globally, depending on the techniques used.

\subsection{Graph pattern mining}
\label{sec:task_graph_pattern_mining}

A global understanding of phenomena occurring in SITS is provided by analysing its corresponding spatio-temporal graph. However, understanding and detecting local dynamics is still limited by human analysis. Mostly basic phenomena following simple pre-established rules can be studied. Although fundamental, these simple rules also form the basis of more complex models that are important to study. To this end, graph-based pattern mining involves identifying frequent patterns, or retrieving patterns that are similar to a user-specified phenomenon type.

Pattern extraction on spatio-temporal graphs is based on graph mining algorithms. Although subgraph mining methods exist, known works on spatio-temporal graphs have focused on sequential patterns to emphasize the temporal evolution of attributes. To this purpose, \citet{lemen2009} tailored a frequent sequence mining algorithm~\cite{zaki2000eclat} for spatio-temporal graphs. This approach uses a depth-first traversal of the prefix tree to search for frequent patterns. Another algorithm~\cite{sanhes2013weighted} based on a similar principle, \ie, extending paths \textit{via} a depth-first search of the prefix tree, is used in~\cite{collin2016}. However, instead of simple paths, this one exploits weighted paths as condensed patterns to better describe the evolution of a region, by specifying the frequency and compactness of the patterns. Extracting all the frequent patterns in a spatio-temporal graph gives end-users a global view of the major phenomena occurring in a study area, while enabling them to carry out a geographically and temporally targeted study of each phenomenon, such as done in~\cite{lemen2009} which provides a theoretical interpretation of the frequent NDVI evolution patterns returned.

When only a particular phenomenon is being studied, it is advisable to retrieve only this specific phenomenon, and not all the frequent ones. In this case, the first step is the formalization of the relevant pattern by the experts, then an automatic detection of similar subgraphs is applied to the whole spatio-temporal graph to localize such phenomenon both spatially and temporally. The formalization of a phenomenon by experts is not an easy task and various methodologies exist in the literature. Basically, the phenomenon can be summarized as a set of rules that can be interpreted in a SITS-based graph context. These can be used to enumerate the subgraphs that respect the rules, as in~\cite{mougel2013}, or to reformulate the similar phenomena detection task as a constraint satisfaction problem, as in~\cite{ayadi2023}. Another strategy is to synthesize a given phenomenon through a reference graph. Based on a remote-sensing scene ontology, \citeauthor{rejichi2015c}~\cite{rejichi2014b, rejichi2015c} identify concepts, such as water surfaces, forests, or crops, and formalize them with attributes based on ground truth data. The reference graph of a phenomenon is then created by the user based on those formalized concepts temporally linked. In a more interactive scenario, it is possible to incrementally learn the definition of the reference spatio-temporal pattern via user-provided positive and negative subgraph examples~\cite{heas2004a, heas2004b, heas2005}. The detection of similar subgraphs in the whole spatio-temporal graph can also be solved in different ways. Given a collection of subgraphs, the retrieval can be based on the similarity between each element of this collection and the reference graph, as in~\cite{heas2004a, heas2004b, heas2005, rejichi2014b, rejichi2015c}. The similarity between graphs can be computed using either a parametric distance model~\cite{heas2004a}, the graph edit distance~\cite{neuhaus2007bridging} or a graph kernel method~\cite{kashima2003}.

These works show how graph mining can be used to help expert analysis of SITS by efficiently retrieving phenomena of interest in the spatio-temporal graph structure. However, graph analysis remains manual, requiring an expert to extract the most useful information for a given case study.

\subsection{Extrinsic prediction from graph}
\label{sec:task_classification}

Although relevant phenomena can be identified through graph mining, much of the analysis work remains manual, whether it involves defining interesting patterns or interpreting extracted patterns. An interesting possibility is to extract the desired information directly from the spatio-temporal graph, \eg, a land cover map over a period or at each date, or even a map of the different change type processes. These maps are the result of a classification task at the level of either the local nodes or the entire graph. The task is known as extrinsic prediction, \ie, predicting variables that are external to the SITS-system being studied. To the best of our knowledge, there is limited work on extrinsic regression for SITS involving graphs~\cite{qiao2023kstage}, so this section focuses solely on classification tasks. Nevertheless, most of the following methods can be easily adapted to extrinsic regression tasks, enabling the estimation of biophysical parameters characterizing SITS, \eg, phenological parameters or yield estimation in agricultural areas, or the concentration of pollutants. It could be done, for example, by modifying the loss function in the deep neural networks. Besides, these predictions can be done with limited supervision to distinguish different phenomena and behaviors. It is also possible to learn from labeled examples to integrate semantics into the results.

\subsubsection{Unsupervised graph partitioning and clustering}

Similar behaviors can be identified at two levels for spatio-temporal graphs, between nodes and between graphs. The former aims to identify spatio-temporal sub-structures in SITS. To achieve that, the corresponding graph can be split into different parts by focusing on the dissimilarities between them, as done in~\cite{tarabalka2013}, which uses GraphCut on a spatio-temporal pixel graph to detect burned areas over time. Another option is to identify sub-structures by clustering nodes with similar attributes, with or without neighborhood constraints. For instance, a Felzenszwalb-based method can detect urban changes on a graph of images~\cite{wang2022}, or Infomap, a community detection algorithm, can distinguish various agricultural landscapes~\cite{interdonato2020}. Markov Random Fields can also be exploited to cluster spatio-temporal subgraphs according to their evolution~\cite{gueguen2006, radoi2015spatio, sziranyi2019fusion}.

The previous works aim to identify similar behaviors inside the same graph. However, it is also interesting to cluster spatio-temporal graphs in order to compare the dynamics of regions.
The works~\cite{khiali2018, khiali2019, kalinicheva2020} cluster evolution graphs with the aim to characterize common behavior of different phenomena and thus analyze the dynamics of the study areas over time. The clustering of such evolution graphs is based on the distance between summarized representations of each graph, called synopsis. Such clustering can be applied to raw synopses with classical Euclidean distances~\cite{khiali2018} or distance metrics tailored to sequences, such as dynamic time warping~\cite{khiali2019}. It can also be applied to synopsis embeddings, obtained, for example, from a GRU autoencoder~\cite{kalinicheva2020}.

\subsubsection{Supervised methods}
\label{sec:supervised_methods}
The graph partitioning and clustering methods can group spatio-temporal structures that share similar behaviors, but they cannot use prior knowledge to label identified patterns and give them semantic meaning. The use of a database of already known phenomena can overcome this limitation by using supervised classification techniques to learn from it.
In this way, the knowledge induced by previously seen examples is used to recognize the learned classes, \ie, spatio-temporal structures, in new data. The spatio-temporal nature of the data in a SITS region classification task can be handled either directly by the classifier or by a post-processing step. In the latter case, a probabilistic graphical model, such as Markov Random Field, can be used to enforce the spatio-temporal consistency of the output~\cite{melgani2003markov, hagensieker2017tropical}. In the former case, the classifier itself should be able to process graph data. An example of such a model is presented in \cite{hoberg2014conditional}, which uses a Conditional Random Field, originally designed to handle graphical data. Traditional classifiers dedicated to tabular data analysis, such as Support Vector Machines (SVM), have also been extended to process graph data. For example, an SVM can employ specific kernel functions, such as a graph-adaptive kernel based on random walks~\cite{kashima2003}, to encode graph information. The use of the kernel trick in conjunction with the training of an SVM enables leveraging evolution graphs such as those in~\cite{lemen2009}. In this way, SVM can classify regions according to their geometrical evolution~\cite{rejichi2011, rejichi2012}, spectral and textural changes~\cite{rejichi2014a}, or a mixture of both~\cite{rejichi2015a, rejichi2015b}. However, SVMs are not suitable for large datasets for scalability reasons, and cannot handle the variance and noise that occur with SITS.

\begin{figure}[!b]
    \centering
    \includegraphics[width=.65\linewidth]{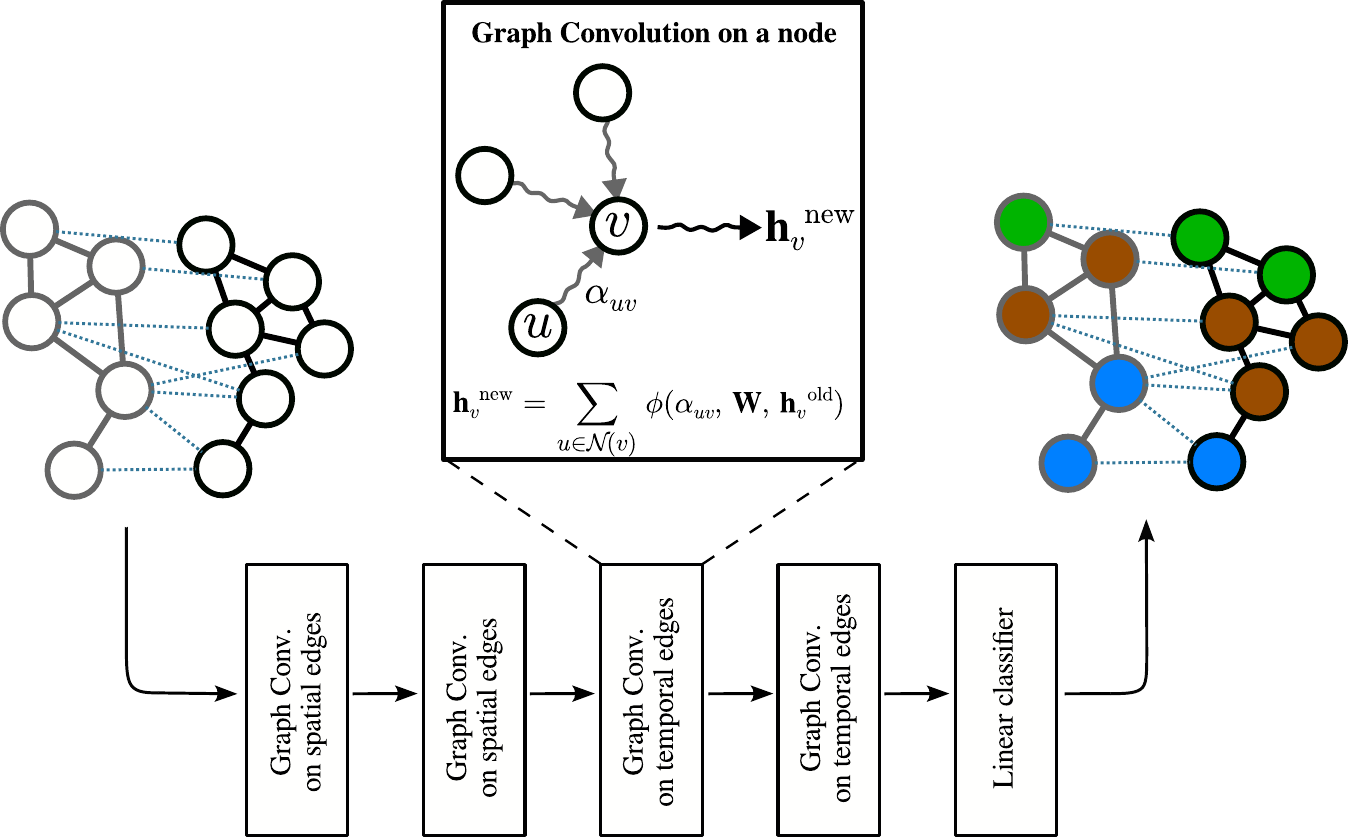}
    \caption{A graph neural network for node classification of spatio-temporal graphs. The spatial dimension is dealt with first by considering only spatial edges, and then the temporal dimension using only spatio-temporal edges. The message-passing principle that occurs in each graph convolutional layer is depicted.}
    \label{fig:task_gnn}
\end{figure}

Following the trend in computer vision and machine learning, recent works have shown that neural networks are well suited to process large quantities of data. Spatio-temporal dimensions of SITS have been successfully handled by recurrent cells and convolutions~\cite{russwurm2018, ji2018, interdonato2019} and attention-based architectures~\cite{garnot2020}. However, those methods work with the entire datacube and do not exploit geographical entities and their relationships, which is one of the reasons for using graphs to process SITS data. Unfortunately, Convolutional Neural Networks (CNNs), Recurrent Neural Networks (RNNs) and Transformers can only process grid and sequence data. To this end, Graph Neural Networks (GNNs) were developed to perform operations on irregular data such as graphs. They are mainly based on the principle of message-passing, which processes each node (referred as target node) and updates its state using information coming from its neighbors (referred as source nodes), as shown in~\Cref{fig:task_gnn}. This process is similar to the localised image processing found in convolutional models applied to grid data. We refer the reader to the works of~\cite{wu2021gnn} for a more in-depth review of GNNs and machine learning on graphs and to~\cite{zhao2025beyond} for Earth observation applications.

Spatio-temporal patterns from graphs can be learned with GNNs to exploit the spatial and temporal dimensions~\cite{zeghina2024deep, corradini2024systematic}. Those networks are of interest when the data structure is non-Euclidean. The discretization of SITS into objects has led to the loss of its Euclidean aspect, and thus to the growing interest in GNNs for processing them. 

Depending on the data, it may be more interesting to deal first with the spatial dimension, then the temporal one, or vice versa, or even to process them together.
Patterns over spatial dimension can be exploited by applying message-passing through spatial edges. Such spatial patterns can appear directly from the raw data, depending on the phenomenon under study, and so be extracted as in~\cite{nazir2023} and~\cite{dufourg2023b} by GNNs before analyzing the temporal dynamics of these patterns. Spatial patterns can also appear after processing the temporal dimension, and GNNs therefore propagate the temporal characteristics of objects, such as temporal profiles~\cite{zangari2021, tulczyjew2022} or deep features~\cite{censi2021}, through the graph's spatial edges to produce a new representation of each node taking into account both spatial and temporal information. Similarly, the use of spatio-temporal edges as a means of passing information offers the possibility of capturing temporal patterns, as in \cite{kavran2023, dufourg2023b}. We note that GNN layers can exploit both spatial and temporal dimensions and thus make the most of the flexibility of spatio-temporal graphs~\cite{dufourg2023b}. It should be noted, however, that \citet{dufourg2023b} treated the spatial and temporal dimensions sequentially, using the GNN plotted in~\Cref{fig:task_gnn}, to extract a spatio-temporal representation of the SITS. To handle both dimensions jointly, a solution can be to use a GNN that does not separate spatial and spatio-temporal edges~\cite{ahmad2023}. GNN layers adapted to heterogeneous graphs could also be a way of learning a joint representation.

Deep learning requires a large amount of labeled data to be effective. However, labeling remote-sensing images is costly and time-consuming. Semi-supervised methods make it possible to limit the amount of data to be annotated while preserving the semantics of the output, unlike unsupervised approaches. For instance, in~\cite{censi2021}, the reference data is a sparse collection of polygons assigned with a land-use class. As the method used is based on neural networks, more data help to provide better embedding and thus unlabeled data are still used for spatial propagation in the graph, together with labeled data. Instead of an embedding-based classifier, \citet{fei2023} proposed a Poisson learning approach based on label propagation solving the Poisson equation. This method builds a weighted spatio-temporal graph that acts as a regularizer to learn how to spread labels over unlabeled nodes. Here, edge weight is a similarity measure based on the spatial, temporal and spectral characteristics of the nodes. The latter approach has been very effective in the case of an extremely limited number of labels, however it is restricted to the transductive setting (see \Cref{sec:background:task}) and cannot induce new results on blank data, unlike embedding-based approaches.

\subsection{Intrinsic regression using graph}
\label{sec:task_forecasting}

The above-mentioned works show the range of tools available to extract new information from SITS using graphs. However, classification and regression of extrinsic data do not provide an internal understanding of a spatio-temporal phenomenon. For this kind of problem, intrinsic regression, such as data imputation and forecasting, is more appropriate. They involve the prediction of variables belonging to the same space as the input data in order to predict missing data or to forecast the future of such variables.

To illustrate the benefits of graphs for these tasks, the works cited in this section cover those using SITS-derived data. These data may include spatio-temporal reanalysis of the atmospheric state, landslide displacement based on sequences of point clouds obtained by radar interferometry, or wildfire evolution maps. Those SITS-derived data face the same challenges than SITS regarding the processing of spatial and temporal dimensions~\cite{zeghina2024deep}. Thus, in order to illustrate the value of graphs for intrinsic regression, despite the paucity of work with raw SITS, we are relying in addition on these SITS-derived data.

\subsubsection{Imputation}
\begin{figure}[h]
    \centering
    \includegraphics[width=.9\linewidth]{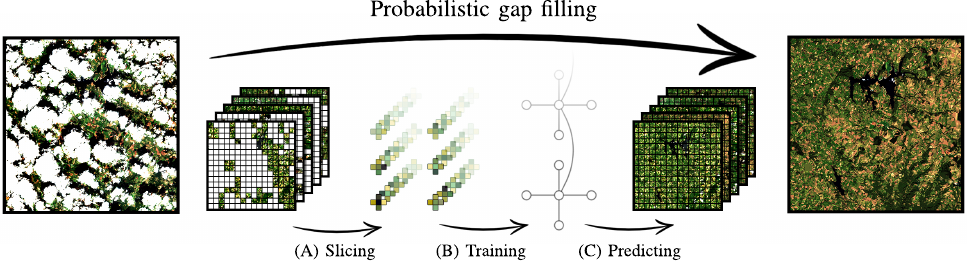}
    \caption{Probabilistic gap filling in cloudy SITS~\cite{fischer2020} (©2020 IEEE)}
    \label{fig:task_gap_filling}
\end{figure}

Firstly, a common intrinsic regression task in remote sensing is to fill gaps in incomplete data due to clouds and their shadows. Probabilistic approaches such as Markov random fields using a spatio-temporal graph can be used for gap filling~\cite{fischer2020} (see~\Cref{fig:task_gap_filling}). Another task that can be seen as intrinsic regression is super-resolution. To improve the spatial resolution of the prediction, learning models can be designed to train on a coarse mesh before inferring on a higher-resolution mesh, as it is done for the prediction of sea surface temperature in~\cite{lienen2022}. Hence, graph processing can be used to infer new data by inducing a structural bias in known spatial and temporal space.

\subsubsection{Forecasting}
The extension for prediction on a future time domain with knowledge of the past relates to forecasting. Current work focuses on predicting the future state of a graph as a prediction of the future node features. Spatio-temporal data modeling thus takes the form of either a static graph, with its nodes carrying time series, or a discrete-time dynamic graph, \ie, a sequence of snapshots sampled from a dynamic graph. With both representations, the spatial structure can be modeled in several ways, as discussed in \Cref{sec:spatialRelations}: adjacency relationships~\cite{peng2023}, distance-based relationships~\cite{zhang2021sea}, neighborhood relationships~\cite{gururani2024developing}, mesh-based structures~\cite{lienen2022}, similarity links~\cite{ning2024harnessing}, adaptive learned edges~\cite{cachay2021world}. The spatial dimension of the data is thus captured using GNNs and the temporal dimension can be managed with other embedding methods frequently encountered with time series, such as CNNs, RNNs or Transformers~\cite{lim2021time, mohammadi2024deep}.

Some works explore the forecasting task as a simple regression task with a sliding temporal window without explicitly exploiting the time information or temporal acquisition order, \eg, an encoding of the acquisition date. In this context, the spatial patterns existing between the time series, carried by the nodes, are exploited with GNNs to predict the future value of each node~\cite{cachay2020,cachay2021world,ning2023graph,ning2024harnessing}. However, the authors also pointed out possible extensions with more complex architectures specific to the spatio-temporal dimension of the data.

To design such models, the spatial and temporal dimensions can be treated sequentially, one taking precedence over the other depending on the phenomenon under study, without there being any consensus---this issue having already been mentioned for the SITS classification architectures in~\Cref{sec:supervised_methods}. If the spatial dimension is processed first, a graph neural network independently encodes the graphs for each acquisition date. The embedding time series thus obtained for each node can then be processed by an LSTM~\cite{liang2023graph}, GRU~\cite{rosch2024data, zhou2021, gururani2024developing, zhao2024adaptive, ge2024non}, temporal convolutions~\cite{ye2023graph}, or Transformer~\cite{wang2024dynamic, liang2024adaptive} to encode the temporal dynamics, which are then sent to a decoder to make a forecast. In order to better manage multi-characteristic phenomena, some works use several modules to distinguish between different time scales~\cite{peng2023enhanced}, or to separately learn periodic evolutions from historical ones and local phenomena~\cite{ye2023graph}. If the temporal pattern is given priority, a GNN for the spatial dimension is applied to a static graph whose nodes' temporal pattern has already been embedded, which can be achieved by random forest~\cite{ou20243d} or by temporal convolutions~\cite{zhang2021sea, gao2023global, sheng2023graph}, for example.

Successive learning in the temporal and then spatial dimension, or vice versa, to extract a representation from the SITS limits the complementary and inseparable aspect of these two dimensions. \citet{peng2023enhanced} have sought to overcome this limitation by integrating graph convolution on the spatial dimension within GRU cells, enabling the convolution to exploit the hidden state containing temporal information. However, although spatial and temporal learning are linked, they remain partly sequential. Joint embedding can be achieved using a GNN on both spatial and spatio-temporal edges simultaneously, with graph nodes corresponding to local observations at single dates~\cite{mu2022enso, peng2023, kuang2022}.

Following a different logic, some works took inspiration from current physical simulation methods to predict the next state of the observed system. Machine learning is playing an increasingly important role in this field, and \citet{keisler2022} followed the work proposed by~\citet{pfaff2021learning} using an Encode-Process-Decode architecture to predict global weather. Analogous to finite element methods, this architecture consists of moving from a large native data space (physical data on the acquisition grid) to a reduced intermediate space (abstract feature data on a mesh) to learn the dynamics of the physical system more simply. GNNs are used here to project one space onto another, and also to approximate at the mesh level the differential operators governing the internal dynamics of most physical systems. Various extensions of this work exist, the best known being GraphCast~\cite{lam2023learning}, an ML-based model for medium-range global weather forecasting that outperforms numerical weather prediction methods. Such performance has encouraged its use for other applications, such as fire prediction~\cite{michail2025firecastnet}, and has led to methodological development. In fact, better learning of system dynamics is achieved by integrating multi-scale information via long-distance relationships~\cite{lam2023learning,lang2024aifs} or using a hierarchical mesh~\cite{oskarsson2023, oskarsson2024probabilistic, holmberg2024regional}. In order to match reality as closely as possible, \citet{alexe2024graphdop} train an encoder-processor-decoder architecture using directly observation data, with no physics-based reanalysis inputs or feedback. A more realistic rendering of forecasts is also possible using generative diffusion models~\cite{price2024probabilistic}. Previous work has focused on global weather forecasting applying a mesh approximating the sphere geometry to low-spatial resolution images. For local applications, regular quadrilateral~\cite{oskarsson2023, holmberg2024regional} or triangular meshes~\cite{flora2024wofscast} can be used to fit the rectangular shape of the images or, in the case of higher-resolution SITS, an irregular mesh to bring more detail to a specific area~\cite{dufourg2024forecasting}. Some works on local regions still need the information of surrounding areas, such as atmospheric and ocean conditions, and thus exploit boundary forcing~\cite{oskarsson2023, holmberg2024regional} or a global stretched-grid to maintain global dynamics while focusing on a specific region~\cite{nipen2024regional}.

One of the needs of forecasting methods, linked among other things to their field of application, is to understand how they work and the predictions they make. This need echoes the notions of interpretability and explainability of an automatic prediction system. While these notions are not specific to any task, the following paragraphs focus mainly on describing their use in the context of SITS-based graph forecasting as, to the best of our knowledge, no work was developed using SITS-based graphs for other tasks. Interpretability and explainability are closely related terms, whose definitions are not yet well established in the field of machine learning, as they may or may not be interchangeable depending on the authors~\cite{linardatos2020explainable}. In this section, in accordance with~\cite{gilpin2018explaining}, interpretability is the extent to present in understandable terms to a human the cause of a decision and it is linked with the cause-and-effect relationships within the model’s inputs and outputs. Explainability concerns the ability to discern the internal logic and mechanisms within a machine learning system. It should be noted that both interpretability and explainability are continuous notions, and that a model may only be interpretable, or explainable, to a certain extent.

Architectural inspiration from physical simulation methods is a first step towards making forecasting methods explainable and physically realistic. The architecture of a network can also be directly inspired by physical processes by dividing it into specialized sub-parts, as \citet{chen2024decomposing} have done by breaking down atmospheric dynamics into advection and convection with a GNN and a multi-layer perceptron. Physical priors can also be integrated into GNNs by incorporating assumptions about the shape of the unknown differential equation, which induces a structural bias towards learning specific processes~\cite{lienen2022}. In addition, graphs are not limited to modeling data geometry, they can also model relationships between variables. This makes it possible, for example, to integrate physical knowledge between several meteorological variables by using them as couplers~\cite{chen2024b}, or, on the contrary, to reinforce and analyze the causal or physical relationships that have been learned between these variables~\cite{mu2021enso, zhao2024causal}.

Another advantage of graphs is their interpretability, thanks to the explicit relationships between objects. Even with the use of GNNs, there are several methods to interpret predictions, such as community detection, feature importance and node importance. Depending on the application domain, the interpretability of a result can be crucial, and work has already been carried out in forest fire prevention~\cite{chen2024a} and weather forecasting~\cite{jeon2024a, jeon2024explainable}.

\vspace{1em}
With the various works and tasks mentioned above, spatio-temporal graphs are proved suited to analyze, understand and exploit spatio-temporal patterns appearing in Earth observation. The flexibility of graphs means they can be adapted to extract useful information for each task and application.

\section{Case studies}
\label{sec:exp}

So far, the design and exploitation of spatio-temporal graphs for SITS has been presented from a general point of view. In order to give all the keys to adapt the proposed pipeline to a specific application, Earth observation problems are studied here through two case studies. The first, inspired by the work of~\cite{dufourg2023b}, aims to provide accurate land cover mapping over time. Frequent land cover updates are crucial for accurate monitoring of landscape changes. The second, derived from~\cite{dufourg2024forecasting}, focuses on the prediction of water resources. These two case studies are complementary both in their task ---classification and forecasting---and in their choice of design. Although these two tasks are far from representing the full range of possible applications, they do illustrate the key questions that must be considered when applying SITS-based spatio-temporal graphs. The source code for these case studies is available at \url{github.com/corentin-dfg/graph4sits/}.

\subsection{Precise and dynamic land cover mapping}
\label{sec:gden}

Land cover is recognized in the United Nations framework as one of the 14 global fundamental geospatial data themes~\cite{unggim2019} as it influences sustainable development, climate change, biodiversity conservation, food security and disaster risk reduction. Frequent updates of precise land cover mapping is therefore crucial for land management policies.

\subsubsection{Data}
Land cover maps can be derived from two data sources: field surveys and remote sensing acquisitions. Satellite images are more suited for regular updates and global monitoring thanks to the high frequency and high coverage of satellite observations. However, a high number of acquisitions requires an automatic process for classification of land cover classes. The current state of the art is based on learning methods, and supervised classification is used to integrate prior knowledge of the classes~\cite{miller2024deep}. In this context, we propose leveraging graph-based learning strategies to SITS using a labelled dataset.

To boost the development of automatic methods for various applications on SITS, we introduce a dataset collection~\cite{dufourg2023a}, containing a list of downloadable publicly-available satellite imagery datasets with a temporal dimension, mainly annotated SITS and satellite videos. Although many datasets exist for SITS classification, most of them only provide land cover at a given date as ground truth. In contrast, DynamicEarthNet~\cite{toker2022dynamicearthnet} provides pixel-wise annotations at multiple timestamps on high-resolution acquisitions.

\begin{figure}
    \centering
    \includegraphics[width=.5\linewidth]{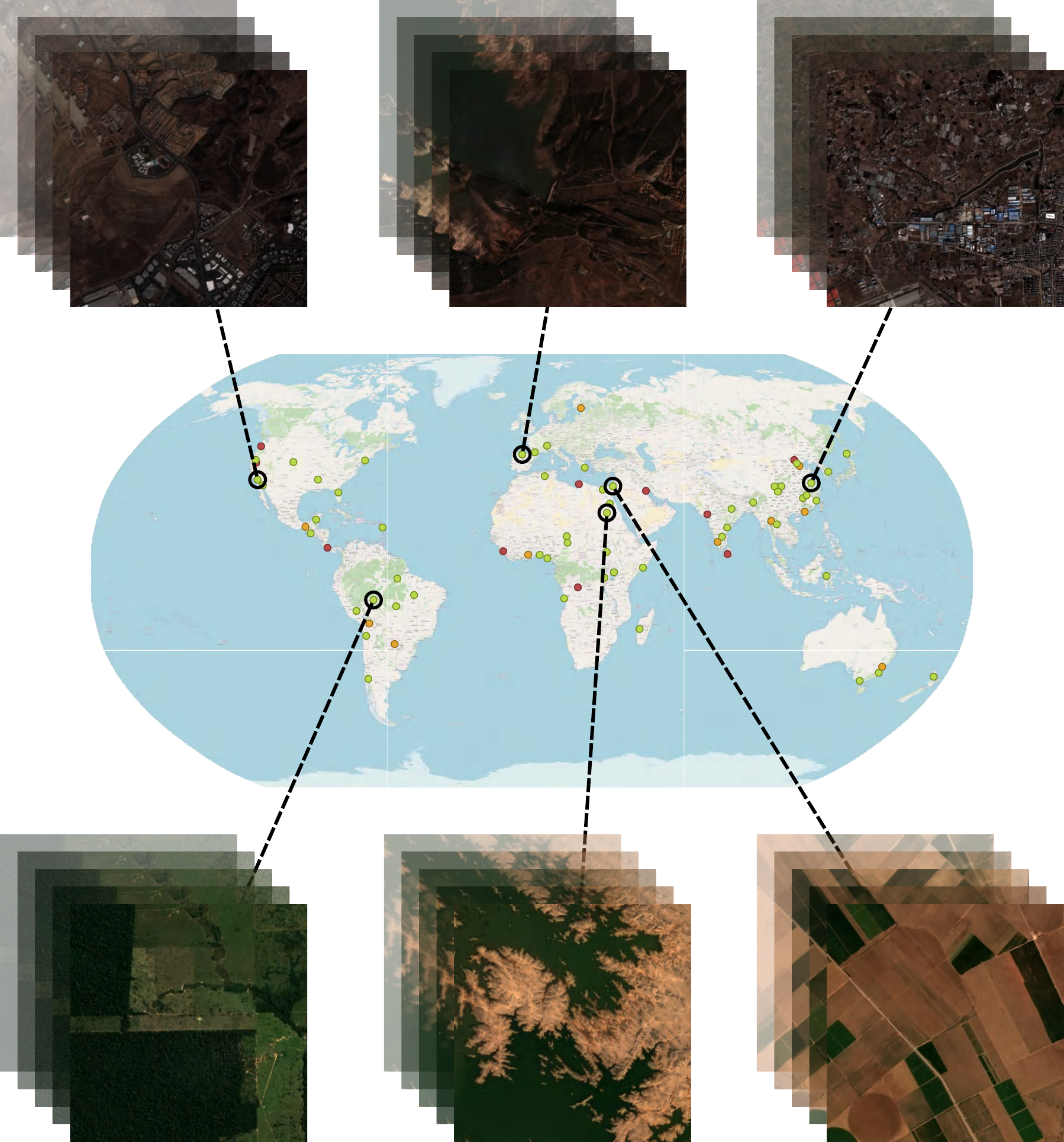}
    \caption{Visualization of the DynamicEarthNet dataset composed of 75 areas divided between train, validation and test sets (resp. green, orange and red dots).}
    \label{fig:den_dataset}
\end{figure}

DynamicEarthNet consists of a dataset of 75~SITS taken from all around the world. \Cref{fig:den_dataset} shows the location of each area, along with a few example SITS. The SITS acquired from January~2018 to December~2019 are composed of multispectral (RGB~+~near-infrared reflectance) images from the Planet Fusion product of size 1024~$\times$~1024~pixels at high spatial (3~meters per pixel) and temporal (daily) resolutions. A land cover ground truth is provided at the beginning of each month (\ie, 24 labeled images among the 730 per SITS) by manual annotation among 7~classes: 7.1\%~\textit{impervious surfaces}, 10.3\%~\textit{agriculture}, 44.9\%~\textit{forest \& other vegetation}, 0.7\%~\textit{wetlands}, 28.0\%~\textit{soil}, 8.0\%~\textit{water}, 1.0\%~\textit{snow}. To keep this use case easily reproducible, we choose to restrict the data to annotated images only, concentrating our task to the fully supervised setting. It is still possible to extend the method to a semi-supervised setting by keeping the unlabeled daily acquisitions, thus enhancing the temporal consistency.

\subsubsection{Graph design}
The downstream task consists of a pixel-wise classification for each date. The high spatial resolution implies a high intra-class variability and a large data volume, therefore it is interesting to aggregate the spatial information by forming spatial objects using an OBIA approach. In addition, land cover is usually consistent both in space and time, so processing each object related with its neighbors improves the spatio-temporal consistency and the classification accuracy. We thus assume that modeling these SITS with spatio-temporal graphs, classifying their nodes then mapping back to the pixel scale is adapted both to the data and the task, enabling a precise and dynamic land cover mapping. We describe below the procedure we used to extract spatio-temporal graphs from SITS.

A spatio-temporal graph is first composed by objects identified on SITS. To fix the intra-class variability, objects are designed as homogeneous regions because, at the local scale, neighboring pixels with similar spectral values probably share the same land cover class. To obtain these regions, we apply a segmentation algorithm to each SITS image producing disjoint objects for each dates. Identifying objects by date is chosen because the interval between acquisition dates is one month, so seasonal effects affecting land cover is significant and can be captured by spatio-temporal edges, keeping the object feature extractor simple. Among possible segmentation algorithms, we choose Felzenszwalb~\cite{felzenszwalb2004} to produce these regions because it extracts multi-scale object respecting the semantic boundaries and limiting the loss of small and thin objects such as roads. It also has a log-linear time complexity suitable for the large-scale images of DynamicEarthNet.

As described in~\Cref{sec:obj_feat}, the second step consists of defining region features discriminant for the land cover mapping task. In our case, we suppose that the spectral statistics of a region are sufficiently expressive, while remaining among the simplest features. The mean, the standard deviation, the minimal and the maximal values of each spectral band are used as the region embedding.

The third and last step is to add spatial and spatio-temporal edges to capture the spatial and temporal context of each node. In fact, for classes with similar spectral signatures, such as agriculture and vegetation, the spatio-temporal context can be useful. For this task, we use relationships with adjacent objects. Spatial edges are obtained using the region adjacency graph, whereas spatio-temporal edges are created if the spatial footprint of regions in two consecutive timestamps shared at least one pixel. Long-range spatio-temporal edges could be added to exploit the seasonal effect of land cover but we choose to remain as simple as possible for the purpose of this case study. Edges are not attributed, as the presence or absence of adjacency relationships is sufficient for our model. In our preliminary experiments, features such as the length of the common boundary or the spatio-temporal distance between two regions were found uninformative.

\subsubsection{Experimental setup}

To study the contribution of spatio-temporal graphs in the analysis of SITS, we compare GNNs to a pixel-based model as baseline, U-Net~\cite{ronneberger2015u}. It exploits only the spatial context at multiple scales and produces a segmentation map of results independently for each date. The graph method is evaluated with different graph convolutional layers to highlight the importance of GNN design choices related to the task, namely GCN~\cite{kipf2017gcn}, GraphSAGE~\cite{hamilton2017graphsage}, GIN~\cite{xu2018how}, GATv2~\cite{brody2021attentive} and ResGatedGCN~\cite{bresson2017resgatedgcn}. We also include a multi-layer perceptron (MLP), which acts as the previous GNNs but on edge-free graphs, \ie, without consideration of the spatio-temporal context, following a traditional OBIA framework.

GNN architectures consist of an encoder of four graph convolution layers, with an hidden dimension of 64, followed by a linear layer for classification. To aggregate information from both the spatial and temporal dimensions, the encoder consists of two layers aggregating only from the spatial neighbors, and two layers using temporal neighbors, as in~\Cref{fig:task_gnn}. Each convolution layer is followed by a batch normalization and a ReLU activation. U-Net experiments are run under the same setting as in the DynamicEarthNet paper~\cite{toker2022dynamicearthnet}. Thus, we refer to this work for the configuration of pixel-based baseline and to our implementation for other details about the setup. Another U-Net setting, denoted U-Net\textsubscript{/SITS}, is also evaluated for which an input batch contains images from only the same SITS, rather than single images extracted from multiple SITS. This setting provides a similar data loading scenario than the one used with GNNs. Although it is not optimal for U-Net by mitigating the learning of varied data in the same batch, this setting helps us to compare spatio-temporal and spatial-only methods more fairly by circumventing the limitation of input data size.

The Felzenszwalb algorithm~\cite{felzenszwalb2004} used to obtain objects relies on the Scikit-Image~\cite{scikit-image} implementation with $scale=100$ and $min\_size=5$ hyperparameters. All the deep learning models are trained using Adam optimizer on cross-entropy loss with a learning rate of 10\textsuperscript{-4}. The training process, for each model, is conducted over 100~epochs with training batch size equal to 1 for GNNs, 4 for U-Net and 24 for U-Net\textsubscript{/SITS}. Detailed hyperparameter values are available for all models at \url{github.com/corentin-dfg/graph4sits/}.

The assessment of the model performance is done considering the Intersection over Union~(IoU) and the Overall Accuracy~(OA) metrics at the pixel level, and the processing time. For graph-based methods, we obtain the output by mapping the node classifications back to the pixel scale using the segmentation map.

We use the provided split of DynamicEarthNet dataset with 55~SITS for training set and 10~SITS each for validation and test sets. This setting differs from our preliminary experiments in~\cite{dufourg2024forecasting} in which only the training set was available. It should be pointed out that the validation set seems more difficult than the test set, which explains the performance gap, also present in the dataset paper. To reduce the bias induced by model initialization, all the results are averaged over three different random initializations. The reported results are taken from the epoch that achieves the highest validation mIoU. For this study, and in line with the DynamicEarthNet baseline~\cite{toker2022dynamicearthnet}, we ignore the \textit{snow} class because this class only appear on 2~SITS of DynamicEarthNet.

Training and testing are carried out on Intel® Cascade Lake 6226 12 cores@2.7GHz with 384~GB of RAM and an Octo-GPU~v100. Preprocessing is done on CPU 12th Gen Intel® Core™ i7-12800H 20 cores@2.4-4.8GHz. Networks are implemented as defined in their original their paper using PyTorch and PyG.

\subsubsection{Experimental results}

\begin{table*}[!b]
\begin{center}
\setlength{\tabcolsep}{1.7pt}
\renewcommand{\arraystretch}{1.3}
\sisetup{table-format = 2.2\textsubscript{$\pm$aaaa},table-align-text-after = true, detect-weight}
\resizebox{\linewidth }{!}{
\begin{tabular}{r r | S S S S S S S S | S S}
     \hline
       & \multirowcell{2}{Nb. of\\params} & \multicolumn{6}{c}{\textit{per class IoU} ($\uparrow$)} & {\multirowcell{2}{Val\\mIoU ($\uparrow$)}} & {\multirowcell{2}{Val\\OA ($\uparrow$)}} & {\multirowcell{2}{Test\\mIoU ($\uparrow$)}} & {\multirowcell{2}{Test\\OA ($\uparrow$)}} \\\cline{3-8}
       & & {Impervious} & {Agriculture} & {Forest} & {Wetlands} & {Soil} & {Water} & & & & \\
     \hline
     U-Net & 983,654 & {\bfseries 29.60\textsubscript{$\pm$0.31}} & {\bfseries 6.48\textsubscript{$\pm$1.37}} & {\bfseries 77.43\textsubscript{$\pm$0.99}} & 0.00\textsubscript{$\pm$0.00} & {\bfseries 37.92\textsubscript{$\pm$1.73}} & 50.70\textsubscript{$\pm$3.08} & {\bfseries 33.69\textsubscript{$\pm$0.71}} & {\bfseries 72.86\textsubscript{$\pm$0.99}} & {\bfseries 37.71\textsubscript{$\pm$1.22}} & {\bfseries 69.85\textsubscript{$\pm$2.82}} \\
     U-Net\textsubscript{/SITS} & 983,654 & 0.00\textsubscript{$\pm$0.00} & 0.00\textsubscript{$\pm$0.00} & 57.93\textsubscript{$\pm$1.53} & 0.00\textsubscript{$\pm$0.00} & 10.49\textsubscript{$\pm$6.83} & 4.12\textsubscript{$\pm$4.42} & 12.09\textsubscript{$\pm$0.53} & 56.81\textsubscript{$\pm$1.32} & 9.25\textsubscript{$\pm$0.59} & 39.45\textsubscript{$\pm$1.89} \\
     \hdashline
     MLP & 14,470 & 12.83\textsubscript{$\pm$2.27} & 0.01\textsubscript{$\pm$0.01} & 75.36\textsubscript{$\pm$0.95} & 0.00\textsubscript{$\pm$0.00} & \Uline{32.77}\textsubscript{$\pm$0.45} & 55.53\textsubscript{$\pm$7.18} & 29.42\textsubscript{$\pm$1.06} & 71.46\textsubscript{$\pm$0.76} & 34.81\textsubscript{$\pm$0.43} & 67.76\textsubscript{$\pm$0.12} \\
     GCN & 14,470 & 12.90\textsubscript{$\pm$7.13} & 0.10\textsubscript{$\pm$0.14} & 74.04\textsubscript{$\pm$0.52} & 0.00\textsubscript{$\pm$0.00} & 32.03\textsubscript{$\pm$0.82} & \Uline{60.22}\textsubscript{$\pm$8.64} & 29.88\textsubscript{$\pm$1.96} & 70.69\textsubscript{$\pm$0.98} & 32.38\textsubscript{$\pm$1.61} & 66.07\textsubscript{$\pm$1.28} \\
     GraphSAGE & 27,782 & \Uline{20.61}\textsubscript{$\pm$2.41} & \Uline{1.01}\textsubscript{$\pm$0.98} & \Uline{75.45}\textsubscript{$\pm$0.71} & 0.00\textsubscript{$\pm$0.00} & 32.20\textsubscript{$\pm$0.70} & 59.59\textsubscript{$\pm$3.48} & \Uline{31.48}\textsubscript{$\pm$0.51} & \Uline{71.67}\textsubscript{$\pm$0.20} & \Uline{35.81}\textsubscript{$\pm$0.98} & \Uline{68.21}\textsubscript{$\pm$0.71} \\
     GIN & 14,474 & 0.00\textsubscript{$\pm$0.00} & 0.00\textsubscript{$\pm$0.00} & 73.33\textsubscript{$\pm$1.05} & 0.00\textsubscript{$\pm$0.00} & 31.57\textsubscript{$\pm$1.82} & 54.57\textsubscript{$\pm$5.58} & 26.58\textsubscript{$\pm$1.23} & 69.42\textsubscript{$\pm$1.01} & 27.71\textsubscript{$\pm$4.29} & 63.79\textsubscript{$\pm$3.39} \\
     GATv2 & 28,550 & 18.36\textsubscript{$\pm$3.29} & 0.06\textsubscript{$\pm$0.09} & 74.08\textsubscript{$\pm$0.62} & 0.00\textsubscript{$\pm$0.00} & 31.17\textsubscript{$\pm$0.14} & {\bfseries 63.22\textsubscript{$\pm$2.86}} & 31.15\textsubscript{$\pm$1.06} & 71.30\textsubscript{$\pm$0.52} & 31.99\textsubscript{$\pm$1.67} & 63.92\textsubscript{$\pm$0.24} \\
     ResGatedGCN & 55,174 & 0.00\textsubscript{$\pm$0.00} & 0.00\textsubscript{$\pm$0.00} & 71.14\textsubscript{$\pm$1.52} & 0.00\textsubscript{$\pm$0.00} & 31.06\textsubscript{$\pm$0.06} & 48.05\textsubscript{$\pm$4.60} & 25.04\textsubscript{$\pm$0.56} & 68.02\textsubscript{$\pm$1.03} & 27.99\textsubscript{$\pm$0.46} & 63.40\textsubscript{$\pm$1.06} \\
     \hline
\end{tabular}
}
\caption{Intersection over Union~(IoU) and Overall Accuracy~(OA) performance of pixel-based and graph-based models on validation and test sets. The results are provided with average and standard deviation on three random initializations (\textbf{best}, \underline{second best}).}
\label{tab:gden_perf}
\end{center}

\end{table*}
 
\begin{figure*}[t]
	\centering
	\captionsetup[subfigure]{labelformat=empty,position=top,skip=1pt}
	\begin{minipage}{\linewidth}
	\subfloat{%
    	\includegraphics[clip,width=.0625\linewidth]{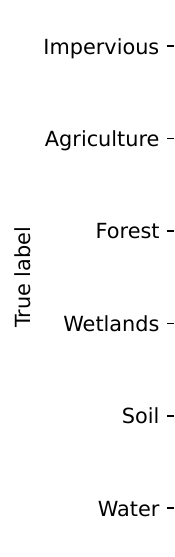}%
    }%
	\subfloat[U-Net]{%
    	\includegraphics[clip,width=.2\linewidth]{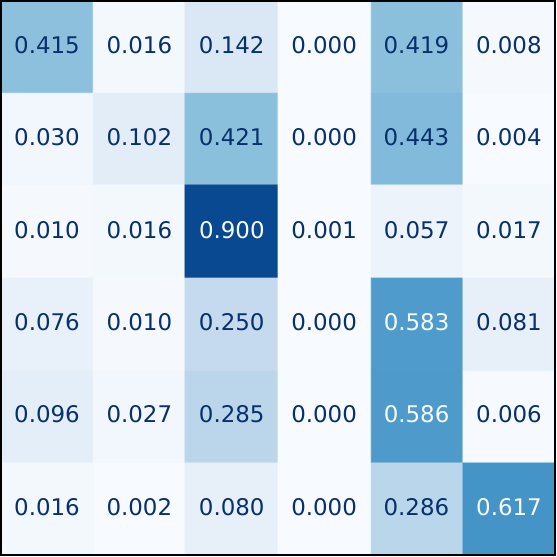}
    }
	\hfill
	\subfloat[U-Net\textsubscript{/SITS}]{
    	\includegraphics[clip,width=.2\linewidth]{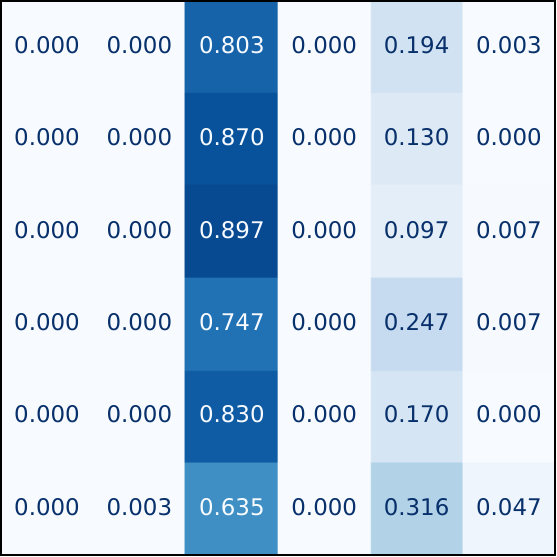}
    }
	\hfill
	\subfloat[MLP]{
    	\includegraphics[clip,width=.2\linewidth]{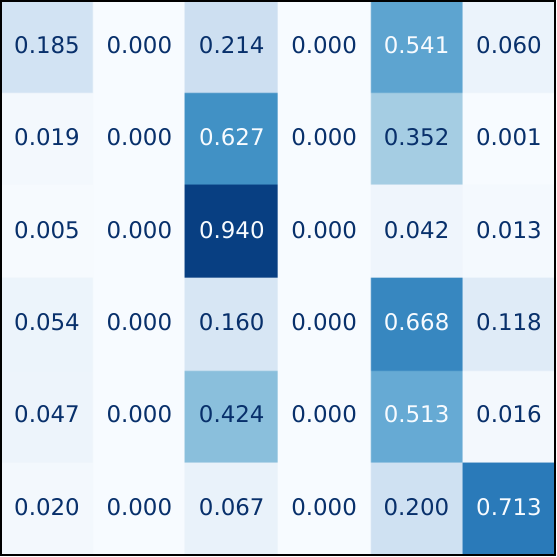}
    }
    \hfill
	\subfloat[GCN]{
	    \includegraphics[clip,width=.2\linewidth]{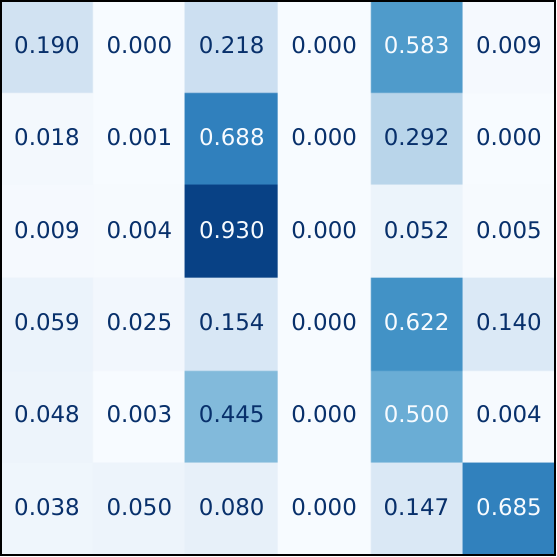}
	}
	\end{minipage}
	\\
	\vspace{1em}
	\begin{minipage}{\linewidth}
	\subfloat{%
    	\includegraphics[clip,width=.0625\linewidth]{img/gden/confusion_matrix_borderless/yaxis.pdf}%
    }%
	\subfloat[SAGE]{%
	    \includegraphics[clip,width=.2\linewidth]{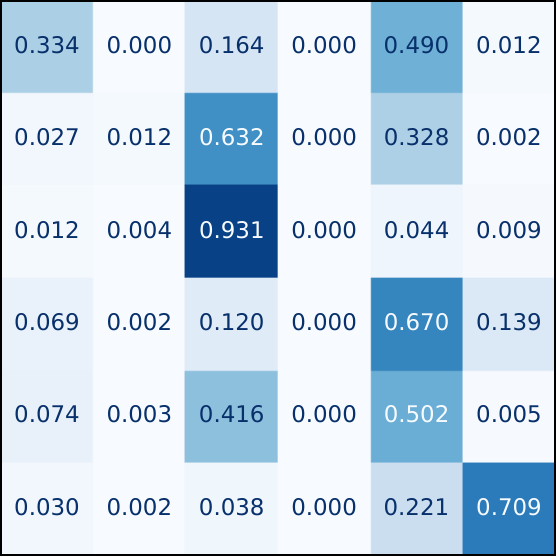}
	}
	\hfill
	\subfloat[GIN]{
	    \includegraphics[clip,width=.2\linewidth]{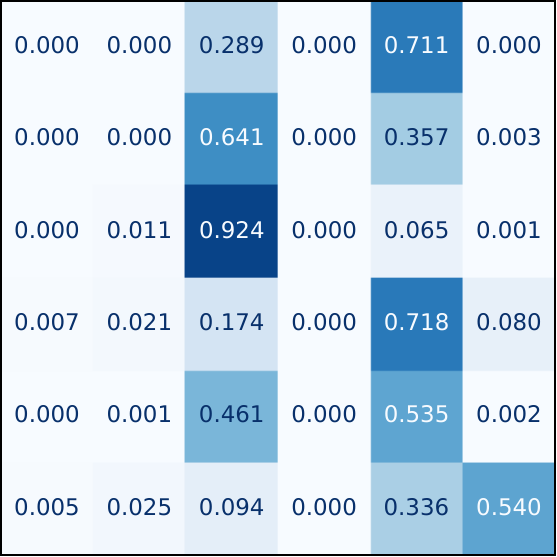}
	}
	\hfill
	\subfloat[GATv2]{
	    \includegraphics[clip,width=.2\linewidth]{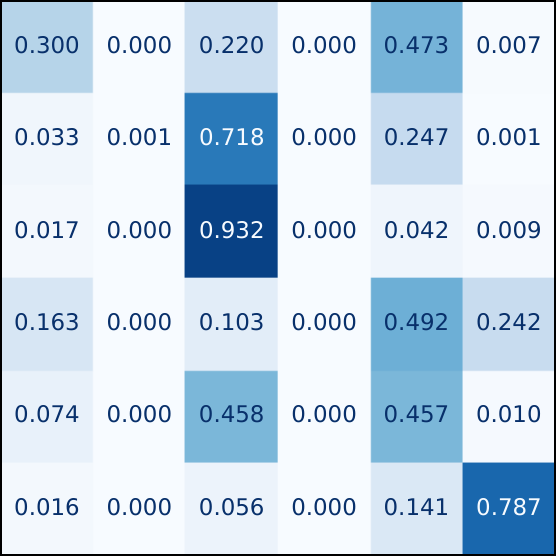}
	}
	\hfill
	\subfloat[ResGatedGCN]{
	    \includegraphics[clip,width=.2\linewidth]{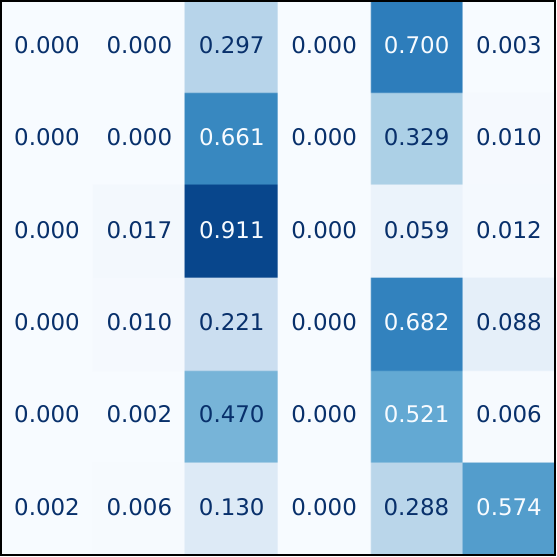}
	}
	\end{minipage}
	\\
	\begin{minipage}{\linewidth}
    \hspace{.0625\linewidth}%
    \subfloat{%
	    \includegraphics[clip,width=.2\linewidth]{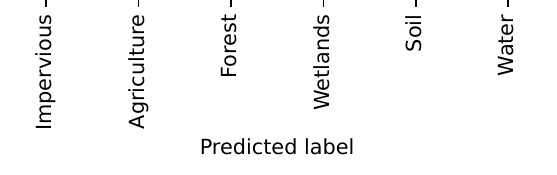}
	}
	\hfill
	\subfloat{
	    \includegraphics[clip,width=.2\linewidth]{img/gden/confusion_matrix_borderless/xaxis.pdf}
	}
	\hfill
	\subfloat{
	    \includegraphics[clip,width=.2\linewidth]{img/gden/confusion_matrix_borderless/xaxis.pdf}
	}
	\hfill
	\subfloat{
	    \includegraphics[clip,width=.2\linewidth]{img/gden/confusion_matrix_borderless/xaxis.pdf}
	}
	\end{minipage}
	
	\caption{Confusion matrices on the validation set. The results are normalized over the targets and provided with average on three random initializations.}
	
	\label{fig:gden_confusion_matrix}
\end{figure*}

\Cref{tab:gden_perf} summarizes the average quantitative performances on DynamicEarthNet achieved by the pixel-based baseline U-Net, MLP and GNNs. To highlight the limitations of the different models, we provide in \Cref{fig:gden_confusion_matrix} complementary information with the confusion matrices over the validation set. In addition, part of the inter-class error is due to class imbalance in the dataset, which is suggested by the similar structure of each confusion matrix, regardless of the model. In the following, we first highlight the role of context information for the case study compared to a context-agnostic model, such as MLP. Then, we inspect the GNN performances related to the pixel-based baseline. We also analyze the inherent error of object-based methods that results from the semantic segmentation step. Finally, we inspect the computational performance of the models in terms of training and testing time.

\paragraph{Local \textit{vs.} context information}

The GraphSAGE model outperforms MLP on accuracy and IoU metrics, demonstrating the value of using the spatio-temporal context of an object in addition to its own features to determine its nature. These performances are significantly better for the classes \textit{agriculture}, \textit{water}, and especially \textit{impervious surface}. These are the classes for which determination by spectral characteristics is limited, either because of the very nature of these classes (\eg, \textit{agriculture} is confused with \textit{forest} and \textit{soil}, as shown by \Cref{fig:gden_confusion_matrix}) or because of high intra-class variability. Indeed, \textit{impervious surface} is confused with \textit{forest}, \textit{soil} and \textit{water}. A possible source of this confusion is because geographical objects in the \textit{impervious surface} class are generally smaller and more difficult to segment into regions. The gap between MLP and GraphSAGE performance thus shows that learning the context of each object decreases the class ambiguity.

However, the way in which the context of objects is understood and integrated into learning is also important, as underlined by the fact that other graph-based models do not outperform MLP. GCN's poorer performance can be explained by the fact that the features of neighboring regions are aggregated in the same way as those of the central node, making it harder for the network to retain local information~\cite{hamilton2017graphsage}, which is crucial in a semantic segmentation task.

Some of the failures are also due to the neighborhood aggregator used. Indeed, the aggregation function is one of the most critical components of GNN as it designs which property of the neighborhood is determinant. For instance, neighborhood distribution is better captured by the mean aggregator, structure information, such as node degree, by the sum aggregator, and specific features of some neighbors by the max aggregator~\cite{xu2018how}. The difference between GraphSAGE and GIN, which are inherently similar except for their aggregation function (mean and sum, respectively), suggests that the neighborhood distribution is more important than the graph structure to classify regions. Therefore, GNNs that use a (weighted) mean aggregator, such as GCN, GraphSAGE and GATv2, perform better.

Improvements over convolutional GNNs include attention (GATv2) or gate (ResGatedGCN) mechanisms, which enable anisotropic aggregation of neighbor information. From a message-passing point of view (see~\Cref{sec:supervised_methods}), self-attention consists in weighting incoming messages according to the similarity of the target node with the source nodes. The gate mechanism is an independent weighting of the features of each incoming message, still depending on the characteristics of the source and target nodes. Contrary to expectations, the performance of these two networks is inferior to that of a GNN with isotropic aggregation such as GraphSAGE, which treats all its neighbors in the same way. Regarding GATv2, even though the benefits of dynamic context awareness are found for \textit{impervious surface} and especially \textit{water} compared to MLP, it seems that performances for \textit{forest \& other vegetation} and \textit{soil} are lower than those of MLP and GraphSAGE. With ResGatedGCN, there is a drop in performance compared to GraphSAGE and even to MLP, similar to the GIN performance. This does not seem to be due to anisotropic aggregation---otherwise a similar observation would be made with GATv2---but rather to the actual aggregation function used. The weighting from edge gates is not normalized, unlike in GATv2, and the aggregator corresponds more to a weighted sum than a weighted average. As a result, despite the greater expressiveness of ResGatedGCN (in fact, gates can be seen as a soft attention process), GATv2 achieves better performance.

This analysis helps explain why GraphSAGE is a better choice of graph convolution for the studied task and data. This comparison allows us to determine which characteristics of graph convolution are important for the node classification of a spatio-temporal graph in our case study. It may help to build a more effective model in the future.

\paragraph{Pixel-based \textit{vs.} graph-based methods}

The performance of GraphSAGE---previously identified as the best graph-based model for the case study---is poorer than that of U-Net on all metrics (with the exception of the \textit{water} class). This could be explained by the more complex feature extraction achieved by U-Net, whereas GNNs rely solely on features carried by nodes, which are handcrafted through the design of the spatio-temporal graph, and therefore sub-optimal. Extracting features from regions is all the more important as the downstream task is local and, as seen in the previous section, the features of the targeted region are crucial for its classification. Deep feature extraction could potentially improve performance by employing, for example, a pixel-set encoder~\cite{garnot2020} to build data-dependent features. Related to the limitation of the graph design, the creation of spatial and spatio-temporal edges is the result of an \textit{a priori} knowledge, but it would be interesting to further study the impact of such choices using interpretability tools. Finally, although the partitioning of the SITS into objects induces an intrinsic error in the model (see~\Cref{sec:gden_segm_error}), this is limited and only partially explains the prediction error at pixel level.

Data batch processing is another criterion influencing the performance of U-Net and GNNs. Indeed, as U-Net uses only the spatial dimension and works by date, it is easier to integrate different locations in the same batch. This makes it easier to take into account data variety and class distribution at each network iteration. On the other hand, the use of spatio-temporal graphs restricts the size of the input batch due to the volume of each graph, recalling that each graph in the current experiment is built from 24 images. The input data at each iteration is therefore less representative of the dataset as a whole, resulting in slower and less stable convergence. To illustrate this phenomenon, the U-Net model was trained with a batch size of 24 corresponding to one SITS per iteration, thus approximating the data loading used with GNNs. The results are reported in \Cref{tab:gden_perf} as U-Net\textsubscript{/SITS} and clearly demonstrate the importance of data representativeness in a batch. The trade-off between representativeness and input data size is crucial in spatio-temporal models. Comparing the results of the U-Net\textsubscript{/SITS} model trained on a similar setting than the graph-based models shows the value of a spatio-temporal approach.

Finally, the performance gap between U-Net and GraphSAGE shows that the use of spatio-temporal graphs and GNNs, despite their advantages, is still in a maturing phase that requires improvements in design and learning strategy depending on the target task to reach the level of pixel-based methods.

\paragraph{Inherent error in object-based methods}
\label{sec:gden_segm_error}

Part of the semantic segmentation error is inherent in object-based approaches evaluated at pixel level. In fact, performance is bounded by an optimal classification of regions, depending on the metric under study. In the current case, the maximum overall accuracy, corresponding to the prediction for each region of its majority class, is 92\% for the train set, 92\% for the validation set and 91\% for the test set.
The magnitude of this error, around 9\%, is smaller than current predictions, which are around 30\% in the best case. Therefore, in this case, the impact of segmentation into regions can be considered as less severe than the error of networks on region classification.

\paragraph{Computational considerations}

Although U-Net offers better performance, it is a heavyweight model. GraphSAGE is much lighter---number of parameters divided by 35---and faster. \Cref{tab:gden_time} shows the runtime for various steps: data preprocessing, training and model testing. The training of GraphSAGE is in fact 5 times faster, as shown by the \textit{Total training time} which includes data preprocessing, training on the train set and evaluation on the train and validation sets, for profiling and model selection respectively.

The bottleneck of graph-based approaches in terms of computational cost is their preprocessing, \ie, converting a SITS into a spatio-temporal graph. This can be seen in the \Cref{tab:gden_time} where 98\% of the test time for graph-based approaches is dedicated to preprocessing. The GNN forward pass is~0.5s, much faster than U-Net. This bottleneck does not appear in the training stage, as preprocessing only needs to be carried out once and then becomes negligible compared to the learning time.

The cost of preprocessing is mainly due to the segmentation algorithm, in this case Felzenszwalb, which is not optimized and runs on the CPU, just like the rest of the preprocessing. Some approaches, such as \cite{xu2023esnet}, claim GPU-suited segmentation methods up to 30 times faster than traditional algorithms such as SLIC. Regarding GPU optimization of neural networks, despite the efforts of the graph deep learning community, GNNs are still a more recent development than CNNs, and more complex to optimize for high-performance computing due to sparse matrix computation, whereas CNNs have been optimized for years in both software and hardware.

Further improvements on the computational cost of graph-based methods may achieve a testing time per SITS similar or even smaller than U-Net while keeping a much lower training time, by reducing the preprocessing duration.

\begin{table}
\begin{center}
\setlength{\tabcolsep}{1.7pt}
\renewcommand{\arraystretch}{1.3}
\begin{tabular}{r r | S | S | S | S }
     \hline
       & {\multirowcell{2}{Nb. of\\params}} & {\multirowcell{2}{Preprocessing\\time (s/SITS)}} & {\multirowcell{2}{Training\\time (s/epoch)}} & {\multirowcell{2}{Total training\\time (h)}} & {\multirowcell{2}{Total testing\\time (s/SITS)}}  \\
       & & & & & \\
     \hline
     U-Net & 983,654 & {-} & 382.66 & 10.63 & 3.86 \\
     \hdashline
     MLP & 14,470 & 34.26 & 57.76 & 2.22 & 34.81 \\
     GCN & 14,470 & 34.26 & 59.43 & 2.27 & 34.82 \\
     GraphSAGE & 27,782 & 34.26 & 61.24 & 2.32 & 34.81 \\
     GIN & 14,474 & 34.26 & 58.09 & 2.23 & 34.81 \\
     GATv2 & 28,550 & 34.26 & 61.25 & 2.32 & 34.85 \\
     ResGatedGCN & 55,174 & 34.26 & 59.50 & 2.27 & 34.82 \\
     \hline
\end{tabular}
\caption{Runtime of preprocessing, training and testing steps. It might be noted that the preprocessing step is not optimized and ran on CPU only. The results are provided with average on three random initializations.}
\label{tab:gden_time}
\end{center}

\end{table}

\subsubsection{Discussion}
This case study highlights a practical downstream task by illustrating the different design and processing questions which can emerge by using spatio-temporal graphs for SITS semantic segmentation over time. The main purpose of spatio-temporal graphs lies in the integration of spatial and temporal information into the same data structure, thus enabling the direct use of spatio-temporal context. They also limit the data redundancy, as the processing is based on homogeneous regions, thus reducing computational costs. However, despite these qualities, graph methods are understudied, and the design choices are poorly explained even for specific applications. To enable better use of graph structure for SITS, we encourage the scientific community to further study these issues to achieve an efficient pipeline for classification tasks, particularly with a view to matching the performance of pixel-based methods. A number of avenues have been highlighted in the previous sections, including optimization of object segmentation (\eg, learned segmentation, 3D spatio-temporal segmentation of the datacube), improvement of deep feature extraction for irregular objects, and study of the properties of the SITS-based graphs (\eg, heterophily~\cite{luan2024heterophilic}, connectivity, heterogeneity, class balancing) to subsequently guide design and processing choices. Concerning the learning part, transferring the latest advances from graph deep learning is necessary, both in terms of learning strategy (subgraph batch training~\cite{hamilton2017graphsage, chiang2019cluster}) and architecture (aggregation functions~\cite{corso2020principal, buterez2022graph}, deeper GNN~\cite{li2021training}, graph transformers~\cite{dwivedi2021generalization}, space-first time-later \textit{vs.} time-first space-later \textit{vs.} joint space-time~\cite{camara2016big, gao2022equivalence}). Finally, it is important that design and architecture choices in future work are explicitly motivated, and the use of interpretability tools looks promising for these purposes. The current maturation phase of spatio-temporal graphs applied to SITS is crucial, and we hope that this case study may aid future work in classification and semantic segmentation.

\subsection{Water resources forecasting with graph neural networks}

Like land cover, water is also one of the 14 global fundamental geospatial data themes defined within the United Nations framework~\cite{unggim2019}. Water is critical to almost every process on Earth and is essential for life. Water management is therefore crucial, and tools are needed to determine the availability of water resources. This second case study explores forecasting in water basin areas and, in the context of this paper, provides a good example of the use of graphs for such a forecasting task.

\subsubsection{Data}
\label{subsec:gc_dataset}

Various types of satellite acquisitions are useful to study water bodies such as altimetry, radar detection and optical imagery. In this use case, the evolution of water bodies and the impact on their environment, in particular agricultural fields, is studied with optical imagery.

To monitor changes in water content, the Normalized Difference Water Index (NDWI)~\cite{mcfeeters1996use} is computed using the green (B03) and near-infrared (B08) 10-meter resolution bands from \mbox{Sentinel-2}, with the formula~$\displaystyle \frac{\text{B03}-\text{B08}}{\text{B03}+\text{B08}}$. This formulation enables the detection of water bodies and fine variations in water content, providing insights into overall turbidity, including suspended sediments and chlorophyll $\alpha_t$. Its sensitivity to water levels also makes it a suitable choice for forecasting water resources within basins. However, NDWI can be influenced by built-up features (\eg, buildings, roads, and bridges) and atmospheric conditions, particularly clouds, which may result in the overestimation of water bodies when using a simple thresholding approach to identify water content \cite{xu2006modification}. Regarding other surfaces such as soil, dry and green vegetation, the NDWI provides zero and negative values but is still informative, even if the difference between land cover is less pronounced than for water.

\begin{figure}
    \centering
    \includegraphics[width=.5\linewidth]{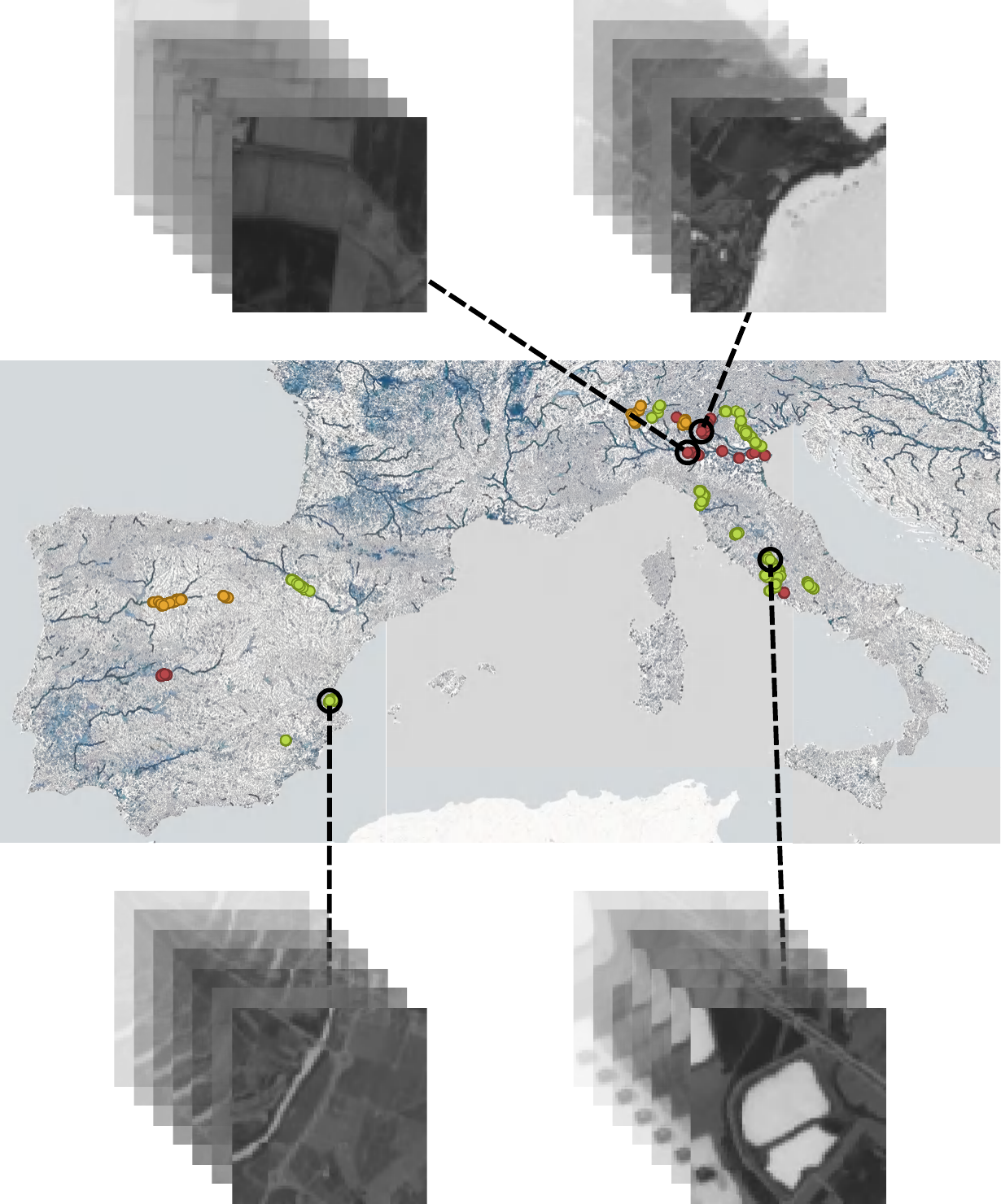}
    \caption{Visualization of the SEN2DWATER dataset composed of 17 water basins split in train, validation and test sets (resp. green, orange and red dots).}
    \label{fig:gc_dataset}
\end{figure}

To conduct the NDWI forecasting task, we used the existing SEN2DWATER dataset~\cite{mauro2023sen2dwater}. The original dataset consists of 5,264 sequences of \mbox{Sentinel-2} image patches gathered in 17 different basins, mainly in Italy and Spain, between July 2020 and December 2022. The areas cover the surroundings around lakes and rivers, mainly in agricultural regions where basin water resources might be used for irrigation. Each SITS patch has a size of 64 $\times$ 64 pixels and a temporal resolution of around two months (only the less cloudy images were processed). We filtered out SITS that contained missing data in the original dataset, leading to a total of 3,682 sequences of \mbox{Sentinel-2} image patches for the experiments. Their locations is shown in~\Cref{fig:gc_dataset}. To facilitate the reproducibility of this study, we provide our curated NDWI version of the SENS2DWATER as well as the code (\url{github.com/corentin-dfg/graph4sits/}).

\subsubsection{Graph design}
\label{subsec:gc_ourMethod}

\begin{figure*}
    \begin{center}
    \includegraphics[width=\linewidth]{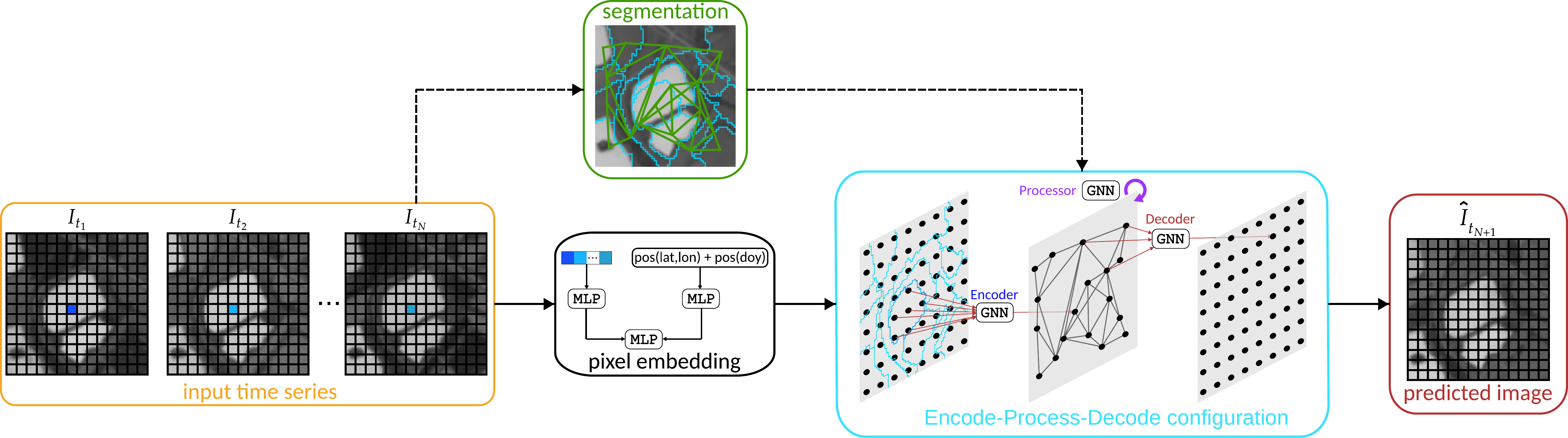}
    \end{center}
    \caption{Overview of our graph-based pipeline. The input is a sequence of satellite image time series acquired at dates $t_1$, $t_2$, ..., $t_N$. The output is the predicted image at $t_{N+1}$.}
    \label{fig:gc_pipeline}
\end{figure*}

To perform forecasting of NDWI, the autoregressive architecture of GraphCast~\cite{lam2023learning} is adapted to SITS with high spatial resolution and low temporal one. The employed graph-based model follows the same encoder-processor-decoder structure. The encoder projects the input state of a local region (\ie, a group of pixels) into the nodes of a coarser mesh graph, at a scale that makes it easier to learn the dynamics of the physical system. Then, the processor learns latent representations of the mesh nodes via message passing. Finally, the decoder maps the learned features to each original pixel coordinates using only the three nearest mesh nodes. It predicts the future state of the system as a residual of the last known state, \ie, it outputs the difference with the most recent input. \Cref{fig:gc_pipeline} displays the full architecture.

At the heart of this architecture is the coarse-mesh graph, which supports learning about system dynamics. This study focuses on location-specific high-resolution images covering small areas. Thus the mesh should result from the discretization of the studied areas. Rather than a regular mesh, using geographical objects to build the processor graph should make it easier to learn about the dynamics of water basins and the local interactions. In satellite imagery, homogeneous regions can help approximate these objects. Therefore, in this study, the simple linear iterative clustering (SLIC) algorithm~\cite{achanta2012} is used on the last available acquisition, assuming that the geographical objects at the targeted date should be similar.

Regarding the spatial edges, they are divided in three categories in this architecture, depending if there are used by the encoder, the processor or the decoder. The encoder maps the initial pixel grid to the coarse mesh by means of oriented edges linking each pixel of a geographic object to the corresponding node in the processor mesh. To learn the relationships between the geographical objects, the processor mesh is based on a region adjacency graph. Finally, the decoder uses oriented edges from the three nearest mesh nodes of the target pixel to learn interpolation. Each of these edges are attributed with the relative distance between the source and the target nodes. The encoder, processor and decoder all use the same type of message-passing layer proposed by~\citet{battaglia2018relational}, and the difference between the three modules lies in the design of the subgraphs.

In contrast to the previous case study, the temporal dimension is not processed within a spatio-temporal graph, but using a pixel-wise temporal embedding as node features of a spatial graph. The NDWI pixel time series is processed by an MLP to obtain its embedding. A spatio-temporal positional embedding is also obtained from the processing by an MLP of the concatenation of the sine of the latitude, the sine and cosine of the longitude, and the sine of the day of the year. Subsequently, the pixel time series embedding and the spatio-temporal positional encoding are concatenated before being passed through a third MLP to obtain the pixel node's features. In this way, temporal information is already aggregated for each pixel along with its geographical coordinates around the world.

\subsubsection{Experimental setup}
\label{subsec:gc_implem}

To study the forecasting performance of this graph-based model, it is compared with three competitors proven to be efficient for SITS analysis \cite{russwurm2017temporal,russwurm2018,mauro2023sen2dwater}: (i) Long Short Term Memory (LSTM), (ii) \mbox{ConvLSTM}, and (iii) Time Distributed CNN \mbox{ConvLSTM} (\mbox{TDCNN-ConvLSTM}), the best-performing strategy (originally called \mbox{TD-CNN}) tested on the SEN2DWATER dataset \cite{mauro2023sen2dwater}. The \mbox{TDCNN-ConvLSTM} architecture is similar to \mbox{ConvLSTM} but first performs an embedding of each image individually with 2D spatial convolutions, enriching the image's context. While SITS patches are inputted in \mbox{ConvLSTM} and \mbox{TDCNN-ConvLSTM}, LSTM is applied pixel-wise and thus is agnostic to the spatial dimension. These three models predict directly the next image in the sequence, not the residuals as in our model.

Two additional weak baselines are also provided: (i) input average that corresponds to predict the temporal average image, and (ii) persistence which means predicting the same image as the last acquisition.

Regarding the inputs, the first six acquisitions of each site in the dataset are used, which corresponds to a period of approximately one year from July 2016, following the setting proposed in~\cite{mauro2023sen2dwater}. This data sampling strategy makes the model less robust in predicting periods other than summer. Let us note that this case study is not intended to be deployed, but to serves as a proof of concept for the use of graphs in a SITS forecasting task.

Besides, the segmentation map used to build the graph structure is generated with the SLIC algorithm implemented in Scikit-Image \cite{scikit-image}. By default, SLIC is applied to the last image of the input sequence with a number of superpixels of 256 and a compactness value of 0.1. All the networks are trained using the Huber loss for 50 epochs and Adam optimizer with an initial learning rate of 0.0001, decreased by 90\% when the validation loss does not reduce over the past five epochs. Detailed hyperparameter values of architectures and other training settings are available at~\url{github.com/corentin-dfg/graph4sits/}.

The assessment of the model performance is done considering widely used quality measures averaged over all the test images. In particular, we use reconstruction and perception-based metrics, including root mean square error (RMSE), peak signal-to-noise ratio (PSNR), and structural similarity index measure (SSIM)~\cite{wang2004image}. Additionally, we evaluate and compare visually the predicted NDWI images.

To guarantee the independence of the train and test sets, we ensure that image patches from the same basin cannot be in different sets, contrarily to the split proposed in the original paper~\cite{mauro2023sen2dwater}. The SITS patches are split using an 85:15 ratio. The training set is further split into train and validation sets, used to monitor the performance and decrease the learning rate during the model's optimization. 

Training and testing are carried out on a CPU 12th Gen Intel® Core™ i7-12800H × 20 and an NVIDIA RTX A1000. Baseline implementations mirror the work of \citet{mauro2023sen2dwater}. Our model implementation in Pytorch and PyG is adapted from the original Python/JAX code of GraphCast.

\subsubsection{Experimental results}

\Cref{tab:gc_results_sen2dwater} summarizes the average quantitative performances on SEN2DWATER achieved by our graph-based method and the baselines. In the following, we first quantitatively assess the reconstruction performance of the models and their ability to handle changes. Then, we visually analyse the predictions thanks to a qualitative study on two representative areas of the dataset: lakeshores and farmland. We also explore the influence of the graph design, notably of the segmentation step. Finally, we explore the choices made concerning the given inputs, especially about the time series length.

\paragraph{Quantitative analysis}

\begin{table*}[h!]
\begin{center}
\sisetup{table-align-text-after = true, detect-weight}
\begin{tabular}{r r S[table-format = 1.4\textsubscript{$\pm$aaaaaa}] S[table-format = 2.2\textsubscript{$\pm$aaaa}] S[table-format = 1.4\textsubscript{$\pm$aaaaaa}] c}
\hline
 & \# Params & {RMSE $\downarrow$} & {PSNR $\uparrow$} & {SSIM $\uparrow$} & Runtime (min) \\
\hline
Input average & - & 0.1550 & 23.32 & 0.7465 & - \\
Persistence & - & 0.1332 & 25.03 & 0.7897 & - \\
\hdashline
LSTM & 17,345 & 0.1162\textsubscript{$\pm$0.0005} & 25.53\textsubscript{$\pm$0.05} & \bfseries 0.8282\textsubscript{$\pm$0.0005} & 22 \\
ConvLSTM & 150,721 & 0.1197\textsubscript{$\pm$0.0029} & 25.28\textsubscript{$\pm$0.19} & 0.8113\textsubscript{$\pm$0.0030} & 26 \\
TDCNN-ConvLSTM & 407,681 & 0.1111\textsubscript{$\pm$0.0008} & 25.68\textsubscript{$\pm$0.08} & 0.8083\textsubscript{$\pm$0.0008} & 46 \\
Graph-based & 228,673 & \bfseries 0.1097\textsubscript{$\pm$0.0035} & \bfseries 26.42\textsubscript{$\pm$0.27} & 0.8170\textsubscript{$\pm$0.0070} & 41 \\
\hline
\end{tabular}
\end{center}
\caption{Number of parameters, RMSE, PSNR, SSIM and runtime for baseline and graph-based models. The mean and standard deviation are reported for three initializations of the model's parameters. Bold values display the best performance for each metric.}
\label{tab:gc_results_sen2dwater}
\end{table*}

According to \Cref{tab:gc_results_sen2dwater}, all learning models outperform the two weak baselines, especially the Persistence one, which is particularly interesting to compare with a residual model. The graph-based model yields the best performance on the reconstruction metrics. The LSTM obtains the lowest RMSE and PSNR results among the competitors, but it yields the best performance on SSIM, a perceptual metric, despite the independent processing of each pixel. To derive water content, we assume that a low reconstruction error is more critical than visually pleasant results. The SSIM metrics is also more sensitive to highly structured areas such as crop fields, as opposed to water bodies. Considering model complexity, the total runtime, including both training and inference, is comparable between the graph-based model and TDCNN-ConvLSTM, which has twice as many parameters. Let us note that the total runtime includes all data preprocessing, including on-the-fly graph construction. The latter is very fast, compared to the previous case study, as a single spatial graph is built, requiring the segmentation of only one image instead of segmenting each image of the SITS. The smaller size of the SITS, both spatially and temporally, also explains the difference in graph construction runtime between the two case studies.

\begin{figure}
    \centering
    \includegraphics[width=.7\linewidth]{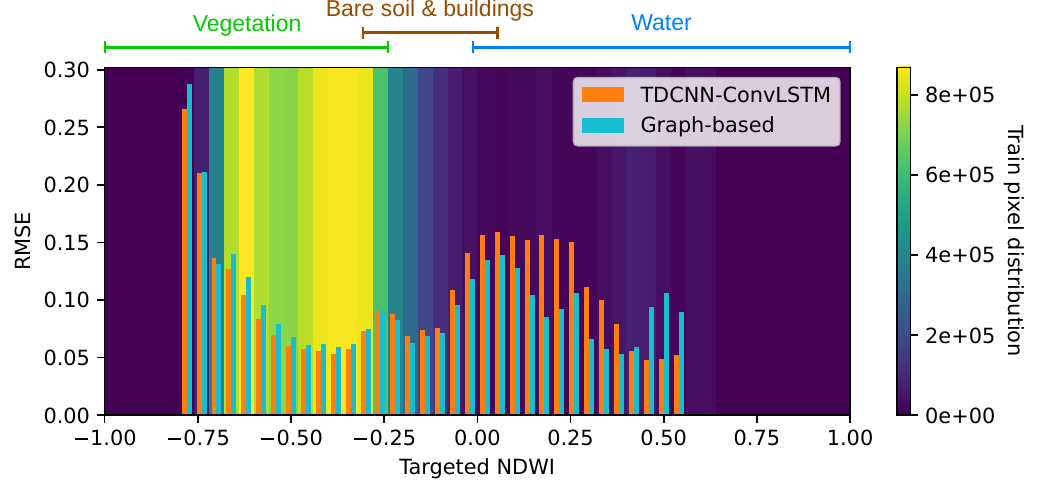}
    \caption{RMSE bar plot depending on the targeted NDWI for each pixel of the test set. The background color represents the training pixel distribution of the targeted NDWI.}
    \label{fig:gc_rmse_ndwi}
\end{figure}

We further analyse the graph-based model and the best-performing competitor, TDCNN-ConvLSTM, by comparing their performances depending on the targeted NDWI value, and by extension on the targeted land cover. By exploring the dataset, comparing it with other sources and in line with the spectral index design~\cite{mcfeeters1996use}, we consider positive NDWI values as water bodies, slightly negative values as roads, buildings and bare soil, and highly negative values as vegetation, such as forests, cultivated fields and grasslands. \Cref{fig:gc_rmse_ndwi} displays the RMSE values for the targeted NDWI values, while the background color displays the training pixel distribution to gauge the generalization ability of the model. Overall, it shows the superior performances of the graph-based model compared to TDCNN-ConvLSTM on impervious surfaces and water bodies, which are the focus of this study. However, we note high graph-based model errors for pixels with the highest NDWI values, which correspond mainly to the Lake Nemi site and are not reproduced in other sites. Further studies should be done to assess and explain this specific behavior. Considering the forecasting of vegetation, TDCNN-ConvLSTM achieves better results, probably thanks to a better understanding of the crop field dynamics, which are less periodic due to human actions, than water bodies which follow a natural seasonal cycle. Moreover, the performance of the models in relation to the distribution of training pixels highlights that the graph-based model generalizes better to less-viewed NDWI values.

\begin{figure}
    \centering   
    \includegraphics[width=.7\linewidth]{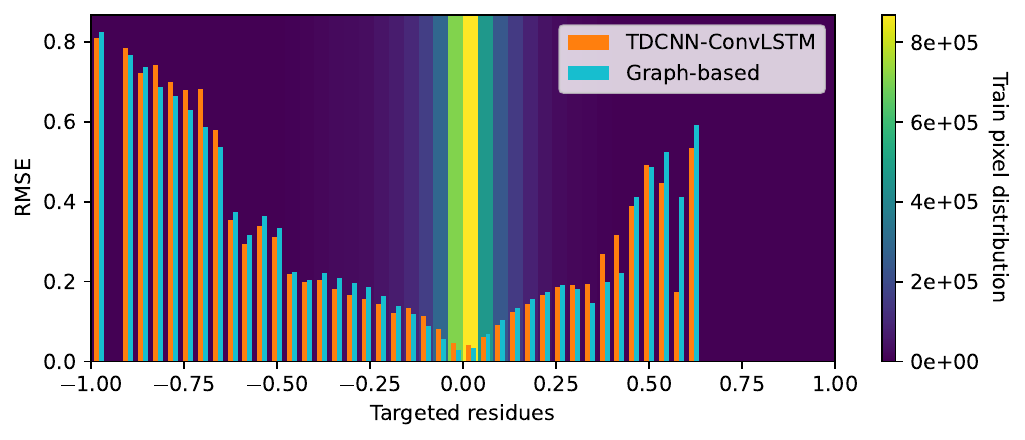}
    \caption{RMSE bar plot depending on the targeted residuals since the last acquisition for each pixel of the test set. The background color represents the training pixel distribution of the targeted residuals. Targeted residuals can theoretically range from $-2$ to $2$, but this does not happen in practice in the dataset.}
    \label{fig:gc_rmse_resid}
\end{figure}

It is also interesting to compare the graph-based model and TDCNN-ConvLSTM related to the type of changes they are able to predict. \Cref{fig:gc_rmse_resid} compares prediction errors with the magnitude of the expected change. The graph-based model tends to produce mainly small changes, surely linked to its residual approach. TDCNN-ConvLSTM seems better for slightly larger changes. We cannot comment on the extreme changes at the ends of the figure, as the number of pixels involved is small and insufficient to perform a statistically significant analysis. However, both models have a tendency towards greater error when significant changes occur between the last observed acquisition and the one to be predicted. This behavior is a consequence of the low temporal frequency of the input, which makes it difficult to capture the first signs of an abrupt phenomenon, whether natural (drought and flood) or anthropogenic (agricultural harvest).

\paragraph{Qualitative analysis}

\begin{figure*}
    \centering
    \subfloat{
        \includegraphics[width=\linewidth]{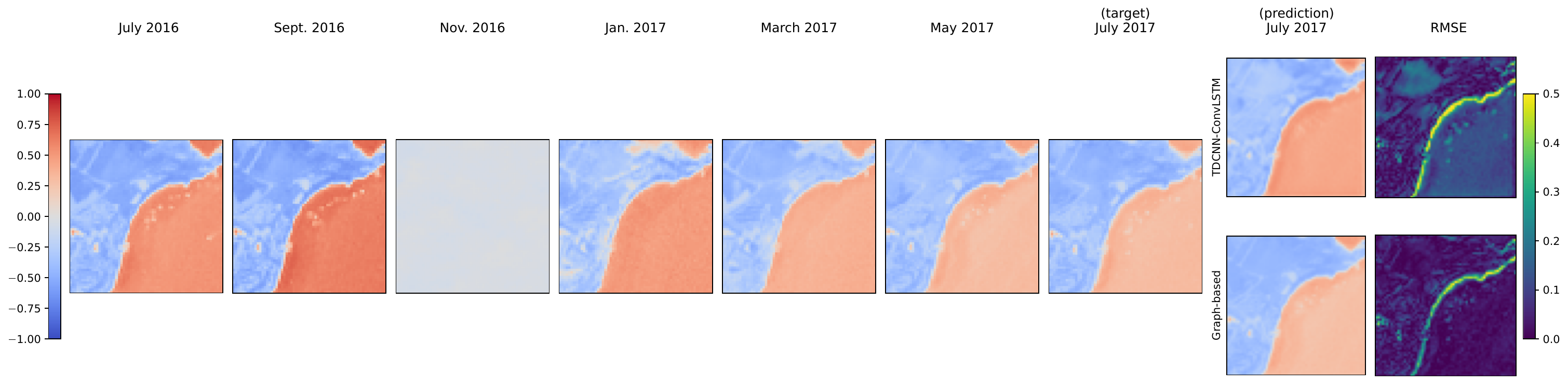}
    }
    \hfill
    \subfloat{
        \includegraphics[width=\linewidth]{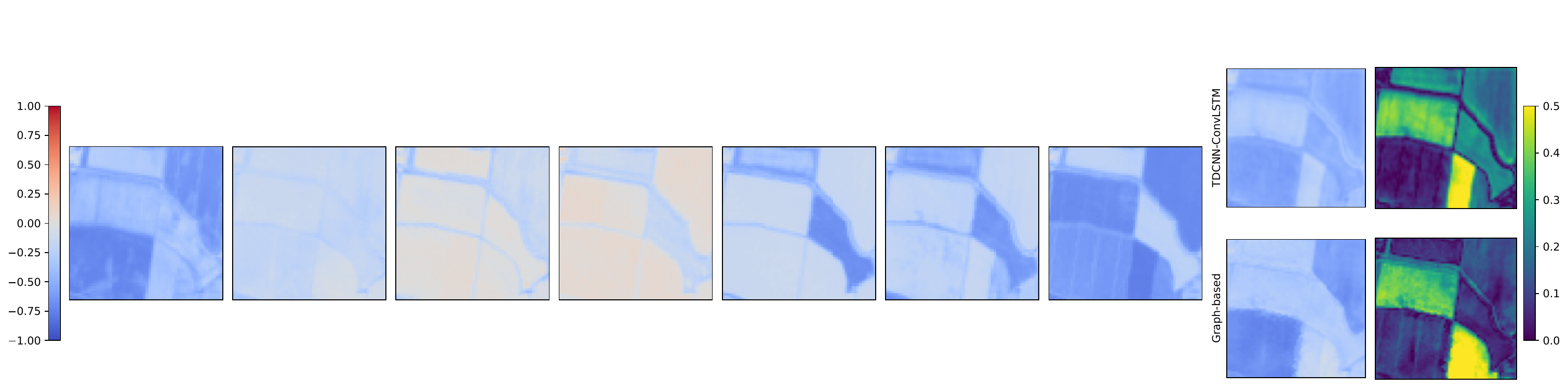}
    }
     
    \caption{Prediction of the next NDWI image of the series by TDCNN-ConvLSTM and our graph-based model. The input images range from July 2016 to May 2017. The objective is to predict an acquisition in July 2017. \textit{(top)} Lake Garda shores; despite data preprocessing including a cloud filtering processing, the November 2016 acquisition is cloudy. \textit{(bottom)} Agricultural area on the banks of the Po River.}
    \label{fig:gc_res_visual}
\end{figure*}

Visual analysis is performed based on \Cref{fig:gc_res_visual}, which displays two examples of NDWI time series and the predictions yielded by the graph-based model and TDCNN-ConvLSTM, with corresponding error maps. It focuses on two types of areas representative of the SEN2DWATER dataset: lakeshores and farmland.

Lakeshore areas with positive NDWI indicate open water, while variations over time reflect turbidity changes. In this SITS, turbidity has no strong seasonal pattern, \eg, the NDWI values in July~2017 are closer to May 2017 than July 2016. The graph-based model best predicted these non-seasonal dynamics. The image also shows terrestrial vegetation with a distinct pattern. Water basins support irrigation, influencing nearby vegetation, which, while seasonal, also depends on water availability. It is therefore interesting to analyze an application of the proposed predictions to nearby fields, as done with the second site of~\Cref{fig:gc_res_visual}.

For the second SITS representing farmland, NDWI distinguishes bare soil (near-zero values) from vegetation (negative values) and helps track seasonal harvesting. The July 2017 acquisition mirrors July 2016 but differs from May 2017, reflecting a seasonal cycle. Both models captured this pattern, as shown by the field prediction at the bottom left, though the graph-based model performed better, especially in the top-right fields, which nevertheless follow the same temporal pattern. However, the significant reconstruction errors which occur in the two central fields highlight that both models struggled with unseen abrupt changes, as already mentioned in the analysis of~\Cref{fig:gc_rmse_resid}.

\paragraph{Influence of the segmentation step}
\label{subsec:gc_mesh}

\begin{figure}
    \centering
    \includegraphics[width=.6\linewidth]{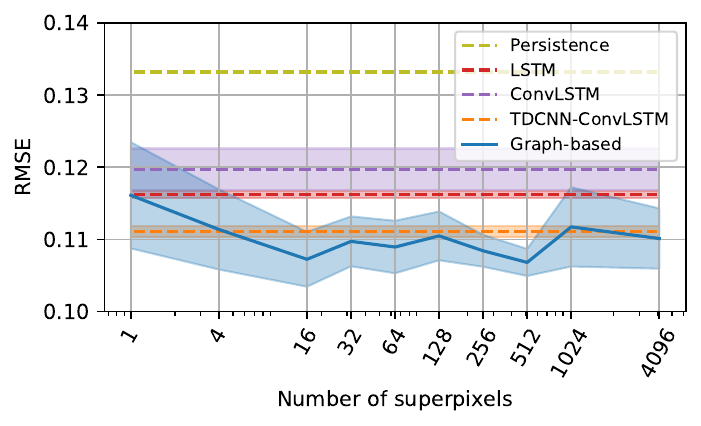}
    \caption{Influence of the number of superpixels used to create the mesh in the graph-based model. The mean and standard deviation are reported for three initializations of the model's parameters.}
    \label{fig:gc_results_nb_spxl}
\end{figure}

In this section, we explore the influence of the segmentation step, which is at the heart of the graph-based setting. In fact, the number of superpixels and their shape influence the topology of the mesh. \Cref{fig:gc_results_nb_spxl} reports the performance for an increasing number of segments ranging from 1 (one segment represents the whole patch) to 4,096 segments (each segment corresponds to one pixel).
The reconstruction error remains stable across a varying number of superpixels, as long as a minimal geometry is maintained. This suggests that SITS dynamics can be analyzed at a regional level rather than solely at the pixel level using a coarse geometric representation. Additionally, employing encoders and decoders to switch between fine geometry and mesh preserves pixel-level precision without compromising visual quality.

Furthermore, \Cref{tab:gc_results_multidate_spxl} displays the performance for two segmentation strategies---SLIC applied to the stack of input images or SLIC applied to the last image of the sequence (default setting). For this experiment, the number of superpixels is set to 256. Although results show a slight performance gain when applying SLIC to the last image, the performance measures are within the same range. We argue that applying SLIC to the stack of input images might be beneficial if cloudy images are not properly curated as it is the case in SEN2DWATER.

\begin{table}
\setlength{\tabcolsep}{2pt}
\begin{center}
\begin{tabular}{rccc}
\hline
 & RMSE $\downarrow$  & PSNR $\uparrow$ & SSIM $\uparrow$ \\
\hline
Last-date SLIC & \bfseries 0.1097\textsubscript{$\pm$0.0035} & \bfseries26.42\textsubscript{$\pm$0.27} & \bfseries0.8170\textsubscript{$\pm$0.0070} \\
Multi-date SLIC & 0.1101\textsubscript{$\pm$0.0034} & 26.12\textsubscript{$\pm$0.45} & 0.8151\textsubscript{$\pm$0.0076} \\
\hline
\end{tabular}
\end{center}
\caption{Influence of the segmentation strategy. The mean and standard deviation of RMSE, PSNR, and SSIM metrics are reported for three initializations of the model's parameters. Bold values display the best performance for each metric.}
\label{tab:gc_results_multidate_spxl}
\end{table}

\paragraph{Influence of the length of the time series}
\label{subsec:gc_tlen}

We evaluate the impact of input time series length on the graph-based model, testing performance with the last 1 to 6 acquisitions, each about two months apart. Unlike recurrent models, our graph-based model lacks a built-in causal mechanism and must learn temporal dependencies directly from the data. Consequently, increasing the number of input acquisitions helps the model capture these dependencies, as shown in \Cref{tab:gc_results_tlen}, where error decreases with longer series. The best performance is achieved covering a full year, allowing the model to forecast at an already-seen season. Interestingly, models using only one or two images perform worse than the weak persistence baseline, emphasizing the need for longer time series to capture annual variations effectively. Although limited by the filtered dataset, we could expect better results with more frequent acquisitions and a greater temporal coverage.

\begin{table}[ht!]
\setlength{\tabcolsep}{2pt}
\begin{center}
\begin{tabular}{ccccc}
\hline
\# input dates & RMSE $\downarrow$  & PSNR $\uparrow$ & SSIM $\uparrow$ \\
\hline
1 & 0.1409\textsubscript{$\pm$0.0085} & 24.28\textsubscript{$\pm$0.65} & 0.7593\textsubscript{$\pm$0.0218} \\
2 & 0.1455\textsubscript{$\pm$0.0013} & 23.72\textsubscript{$\pm$0.18} & 0.7294\textsubscript{$\pm$0.0098} \\
3 & 0.1330\textsubscript{$\pm$0.0157} & 24.75\textsubscript{$\pm$1.28} & 0.7538\textsubscript{$\pm$0.0421} \\
4 & 0.1319\textsubscript{$\pm$0.0020} & 24.77\textsubscript{$\pm$0.09} & 0.7662\textsubscript{$\pm$0.0092} \\
5 & 0.1200\textsubscript{$\pm$0.0034} & 25.42\textsubscript{$\pm$0.40} & 0.7957\textsubscript{$\pm$0.0082} \\
6 & \bfseries 0.1097\textsubscript{$\pm$0.0035} & \bfseries 26.42\textsubscript{$\pm$0.27} & \bfseries 0.8170\textsubscript{$\pm$0.0070} \\
\hline
\end{tabular}
\end{center}
\caption{Forecasting results of the graph-based model for varying input time series length. The mean and standard deviation are reported for three initializations of the model's parameters. Bold values display the best performance for each metric.}
\label{tab:gc_results_tlen}
\end{table}

\subsubsection{Discussion}

This case study illustrates another downstream task that still uses graphs, albeit with different role, design and processing choices. SITS forecasting is a pixel-to-pixel task, thus, in contrast to the previous case study, graphs are used both to link pixel and object scales and to model inter-object relationships. Indeed, it brings a prior spatial information using geographical objects to learn local dynamics of water basins and their surroundings. In this work, we use a simple region adjacency graph, which prevents to capture long-range dependencies. To improve generalization capability of the model in larger images, the proposed architecture should be extended to handle multiple scales, \eg, using multi-meshes or hierarchical meshes. 
Besides, the segmentation step giving prior spatial information to the model still results from an arbitrary setting, \ie, the number of superpixels. 
Other segmentation algorithms, which can produce different types of regions, could influence performance, and thus optimizing this object segmentation step is a means of improvement.
Moreover, temporal information is not processed using a graph, but embedded at pixel level with a simple MLP not tailored to learn temporal patterns from time series. This results from the need to use a static spatial graph to represent the prior spatial information which is considered constant. It appears as a current limitation for highly dynamics phenomena where relationships between the object can vary. Solutions including time-adaptive meshes~\cite{baker2005adaptive, pfaff2021learning, hu2023graph} and discrete-time dynamic graphs~\cite{yang2024dynamic} need to be explored in order to retain the representativeness of graphs to model a time-evolving scene. 
Finally, as in the previous study, interpretability of models and results is crucial to motivate further developments. A way to improve it is realizing a more robust analysis of the results by considering disaggregated results as done in~\Cref{fig:gc_rmse_ndwi} and \Cref{fig:gc_rmse_resid}. The distinction can be done on various criteria which add semantics, such as seasonality and rareness of the events, classification of the phenomenon or of the geographical site. Therefore, such an analysis should enhance the evaluation and comparison of different models and help explain the pros and cons of each model. We hope that this case study may help future works to better consider advantages and current limitations of graphs for a SITS forecasting task, thus encouraging the development of new methods.

\section{Challenges and Perspectives}
\label{sec:perspectives}

Based on our extensive review and the practical insights gained from the two case studies, we highlight several key perspectives for the future of the graph-based processing and analysis of SITS.

\paragraph*{Spatio-temporal objects}
In \Cref{sec:st_obj}, the definition of spatio-temporal objects is deliberately kept as broad as possible to encompass future developments. Specifically, existing works adopting the perdurantism often disregard fully arbitrary segmentation of SITS, which results in spatio-temporal objects with an irregular 3D shape (two spatial dimensions and one temporal). A number of methodological and technical hurdles need to be overcome to achieve effective spatio-temporal segmentation and to process the resulting objects.

\paragraph*{Graph interpretability} Despite the whole section reviewing the different design choices to build a spatio-temporal graph (\Cref{sec:graphExtract}), they remain poorly explained in practical use cases or are only available for specific cases. Graph structure is flexible and demonstrates high potential if combined with a data-centric workflow. A significant improvement would be to democratize interpretability tools to quantitatively explain design choices for specific applications, while maintaining the pluggability of the design pipeline. Prior expert knowledge would therefore be supported and extended by indicators on application data.

This improvement concerns both the graph construction and its processing. The latter one still misses arguments in favor of a time-then-space, a space-then-time or a time-and-space approach. It certainly depends on the application and the data, but it would be interesting to carry out a more in-depth study to gain a better understanding of architecture design choices according to those applications and data.

\paragraph*{End-to-end graph-based pipeline} Without losing sight of the need for interpretability and explainability, integrating graph structure learning from SITS directly into a fully end-to-end graph-based approach should avoid arbitrary choices regarding the segmentation step, and allow to build data-dependent graphs. This future prospect has already started to be explored in some previously cited works for region segmentation~\cite{yang2020superpixel}, feature extraction~\cite{ma2024deep} and edge learning~\cite{cachay2021world}. However, to the best of our knowledge, a fully end-to-end pipeline with both graph structure learning and feature processing is still missing.

\paragraph*{Scalability and dynamic scenarios} Spatio-temporal graphs appear as a solution to pixel variability due to high resolution. Graph-based approaches also solve, to some extent, a scalability problem by reducing the number of working units. However, a spatio-temporal graph can still be large, with a high number of objects and edges, which calls for improvements in large-scale processing as shown in a previous case study (\Cref{sec:gden}). This scalability problem exists for every processing methods, involving learning or not, and for all downstream tasks previously mentioned, \ie, visualization, graph mining, classification and forecasting.

The abundance of data in Earth observation is the main cause of the scalability problem, with a constant flow of new data which are sparsely labelled. Various propositions of solution to handle this data volume exist, such as foundation models~\cite{zhu2024foundations,marsocci2024pangaea} and continual learning. Exploring these solutions in the setting of spatio-temporal graphs could be promising. Adapting the spatio-temporal framework to dynamic scenarios to allow real-time updating, for example developing continuous-time dynamic graphs~\cite{kazemi2020representation,barros2021survey} adapted to SITS, could also improve the scalability of the model.

\paragraph*{Extending spatio-temporal graph to new modalities}
In this paper, graphs are principally used as tool to analyze and process SITS from a single source, mainly optical. Only a few of the aforementioned works have tackled other data sources, such as SAR~\cite{yi2014, debusscher2019object, hagensieker2017tropical}, atmospheric~\cite{keisler2022, lam2023learning} and oceanic~\cite{li2021, xue2019} variables. However, \Cref{sec:SITSacqui} gave an overview of the wide range of sensors imaging the Earth. This diversity of sources brings development opportunities to adapt to each setting, even those with little data currently available (hyperspectral SITS, off-nadir and multi-angle acquisitions, 3D point clouds sequence). Graph-based approaches can also be adapted for data similar to SITS, such as aerial or drone imagery and numerical weather simulations.

Beyond the analysis of SITS data, we envision that graph-based frameworks, with their ability to model spatio-temporal relationships, will serve as powerful tools for integrating and analyzing multiple data sources collected at varying spatial and temporal resolutions. In fact, graphs are already applied to many different modalities of Earth observation~\cite{zhao2025beyond}, from remote sensing to in situ measurements, and they can naturally tackle the heterogeneous nature of these data. Moreover, the flexibility of graphs makes it possible to envisage multimodal fusion that is more respectful of raw data and their geometry, without intermediate approximation such as realignment. Graphs hold potential for multisource and multimodal fusion within a unified data structure.

\paragraph*{A cross-disciplinary field}
By virtue of its history and the themes it addresses, the use of spatio-temporal graphs for SITS is at the intersection of several disciplines, both theoretical (mathematics with graph theory, computer science with data structure, data science for graph processing, and metaphysics for reflection on the nature of space and time and its objects) and thematical (remote sensing for data acquisition and understanding, application fields such as geography, meteorology and oceanography). By its very nature, this is fertile ground for transdisciplinary work. The collaboration between experts from different backgrounds should enable them to bring the latest advances from their different fields to the graph-based SITS analysis.

\section{Conclusion}
\label{sec:ccl}

In this paper, we highlight the potential and versatility of graph-based pipelines for analyzing geospatial temporal data. As an extension of the OBIA paradigm, graph-based approaches improve the SITS processing by incorporating spatial and spatio-temporal relationships. Graphs help to preserve the spatio-temporal structure of SITS while retaining the representation power and memory efficiency of using geographical objects as working units. It offers also new strategies to process SITS data, from visualisation tools to advanced deep learning techniques.

This review shows how the use of graphs for SITS analysis has evolved during the last twenty years. The recent resurgence of this data structure in various fields, especially with graph neural networks, highlights its efficiency, but also appeals for improvement to overcome its shortcomings, namely the need for scalability, interpretability and data-dependent design, potentially learned. A fair comparison of graph-based methods with their pixel-based counterparts is also crucial when studying them. In this review, we showcase the potential of graphs for SITS for two original case studies, illustrating both strengths and limitations which should be addressed in the future. Finally, we discuss numerous perspectives for resolving the limitations cited and extending the use of spatio-temporal graphs to new modalities. We hope that this work will foster the emergence of this field among different scientific communities and encourage transdisciplinary collaborations.

\section*{Acknowledgments}
The authors thank the French Spatial Agency (CNES, Centre National d'Études Spatiales) and the Brittany region (\textit{Groupement d'Intérêt Scientifique Bretagne Télédétection}, GIS BreTel) for their financial support. This work was granted access to the HPC resources of IDRIS under the allocation 2024-AD011014108R1 made by GENCI. Charlotte Pelletier is partially funded through project DECOL ANR-23-CE56-0003. 

\bibliographystyle{IEEEtranN}
\bibliography{main}

\end{document}